\documentclass{article}

\usepackage{microtype}
\usepackage{graphicx}
\usepackage{subcaption}
\usepackage{booktabs} 

\usepackage{hyperref}

\usepackage[accepted]{icml2026_fogen}

\usepackage{amsmath}
\usepackage{amssymb}
\usepackage{mathtools}
\usepackage{amsthm}

\usepackage[capitalize,noabbrev]{cleveref}

\theoremstyle{plain}
\newtheorem{theorem}{Theorem}[section]
\newtheorem{proposition}[theorem]{Proposition}

\theoremstyle{definition}

\theoremstyle{remark}

\usepackage{booktabs}
\usepackage{array}
\usepackage{tabularx}
\usepackage{multirow}
\usepackage{multicol}
\usepackage{wrapfig}
\usepackage{subcaption}   
\usepackage{enumitem}

\usepackage{algorithm}
\usepackage{algpseudocode}
\DeclareUnicodeCharacter{25C6}{$\blacklozenge$}

\usepackage{amsmath,amsfonts,amsthm,bm}
\usepackage{bbm}
\usepackage{color,soul}

\newcommand{\RNum}[1]{\uppercase\expandafter{\romannumeral #1\relax}}
\newcommand{\Rnum}[1]{\lowercase\expandafter{\romannumeral #1\relax}}

\def\Secref#1{Section~\ref{#1}}
\def\Appref#1{Appendix~\ref{#1}}

\def\Figref#1{Figure~\ref{#1}}

\def\Tabref#1{Table~\ref{#1}}

\def\Secref#1{Section~\ref{#1}}

\def\eqref#1{equation~(\ref{#1})}
\def\Eqref#1{Equation~(\ref{#1})}

\def\peqref#1{(\ref{#1})}

\def\0{\bm{0}} 
\def\1{\bm{1}}

\def\rvw{{\mathbf{w}}}
\def\rvx{{\mathbf{x}}}

\DeclareMathAlphabet{\mathsfit}{\encodingdefault}{\sfdefault}{m}{sl}
\SetMathAlphabet{\mathsfit}{bold}{\encodingdefault}{\sfdefault}{bx}{n}

\usepackage[textsize=tiny]{todonotes}

\icmltitlerunning{Not Every Time and Frequency Need to Be Forgotten in Diffusion Unlearning}

\begin{document}

\noindent{\small ICML 2026 Workshop on Foundations of Deep Generative Models:
Understanding Memorization, Generalization, and Reasoning}
\vskip 0.35in

\icmltitle{Not Every Time and Frequency Need to Be Forgotten in Diffusion Unlearning}

\begin{center}
  \begin{tabular}{ccc}
    Jinseong Park &  Mijung Park \\
    \texttt{jinseong@kias.re.kr} &
    \texttt{mijung.park@ubc.ca} \\
     Korea Institute for Advanced Study &  University of British Columbia
  \end{tabular}
\end{center}

  \icmlkeywords{Machine Learning, ICML}

\vskip 0.3in



\printAffiliationsAndNotice{}  

\begin{abstract}
Data unlearning aims to remove the influence of specific training samples from a trained model.
In fine-tuning methods, data unlearning relies primarily on loss maximization over forget samples, which often leads to quality degradation or incomplete forgetting.
Existing methods perform unlearning uniformly across diffusion stages, ignoring diffusion dynamics from noise to data. Our systematic study of diffusion phases shows that forgetting in diffusion models is \textit{uneven} across time and frequency, with theoretical justification of distributive distortion and forgetting-utility trade-off.
By selectively forgetting time and frequency in diffusion models, we achieve both higher unlearning success rates and improved generation quality across diverse settings, including both conditional and unconditional scenarios. We also introduce an improved SSCD metric that measures dissimilarity using a normalized perturbation distance. Together, we provide practical insights for understanding and improving data unlearning in diffusion models.
\end{abstract}

\section{Introduction}
\begin{wrapfigure}{r}{0.49\textwidth}
    \centering \vspace{-5mm}
    \includegraphics[width=0.49\textwidth]{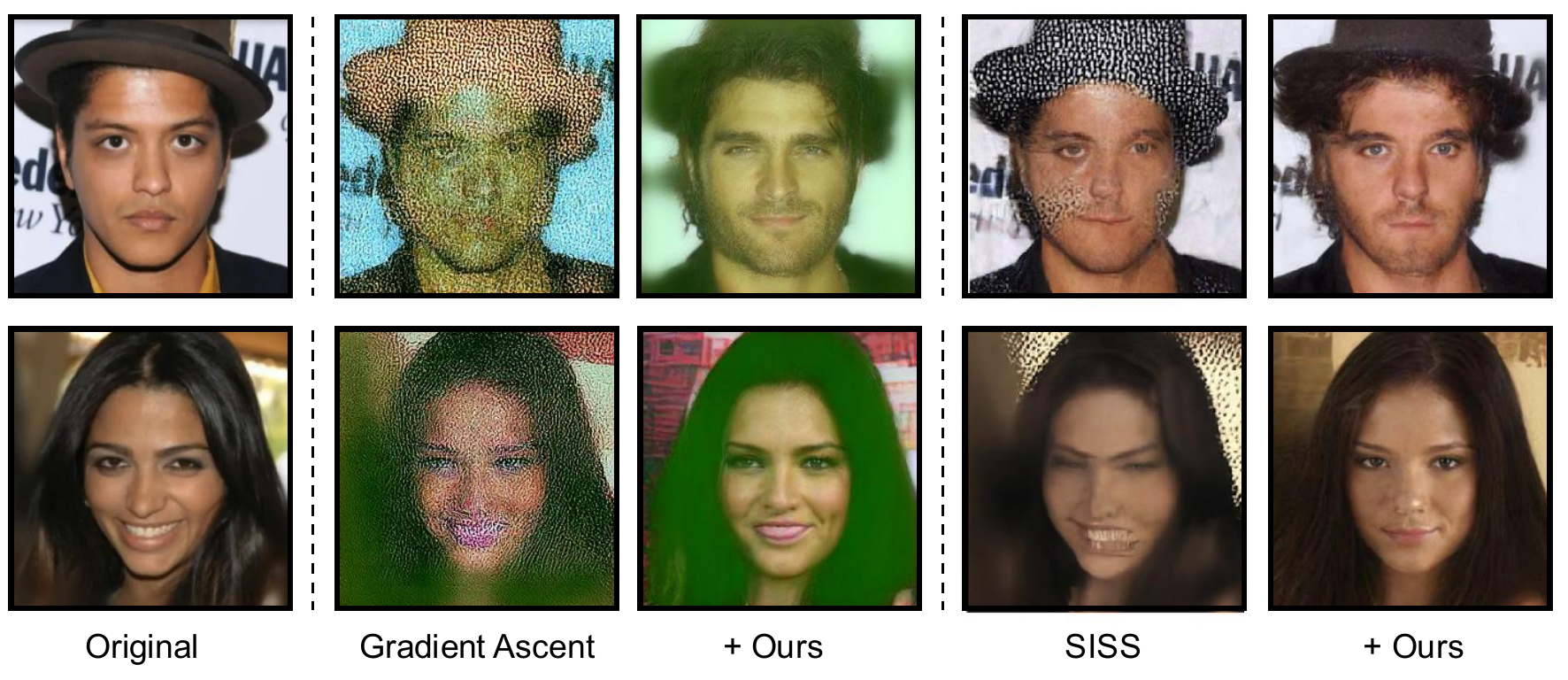}\vspace{-1mm}
    \caption{Illustration of quality degradation of unlearned images after data unlearning. \textbf{Left:} two images to unlearn. \textbf{Middle:} Gradient Ascent (GA, first column) and GA with our approach (second column). \textbf{Right:} SISS \cite{alberti2025data} (first) and SISS with our approach (second). Our approach improves the quality of existing unlearning methods.}\label{fig:proposed_method}
\end{wrapfigure}
The ability to remove the influence of some training samples from a learned model, often referred to as \textit{machine unlearning} \citep{bourtoule2021machine}, has become increasingly important. Regulatory frameworks such as the ``right to be forgotten” in the General Data Protection Regulation (GDPR) by the European Union and growing concerns about sensitive or proprietary data have created demand for methods that allow models to forget without costly retraining from scratch. Recently, with the development of diffusion models \citep{ho2020denoising}, unlearning the unsafe concept or memorization has been actively explored through finetuning \citep{wang2025diffusionnpo}, output filtering \citep{yoon2025safree}, and training-free sampling \citep{kim2025training, kim2026safetyguided}.


In machine unlearning with generative models, two major objectives exist: (a) concept or class unlearning, which refers to preventing the generation of a particular concept or class in samples \citep{fan2024salun}, and (b) data unlearning, which focuses on removing individual samples rather than concepts. As noted by \citet{alberti2025data}, data unlearning is less studied than concept unlearning. Furthermore, removing a single individual while preserving the generation of similar outputs is more challenging than deleting the entire mode of similar examples.
Finetuning-based unlearning methods, such as gradient ascent \citep{alberti2025data} or negative preference optimization \citep{wang2025diffusionnpo}, rely on loss maximization objectives.
The generated images often become noisier and aesthetically degraded, while forgetting might not be complete nor precise, as shown in \Figref{fig:proposed_method}.

To mitigate the drawbacks of the data unlearning in diffusion models, we analyze the dynamics of data unlearning. We argue that forgetting does not occur uniformly, but rather unevenly across time and spectral domains.
Diffusion models learn phase-dependent behaviour during training \citep{choi2022perception}: later diffusion time steps close to Gaussian noise capture coarse semantics, while earlier time steps close to the data refine fine-grained details. However, previous unlearning methods treat all diffusion time steps equally, potentially leading to the inadvertent forgetting of unintended details, as illustrated in \Figref{fig:toy_exp}. Furthermore, we investigate the spectral domain to selectively remove unwanted high-frequency components in the forget set. 
%
%
%
Overall, our framework effectively implements forgetting and preserves sample quality by focusing on selective components in both the time and frequency domains.
We summarize our contributions below:
\begin{itemize}[topsep=1pt,itemsep=1pt,parsep=1pt]
    \item We provide a \textit{systematic} study of data unlearning in diffusion models in time and spectral space, considering both conditional and unconditional scenarios.
    \item Our time- and frequency-selective unlearning approach is \textit{simple, compatible} with existing unlearning objectives, and performs effective unlearning across various settings.
    \item We devise an \textit{improved} metric, SSCD$^{\text{norm}}$. Unlike the original SSCD \citep{pizzi2022self}, which only measures dissimilarity, SSCD$^{\text{norm}}$ standardizes perturbation distance for measuring quality.
\end{itemize}

\section{Background}
\subsection{Diffusion Models}
Diffusion models \citep{ho2020denoising} 
aim to generate samples from an unknown data distribution by considering forward (diffusion) and reverse (denoising) processes. 
The forward process gradually injects noise into samples $\rvx_0$ drawn from the data distribution $p_0$, forwarding them into a fixed standard Gaussian distribution $p_T = \mathcal{N}(\boldsymbol{0},\boldsymbol{I})$. The corresponding forward stochastic differential equation (SDE) is: 
\begin{equation}
    d\rvx_t = f(\rvx_t,t)dt + g(t)d\rvw_t,
\end{equation}
where $f:\mathbb{R}^d \times [0,T] \rightarrow \mathbb{R}^d$ represents the drift function, $g:[0,T] \rightarrow \mathbb{R}$ denotes the diffusion coefficient, and $\rvw_t$ is the Wiener process. As diffusion models learn to reverse the above process, the reverse SDE \citep{anderson1982reverse} is formulated as follows:
\begin{equation}
    d\rvx_t = [f(\rvx_t,t) - g(t)^2\nabla_{\rvx_t} \log p(\rvx_t)]dt + g(t)d\bar{\rvw}_t,
\end{equation}
where $p(\rvx_t)$ refers to the probability density of $\rvx_t$. Diffusion models learn to match the score function $\nabla_{\rvx_t} \log p(\rvx_t)$ (parameterized by a score network $s_\theta(\rvx_t,t)$) for denoising \citep{song2021scorebased}. 
With standard DDPM forward noising process $q(\rvx_t|\rvx_0) = \mathcal{N}\left(\sqrt{\bar\alpha_t}\rvx_0,(1-\bar\alpha_t)\mathbf{I}\right)$, 
the conditional score function can be written as $\nabla_{\rvx_t}\log q(\rvx_t|\rvx_0)$. With a weighting factor $w_t$, the training objective $\mathcal{L}_\mathcal{D}(\theta)$ is
\begin{equation}
    \mathbb{E}_{\rvx_0\sim p_0}\mathbb{E}_{t\in[0,T]}\Bigl[
      w_t\bigl\| s_\theta(\rvx_t,t) - \nabla_{\rvx_t}\log q(\rvx_t|\rvx_0) \bigr\|_2^2
    \Bigr].
    \label{eq:diffusion_loss}
\end{equation}

\subsection{Data Unlearning in Diffusion Models}
\paragraph{Machine Unlearning. } Suppose a model $\theta^{*}$ is already trained on a dataset $\mathcal{D}$. Then, our goal is to delete the influence of the forget set $\mathcal{D}_F$ from $\theta^{*}$, while maintaining the model utility on the retain set $\mathcal{D}_R=\mathcal{D} \setminus \mathcal{D}_F$. Specifically, our goal is to fine-tune the trained model $\theta^{*}$ using the forget dataset $\mathcal{D}_F$, often using the retain dataset $\mathcal{D}_R$ to mitigate quality degradation \citep{bourtoule2021machine}.

\paragraph{Diffusion Unlearning. }
Unlearning in diffusion models can be categorized as concept unlearning and data unlearning. \textbf{Concept unlearning} \citep{gandikota2023erasing} refers to prohibiting a diffusion model from producing images categorized in a particular type of high-level concept, including not safe for work (NSFW) material (nude and violent images, for example), or copyrighted content \citep{gandikota2023erasing, srivatsan2025stereo, park2024direct}.

Due to the high flexibility of diffusion models in sampling, unlike GANs \citep{goodfellow2020generative}, training-free steering methods have been actively investigated \citep{singhal2025a,kim2025training,koulischer2025dynamic}. These methods leverage guidance techniques to repel the generation from specific points or embeddings, achieving concept erasure without finetuning. 


\textbf{Data unlearning}, on the other hand, is closely related to individual data memorization and aims to remove specific data examples in accordance with the ``Right to be Forgotten" by the aforementioned GDPR. For example, if a user requests the deletion of a particular face image used to train a diffusion model, the goal is to eliminate the influence of that image from the trained model.
In contrast to concept erasing, for which various practical solutions have been developed, data unlearning remains relatively underexplored \citep{alberti2025data}. In general, the objective of diffusion unlearning $\min_\theta \mathcal{L}_{\text{UL}}(\theta)$ is
\begin{equation}
\begin{cases}
\max_\theta \mathcal{L}_{\mathcal{D_F}}(\theta) = \min_\theta -\mathcal{L}_{\mathcal{D_F}}(\theta),
& \text{$\mathcal{D}_F$ only} \\
\min_\theta \bigl( - \mathcal{L}_{\mathcal{D_F}}(\theta)+ \mathcal{L}_{\mathcal{D_R}}(\theta) \bigr), & \text{$\mathcal{D}_F$ and $\mathcal{D}_R$. } 
\end{cases}
\label{eq:unlearn}
\end{equation}

Gradient ascent (GA) or negative gradient on forget data samples $\mathcal{D}_F$ is a key component for data unlearning \citep{golatkar2020eternal}. For GA, the loss function on the forget set $\mathcal{L}_{\text{GA}}(\theta)$ is formulated as follows:
\begin{equation}
-\underbrace{\mathbb{E}_{ \rvx_0 \sim \mathcal{D}_F}\mathbb{E}_{t\in[0,T]}
\Bigl[
  w_t\bigl\| s_\theta( \rvx_t,t) - \nabla_{ \rvx_t}\log q( \rvx_t| \rvx_0) \bigr\|_2^2
\Bigr]}_{\text{Negative loss in \Eqref{eq:diffusion_loss} on forget set }\mathcal{D}_F}.
\label{eq:ga}
\end{equation}
EraseDiff \citep{wu2025erasing} replaces the true score of the forget samples with a randomly sampled noisy image. 
SISS \citep{alberti2025data} investigated the unlearning as a mixture distribution of forget and retain data with importance sampling. 
In conclusion, previous methods investigate how to formulate \Eqref{eq:unlearn} well.

\begin{figure*}[!t]
    \centering \vspace{-2mm}
    \includegraphics[width=0.8\textwidth]{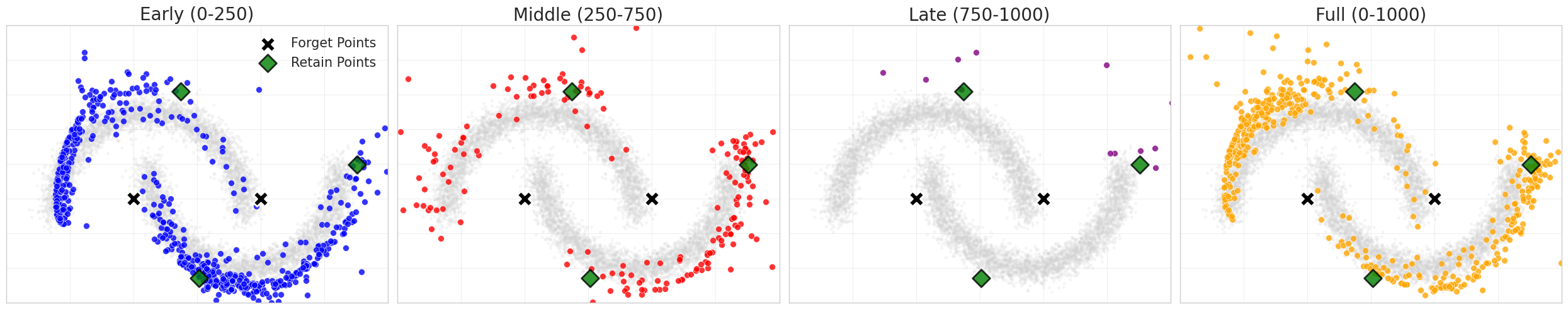}\vspace{-1mm}
    \caption{Illustration of unlearning a specific stage of time steps. See Section \ref{subsubsection:motivation} for details.}
    \label{fig:toy_exp} \vspace{-3mm}
\end{figure*}
\subsection{Evaluation of Data Unlearning in Diffusion Models}
\label{sec:scenario}
We follow the evaluation of \citet{alberti2025data} based on unconditional or conditional scenarios.

\textbf{Memorization on Unconditional Reconstruction.}
In unconditional generation, memorization is assessed by the diffusion model's ability to reconstruct the training data. 
In particular, we noise a clean image $\rvx_0$ via the DDPM forward process to an intermediate time step $\rvx_t$ (e.g., $t=250$) where the model can easily reconstruct, and measure the similarity between the original $\rvx_0$ and the reconstructed $\hat \rvx_0$ using model $\theta$ from $\rvx_t$.

\textbf{Memorization on Conditional Guidance.} 
In conditional generation, conditioning on text or class embeddings can lead to stronger overfitting to the conditional distribution over the images given those conditions \citep{zhai2024membership}. Therefore, measuring the effects of condition $\epsilon_\theta(\rvx_t, e_p)$ or its guidance norm $\|\epsilon_\theta(\rvx_t, e_p) - \epsilon_\theta(\rvx_t, e_\emptyset)\|^2_2$ is crucial to track conditional memorization \cite{wen2024detecting,jain2025classifier,alberti2025data}, with conditional embedding $e_p$ and null embedding $e_\emptyset$. Thus, unlike unconditional memorization, its evaluation starts by sampling from pure noise $t=T=1000$ with the conditions.

\section{Time and Frequency Selection for Data Unlearning in Diffusion Models}

In this section, we first empirically investigate the effects of time steps and frequency on diffusion unlearning and propose a selective framework for data unlearning, followed by a theoretical analysis.

\subsection{Not Every Time Step Needs to Be Forgotten}
\label{sec:time}

\subsubsection{Motivation}\label{subsubsection:motivation}

\begin{figure}[!t]
    \centering \vspace{-2mm}
    \begin{minipage}{0.49\textwidth}
        \centering
        \begin{subfigure}[t]{0.49\linewidth}
            \centering
            \includegraphics[width=0.85\linewidth]{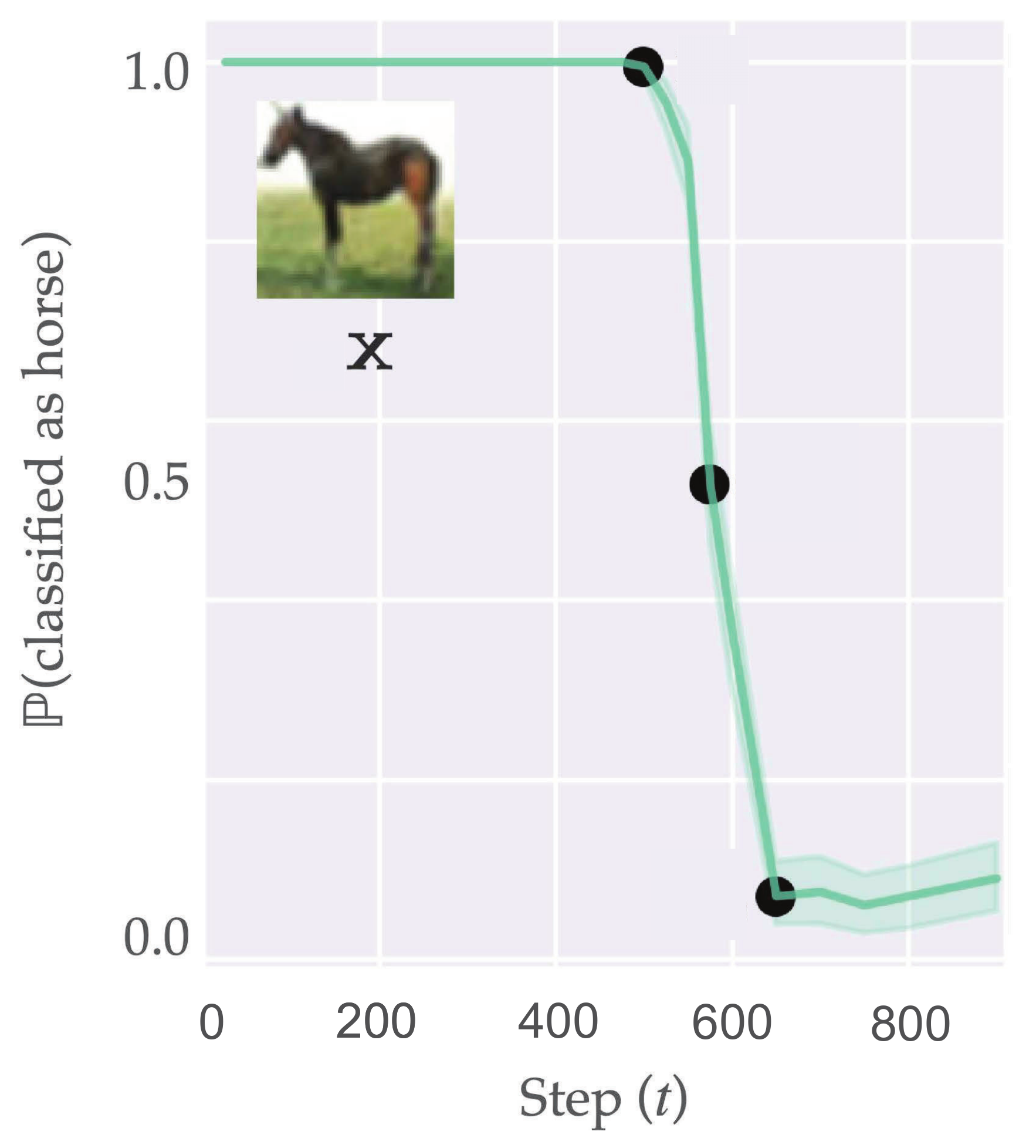}
            \caption{Speciation time $t_S$ (Phase I-II) \citep{georgiev2023journey}}
            \label{fig:stage_analysis_middle}
        \end{subfigure}
        \hfill
        \begin{subfigure}[t]{0.48\linewidth}
            \centering
            \includegraphics[width=0.83\linewidth]{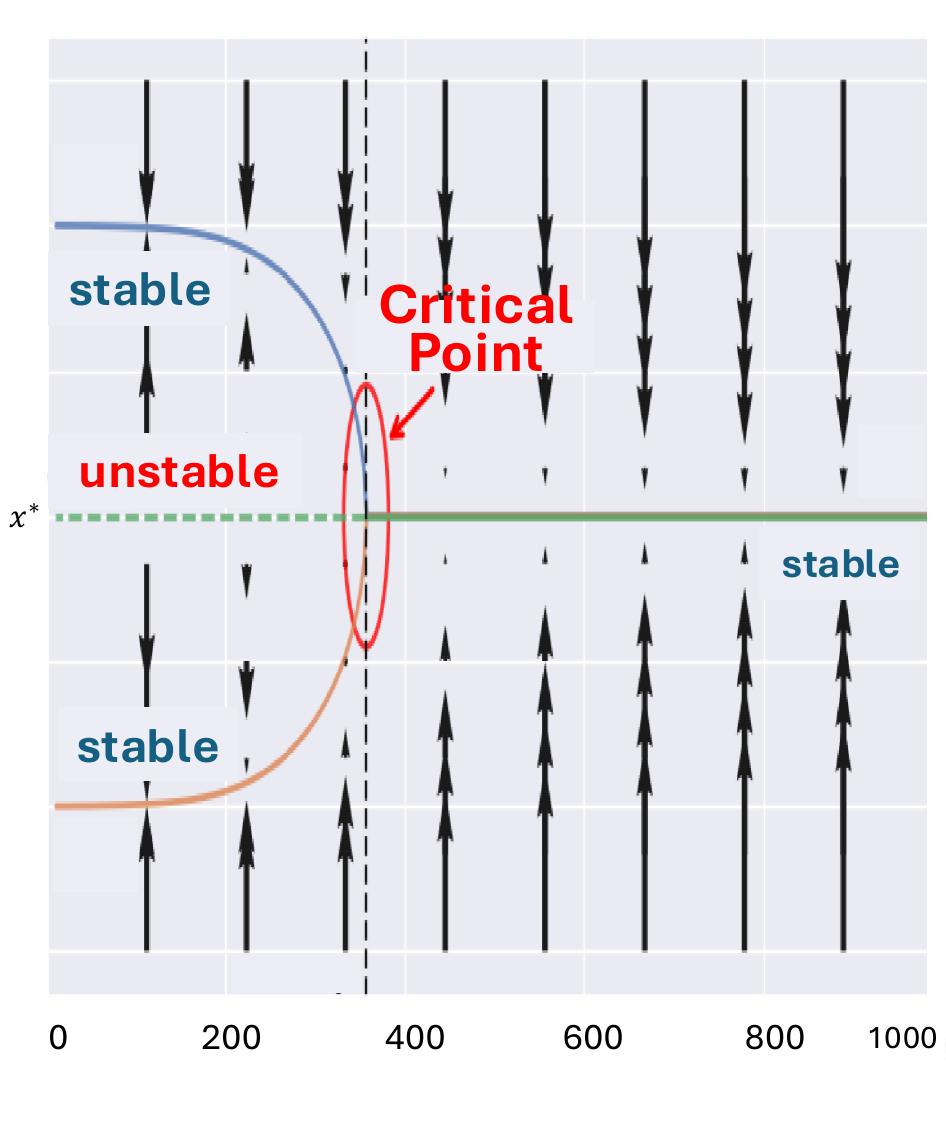}
            \caption{Collapse time $t_C$ (Phase II-III) \citep{raya2023spontaneous}}
        \end{subfigure}
        \caption{Empirical evidence on transition points of speciation time $t_S$ and collapse time $t_C$.}
        \label{fig:stage_analysis_existing}
    \end{minipage}
    \hfill
    \begin{minipage}{0.49\textwidth}
        \centering
        \begin{subfigure}[t]{0.49\linewidth}
            \centering
            \includegraphics[width=\linewidth]{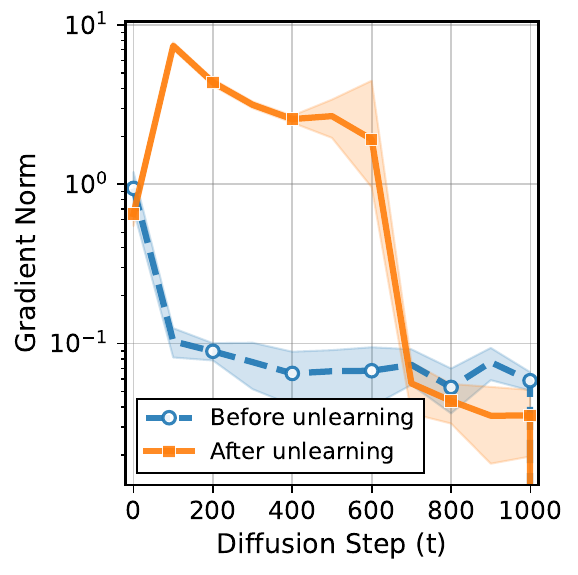}
            \caption{Gradient norm of forget set during unlearning}
            \label{fig:distribution_time_b}
        \end{subfigure}
        \hfill
        \begin{subfigure}[t]{0.48\linewidth}
            \centering
            \includegraphics[width=\linewidth]{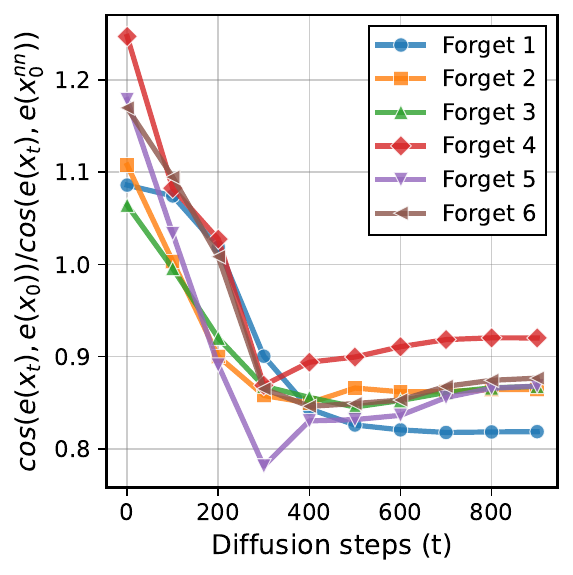}
            \caption{Embedding similarity of $\rvx_t$ w.r.t. $\rvx_0$ and $\rvx_0^\text{nn}$}
            \label{fig:distribution_time_a}
        \end{subfigure}
        \caption{Analysis of diffusion time steps.}
        \label{fig:distribution_time}
    \end{minipage}
    \vspace{-3mm}
\end{figure}

Several theoretical papers analyze the dynamics of diffusion models \citep{li2024criticalwindowsnonasymptotictheory, Biroli2024-pa, Sclocchi.pnas.2408799121, pmlr-v267-li25at}. Starting from the Gaussian noise, diffusion models learn distinct attributes across different phases. \textbf{Phase I} involves learning coarse features at late time steps near time $T$. \textbf{Phase II} involves generating content at intermediate steps, and \textbf{Phase III} involves refining details toward convergence at early time steps near $ t=0$ \citep{choi2022perception}.
\citet{Biroli2024-pa} describes the diffusion dynamics in terms of these three distinct phases divided by two important transition points in time: (a) \textbf{speciation} time $t_S$; and (b) \textbf{collapse} time $t_C$. 
The denoising trajectories begin to separate by semantic class at the \textbf{speciation time} $t_S$ (phase I-II), for example via embedding similarity or consistency scores. For $t > t_S$, trajectories are nearly class-agnostic noise, whereas for $t < t_S$, they encode semantic identity. Afterwards, trajectories converge toward individual training samples at the \textbf{collapse time} $t_C$ (phase II-III), which quantifies instance-level attraction, such as nearest-neighbor consistency. For $t < t_C$, diffusion paths are drawn to a single point and lose their generalization capability, indicating memorization.  Empirically, as shown in \Figref{fig:stage_analysis_existing}\footnote{We clarify that Figure~\ref{fig:stage_analysis_existing} is a slight modification of figures from \citep{raya2023spontaneous} and \citep{georgiev2023journey}, adjusted to align the time steps with ours.}, \citet{georgiev2023journey}  observed that the likelihood that the noisy data $\rvx_t$ is classified into a certain class in speciation time $t_S$. Also, \citet{raya2023spontaneous} observed that data converges into a stable point, indicating collapse time $t_C$.

To investigate data unlearning in the aforementioned view, we will exploit the use of the transition points in data unlearning. We conduct a \textit{toy experiment} with a shallow diffusion model on a half-moon dataset, as visualized in \Figref{fig:toy_exp}. We only change the unlearning time steps while using the same data samples, with the loss function of \Eqref{eq:ga}. Deleting only the early time steps (phase III) does not help forget data samples; instead, it leads to quality degradation in real samples. Unlearning the middle time steps (phase II) is most aligned with the instance-level attraction before reaching the collapse time $t_C$.
For stronger forgetting of semantic conditioning, targeting the late time steps (phase I) preceding speciation at $t_S$ is effective.

\subsubsection{Time-selective Diffusion Forgetting} 
Based on the previous analysis, we suggest a simple framework of selective time steps as 
\begin{equation}
\mathbb{E}_{\rvx_0\sim p_0}
\underbrace{\mathbb{E}_{t \sim P(t)}}_{\text{Time selection}}
\Bigl[
  w_t \| s_\theta(\rvx_t,t) - \nabla_{\rvx_t} \log q({\rvx_t}| \rvx_0) \|^{2}
\Bigr].
\label{eq:ours_time}
\end{equation}
Here, $P(t)$ is a time step weighting.
In detail, we utilize the time scales that give a higher probability to a specific interval $t_1 \leq t \leq t_2$ for the forget data samples as follows:
\begin{equation}
P(t) =
\begin{cases}
{1-k}{/(t_2 - t_1)}, & t_1 \le t \le t_2, \\
{k}/({T - (t_2 - t_1))}, & \text{otherwise},
\end{cases}
\label{eq:time}
\end{equation}
where $0 \le k \le 1$ is the suppression intensity and $t_1,t_2 \in (0,T]$ with $t_1 < t_2$ and we use $k=0$ for simplicity.
As explained in \Secref{sec:scenario}, the evaluation scenarios for unconditional and conditional settings differ significantly. Therefore, we investigate appropriate time steps $P(t)$ in \Eqref{eq:time}.


\paragraph{Unconditional Unlearning.}
The gradient norm $\|\nabla_\theta\mathcal{L}(\rvx;\theta)\|_2^2$ reflects the magnitude of the update required to forget a sample $\rvx$ \citep{paul2021deep,pal2025llm}, allowing us to identify which stages are most affected by the unlearning. In \Figref{fig:distribution_time_b}, to compare the actual changes after standard unlearning with uniform time steps in the image-level unlearning with CelebA-HQ, {we unlearn a pre-trained model with SISS}. The gradient norm increases significantly after unlearning, with the gap being most pronounced in the middle time steps (Phase II). The gradient gap indicates that unlearning induces a substantial shift in the model's parameter space during Phase II.

Furthermore, as shown in \Figref{fig:distribution_time_a}, when we calculate the similarity on DINOv3 \citep{simeoni2025dinov3} of the noisy data $\rvx_t$ between other clean data samples, the similarity towards the original data $\rvx_0$ is drastically increased in the refinement stage, compared to other nearest samples. Thus, to prevent the diffusion trajectory from heading toward memorization, we need to control the preceding memorization at $t_C$. Note that the evaluation scheme also compares the difference between the original and reconstructed images from intermediate time steps, such as $t=250, 500$.

\begin{figure}[!t]
    \centering \vspace{-2mm}
    \begin{minipage}{0.45\textwidth}
        \centering
        \begin{subfigure}[b]{0.48\linewidth}
            \centering
            \includegraphics[width=\linewidth]{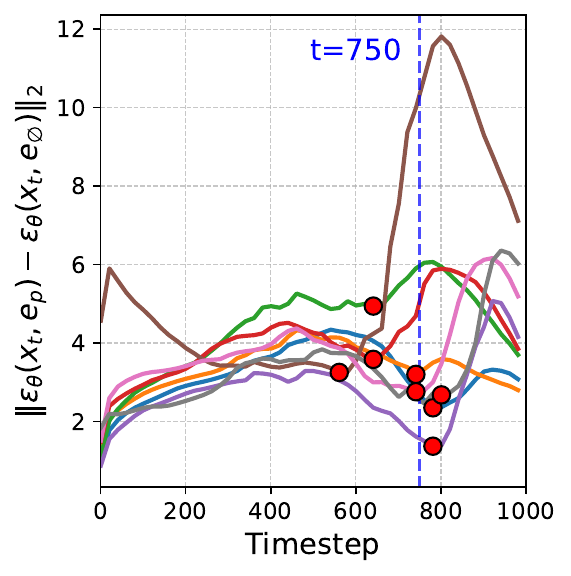}
            \caption{Guidance norm}
            \label{fig:adaptive_a}
        \end{subfigure}
        \hfill
        \begin{subfigure}[b]{0.48\linewidth}
            \centering
            \includegraphics[width=\linewidth]{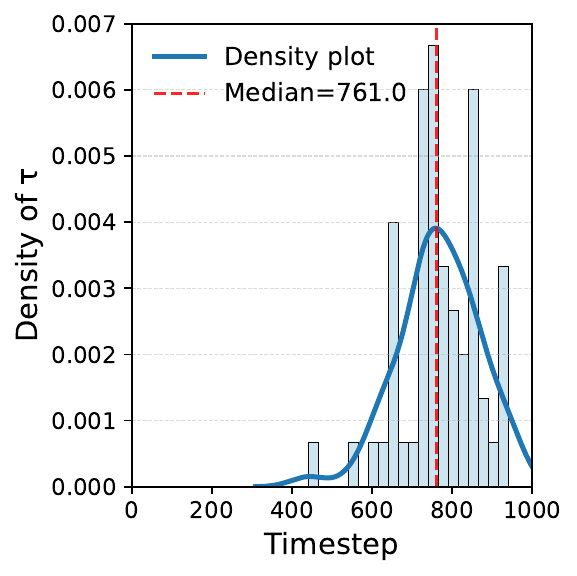}
            \caption{Distribution of $\tau$}
            \label{fig:adaptive_b}
        \end{subfigure}
        \caption{Adaptive conditional memorization.}
        \label{fig:adaptive_combined}
    \end{minipage}
    \hfill
    \begin{minipage}{0.53\textwidth}
        \centering
        \begin{subfigure}[t]{0.48\linewidth}
            \centering
            \includegraphics[width=\linewidth]{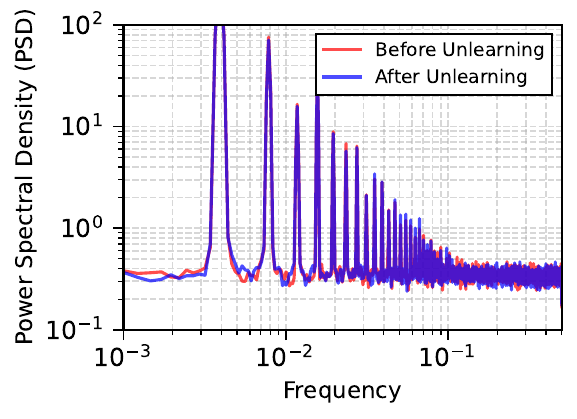}\vspace{-2mm}
            \caption{Retain samples}
            \label{fig:Retain}
        \end{subfigure}
        \hfill
        \begin{subfigure}[t]{0.48\linewidth}
            \centering
            \includegraphics[width=\linewidth]{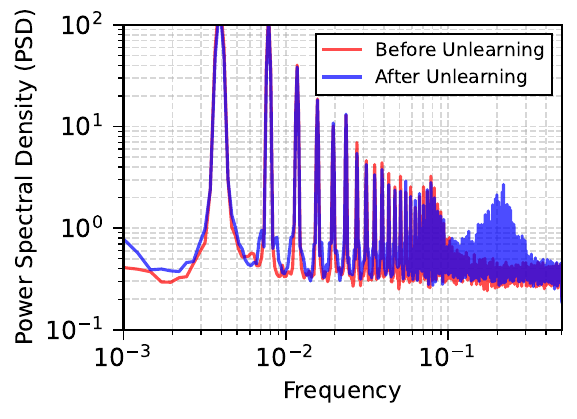}\vspace{-2mm}
            \caption{Over-forgotten samples}
            \label{fig:Collapsed}
        \end{subfigure}
        \caption{Power spectral density before and after unlearning. Over-forgotten samples differ in high frequencies, while retain samples remain unchanged.}        
        \label{fig:psa}
    \end{minipage}
    \vspace{-4mm}
\end{figure}

\paragraph{Conditional Unlearning.} 
In conditional memorization, the effect of conditional guidance, such as class or text embeddings, acts more strongly than its unconditional counterpart \citep{zhai2024membership}. Therefore, calculating the guidance magnitude $\|\epsilon_\theta(\rvx_t, e_p) - \epsilon_\theta(\rvx_t, e_\emptyset)\|_2^2$ is key for conditional memorization \cite{wen2024detecting, jeon2025understanding}. In \Figref{fig:adaptive_a}, we measured the trajectory of conditional guidance with memorized prompts in Stable Diffusion v1.4 \cite{rombach2022high}, with detailed settings provided in \Secref{sec:setup}. We can observe a peak in guidance norms in phase I before speciation begins \cite{jeon2025understanding,jain2025classifier}.

Based on the experiments, we can hypothesize that controlling the diffusion trajectory for $t>t_S$ before specification begins is important for the conditional case.  Similarly, training-free methods observed that guiding only late time steps $t>t_S$ is sufficient to steer the generation away from harmful concepts. For example, \citet{kim2025training} applied guidance within $t>780$, while \citet{kirchhof2025shielded} noted that their repellency terms are significant in late time steps.

In summary, $t_C<t<t_S$ (phase II) is critical for controlling unconditional behavior, while $t>t_S$ (phase I) matters for conditional generation, where conditional guidance dominates the diffusion dynamics after speciation, and the trajectory should be regulated before each critical time step begins.


\definecolor{myblue}{HTML}{2E86C1}
\definecolor{myred}{HTML}{E74C3C}

\begin{wraptable}{r}{0.55\textwidth}
    \centering
    \vspace{-4mm}
    \caption{Summary of diffusion transition points $t_S$ and $t_C$ ($T=1000$) across existing studies. $^\dagger$ denotes intervals. }
    \label{tab:time_selection}
    \small
    \resizebox{\linewidth}{!}{
        \begin{tabular}{l l c c}
            \toprule
            \textbf{Reference} & \textbf{Basis of Analysis} & $\mathbf{t_S}$ \textbf{(phase I-II)} & $\mathbf{t_C}$ \textbf{(phase II-III)} \\
            \midrule
            \citet{Biroli2024-pa}$^\dagger$ & Spectral / Entropy & $[500,800]$ & $[100,250]$ \\
            \citet{choi2022perception} & SNR ($10^{-2}, 10^0$) & $675$ & $259$ \\
            \citet{yang2024denoising} & Structure Pruning & $750$ & $250$ \\
            \citet{kim2025training} & Conditional Sampling & $780$ & -- \\
            \citet{jain2025classifier}$^\dagger$ & Conditional Sampling & $[600,800]$ & -- \\
            \bottomrule
        \end{tabular}
    }
    \vspace{-4mm}
\end{wraptable}
\paragraph{Optimal Selection of Transition Time $t_S$ and $t_C$. }
Finding the optimal $t_S,t_C$ is an open question in diffusion models. 
We summarize previous researches in Table \ref{tab:time_selection}, where $t_S$ and $t_C$ are approximately similar to the transition points $(t_C,t_S)=(250,750)$ used in \Figref{fig:toy_exp}.


Since the guideline in Table \ref{tab:time_selection} is not sample-wise, we further consider an \textbf{adaptive method on conditional unlearning} based on \citet{jain2025classifier}, analogue to the speciation time $t_S$: (i) the guidance magnitude $\|\epsilon_\theta(\rvx_t, e_p) - \epsilon_\theta(\rvx_t, e_\emptyset)\|_2^2$ is computed, and (ii) set the transition point $\tau$ as the first local minimum observed in sampling, detailed in Algorithm \ref{alg:adaptive_unlearning} (\Appref{app:exp}). As the $\tau$ searching method is prompt adaptive, we can consider the transition $\tau$ as $t_S$. As shown in \Figref{fig:adaptive_b}, although the optimal $\tau$ varies across prompts, the median of $\tau$ is $761.1$, similar to the fixed $t_S=750$.

\subsection{Not Every Frequency Mode Needs to Be Forgotten}

\subsubsection{Motivation}
From a time-selective perspective, Phase I ($t > t_S$) and Phase II ($t_C < t < t_S$) represent the primary stages where semantic- and individual-level memorization are established, respectively. 
Maintaining time steps $t < t_C$ (Phase III) is essential for quality preservation because this phase only refines the quality of the generated sample determined in earlier stages. This motivates the use of low-pass filters in diffusion models to avoid forgetting fine-grained and high-frequency details \cite{lian2025unveiling,choi2025enhancing}.

We evaluate the impact of unlearning in the frequency domain via the power spectral density in \Figref{fig:psa}. For the retain set, the high-frequency distribution is preserved post-unlearning. Conversely, the forget set exhibits \textit{unwanted high-frequency components} that risk degrading sample quality. 



\subsubsection{Frequency-selective Diffusion Forgetting}
Thus, to preserve fine-grained image quality and make less perturbation in the manifold during unlearning, we add a low-pass filter to remove the high-frequency components of the forget set as
\begin{equation}
\mathbb{E}_{\mathbf{x}_0\sim p_0}\mathbb{E}_t\Bigl[w_t\bigl\|\underbrace{\mathcal{F}\bigl(s_\theta(\mathbf{x}_t,t)-\nabla{\mathbf{x}_t}\log q(\mathbf{x}_t\mid\mathbf{x}_0)\bigr)}_{\text{Low-pass filter}}\bigr\|^2\Bigr]
\label{eq:ours}
\end{equation}
Notice that the denoiser input remains the noisy natural sample $\mathbf{x}_t$ during unlearning; \textbf{only the loss residual is projected to low frequencies}. Hence, the input distribution during unlearning and inference is identical, and only the gradient signal used for unlearning is projected.
In practice, we employ the discrete Fast Fourier Transform (FFT) as follows:
\begin{equation}
\mathcal{T}(u,v) = \text{FFT}(\rvx) = \sum_{x=1}^{H}\sum_{y=1}^{W} \rvx(x,y) e^{-j2\pi\left(\frac{u}{H}x + \frac{v}{W}y\right)},
\end{equation}
where $\rvx(x,y)$ is the pixel intensity at position $(x,y)$, $\mathcal{T}(u,v)$ is the complex coefficient at frequency $(u,v)$, and $e$ and $j$ represent Euler’s constant and the imaginary unit, respectively. 
To remove only the high-frequency terms, we apply a masking function to the FFT results and then reconstruct the image using the inverse FFT with radius $r$ as
$\mathcal{F}(\rvx_{i,t})
= \text{IFFT} \left(\text{FFT}(\rvx_{i,t}) \odot \beta_{i,t}(r)\right),$
where $\beta_{i,t}(r) = s \text{ if the radius } r > r_t, 1 \text{ otherwise}$. $r_t$ is a radius cutoff frequency threshold. $0 \leq s \leq 1$ is a high frequency weight, where we set $s=0$ to exclude high frequency in unlearning.

\subsection{Theoretical Justification}

Data unlearning must remove the influence of specific samples while preserving the overall data manifold. To this end, we analyze the enhanced forgetting within the selected region and the minimized distortion in the non-selected region, showing the strength in the forgetting–preservation trade-off.

\noindent\textbf{Setup.} 
Let $\mathbb{P}_{\mathrm{orig}}$ and $\mathbb{P}_m$ ($m \in \{\mathrm{full}, \mathrm{sel}\}$) be the path measures of the original and unlearned diffusion models. We denote the score perturbation as $\Delta s_m(x_t, t) := s_{\theta_m}(x_t, t) - s_{\theta_{\mathrm{orig}}}(x_t, t)$. 
By Girsanov's theorem~\cite{liptser1977statistics}, the KL divergence between the original and unlearned path measures is 
\begin{equation}
     D_{\mathrm{KL}}(\mathbb{P}_{\mathrm{orig}} \,\|\, \mathbb{P}_m) = \frac{1}{2}\int_0^T g(t)^2 \mathbb{E}_{x_t} \left[ \|\Delta s_m(x_t, t)\|_2^2 \right] dt. 
\end{equation}
To evaluate unlearning efficacy, we decompose this total unlearning-induced distortion into a selected target window $\mathcal{S}$ (e.g., selected time-steps or frequencies) and its complement $\mathcal{S}^c$. Let $D_{\mathrm{KL}}^{\mathcal{W}}(\mathbb{P}_{\mathrm{orig}} \,\|\, \mathbb{P}_m) := \frac{1}{2}\int_{\mathcal{W}} g(t)^2 \mathbb{E}_{x_t} [\|\Delta s_m\|_2^2] dt$ denote the regional divergence over a window $\mathcal{W}$. Thus, the total divergence is
$ D_{\mathrm{KL}}(\mathbb{P}_{\mathrm{orig}} \,\|\, \mathbb{P}_m) = D_{\mathrm{KL}}^{\mathcal{S}}(\mathbb{P}_{\mathrm{orig}} \,\|\, \mathbb{P}_m) + D_{\mathrm{KL}}^{\mathcal{S}^c}(\mathbb{P}_{\mathrm{orig}} \,\|\, \mathbb{P}_m). $
For $\beta > 0$, we define the  forgetting--preservation utility as:
\begin{equation}\label{eq:trade_off}
    \mathcal{J}_{\beta}(m) := D_{\mathrm{KL}}^{\mathcal{S}}(\mathbb{P}_{\mathrm{orig}} \,\|\, \mathbb{P}_m) - \beta D_{\mathrm{KL}}^{\mathcal{S}^c}(\mathbb{P}_{\mathrm{orig}} \,\|\, \mathbb{P}_m).
\end{equation}
Note that $D_{\mathrm{KL}}^{\mathcal{S}}(\mathbb{P}_{\mathrm{orig}} \,\|\, \mathbb{P}_m)$ is a proxy for the \textit{intended forgetting} and, $D_{\mathrm{KL}}^{\mathcal{S}^c}(\mathbb{P}_{\mathrm{orig}} \,\|\, \mathbb{P}_m)$ is a proxy for \textit{collateral distortion} without intend to unlearn, detailed in Appendix~\ref{app:proof}.

\noindent\textbf{Assumptions.} 
(i) \textit{Finite divergence}: $D_{\mathrm{KL}}(\mathbb{P}_{\mathrm{orig}} \,\|\, \mathbb{P}_m) < \infty$.  (ii) \textit{Focused updates in $\mathcal{S}$}: Due to the frequent update on the selected region under the same training budget, selective unlearning amplifies changes in the target region ($\|\Delta s_{\mathrm{sel}}\|_2 \ge \mu \|\Delta s_{\mathrm{full}}\|_2$ for $\mu \ge 1$ in $\mathcal{S}$) and weaker changes in $\mathcal{S}^c$ ($\|\Delta s_{\mathrm{sel}}\|_2 \le \eta \|\Delta s_{\mathrm{full}}\|_2$ for $\eta \in [0,1]$ in $\mathcal{S}^c$). See Appendix~\ref{app:proof} for empirical validation and proofs.

\begin{proposition}
\label{prop:3.1}
Under the assumptions above, for every  $\beta>0$, and $\lambda := \frac{D_{\mathrm{KL}}^{\mathcal{S}^c}(\mathbb{P}_{\mathrm{orig}} \,\|\, \mathbb{P}_{\mathrm{full}})}{D_{\mathrm{KL}}(\mathbb{P}_{\mathrm{orig}} \,\|\, \mathbb{P}_{\mathrm{full}})} \in [0,1]$,  the trade-off utility gap of $\mathcal{J}_{\beta}(\mathrm{sel})$ and $\mathcal{J}_{\beta}(\mathrm{full)}$ satisfies:
\begin{equation}
    \mathcal{J}_{\beta}(\mathrm{sel}) - \mathcal{J}_{\beta}(\mathrm{full}) \ge \left[ (\mu^2-1)(1-\lambda) + \beta(1-\eta^2)\lambda \right] D_{\mathrm{KL}}(\mathbb{P}_{\mathrm{orig}} \,\|\, \mathbb{P}_{\mathrm{full}}),
\end{equation}  
where $\lambda$ indicates the proportion of collateral distortion compared to the full-unlearning.
\end{proposition}
\noindent\textbf{Remark 1.} \textit{The bracketed term $(\mu^2-1)(1-\lambda) + \beta(1-\eta^2)\lambda$ is always non-negative under our assumptions ($\mu \ge 1, \eta \le 1, \beta>0, \lambda \in [0,1]$). Then, the improvement in utility is strictly positive.}

\noindent\textbf{Remark 2.} \textit{The utility gap $\mathcal{J}_{\beta}(\mathrm{sel}) - \mathcal{J}_{\beta}(\mathrm{full})\geq0$ indicates our selective mechanism focuses score updates on the selected forget-region while mitigating unwanted distortion in non-selected parts.}

\section{Experiments}
\label{sec:exp}

\subsection{Experimental Setups}
\label{sec:setup}
\paragraph{Datasets and Baselines.} We follow the settings of \citet{alberti2025data}. For CelebA-HQ \citep{liu2015faceattributes}, we evaluate the deletion of individual samples at the image level in unconditional image generation. For Stable Diffusion v1.4 \citep{rombach2022high}, we evaluated the memorization of the LAION dataset corresponding to each prompt in text-to-image generation. 
Further experimental details are in the Appendix~\ref{app:exp}.

\paragraph{Hyperparameters.}
We use both time selection \peqref{eq:ours_time} and frequency selection \peqref{eq:ours} for experiments. Although we acknowledge that the choice of thresholds might vary, we fix $(t_C,t_S)=(250,750)$ based on \Tabref{tab:time_selection}. Thus, we mainly use $(t_1,t_2)=(250,750)$ for the unconditional case and $(t_1,t_2)=(750,1000)$ for the conditional case.
Based on the observation in \Figref{fig:psa}, as our goal is to filter out the high-frequency, we tune $r_t \in [0.1, 0.15]$, and unless otherwise specified, we set $r_t=0.15$. 
For CelebA-HQ and Stable Diffusion from the Hugging Face Diffusers, we fine-tune the model with Adam with a batch size of 64 and 16 with a learning rate of $5\cdot10^{-6}$ and $10^{-5}$, respectively. 

\paragraph{Evaluation Metrics.}
The traditional evaluation of unlearning methods is two-fold: the quality of retained data samples, measured by Frechet Inception Distance (FID) \citep{heusel2017gans}, and the dissimilarity of forgotten samples from the originals, measured by SSCD \citep{pizzi2022self}. However, SSCD ignores the quality of the generated samples, e.g., a blurry image achieves a very low SSCD, though it is not a meaningful outcome. To address this, we introduce a \textbf{normalized SSCD score} that considers the normalized distance between the generated image $\hat\rvx_0(\rvx_t,\theta)$ and the original image $\rvx_0$. Motivated by adversarial attacks \citep{madry2018towards}, we project the difference onto $\ell_2$ ball with radius $\rho$ as
\begin{equation}    
\small
\mathrm{SSCD}^\text{norm} = \mathrm{SSCD}(\rvx_0,\rvx_0 + \rho \frac{\hat\rvx_0(\rvx_t,\theta) - \rvx_0}{\|\hat\rvx_0(\rvx_t,\theta) - \rvx_0\|_2^2+\varepsilon}).
\label{eq:sscd_norm}
\end{equation}
Here, $\varepsilon$ is a small constant to avoid division by zero, and we use $\rho=100$ with a normalized image with 256x256 resolution. As a quality metric, we compute sample-wise aesthetic scores using the LAION-Aesthetic V2  predictor\footnote{https://github.com/christophschuhmann/improved-aesthetic-predictor}.
For Stable Diffusion, we evaluate generation quality using CLIP-IQA \citep{wang2023exploring} and measure unlearning performance through the unlearning success rate, defined as the proportion of cases where all 16 memorized samples are removed. 
We tested with the \textbf{five runs} and report their mean. Variance is comparably small, shown in \Tabref{tab:variance}.

\begin{figure*}[!t]
    \centering
    \includegraphics[width=0.85\textwidth]{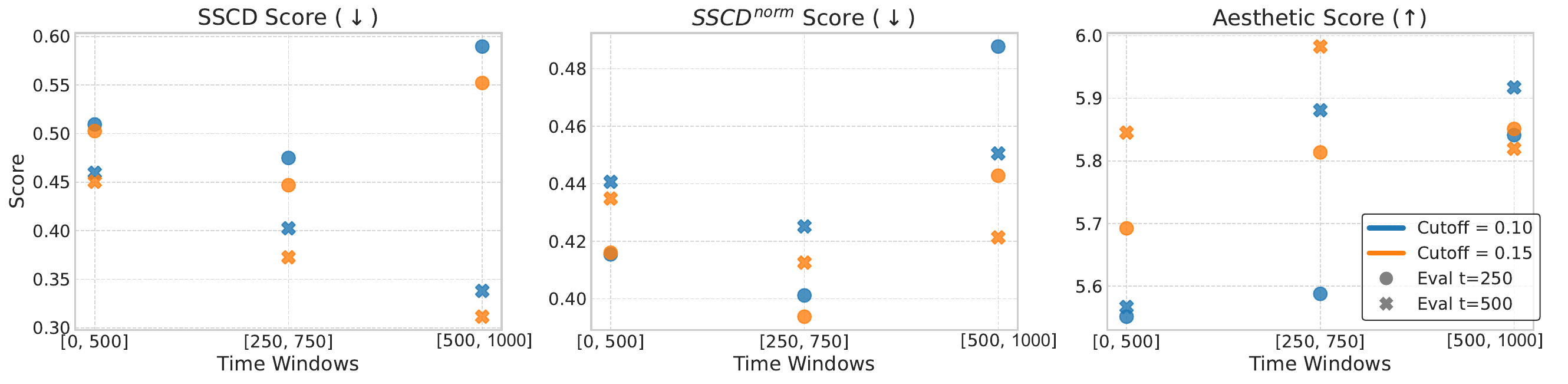}
    \caption{Ablation study with different diffusion time steps and cutoff for low-pass filter.}
    \label{fig:metric_grid_final}
    \vspace{-4mm}
\end{figure*}

\begin{table*}[!t]
\centering
\caption{Comparison of unconditional unlearning. We show the baseline scores, the scores with our method applied (+ Ours), and the resulting relative gain (\%) in a separate row. For the 'Gain (\%)' row, a higher value always indicates better results. Positive gains are in \textcolor{myblue}{blue}, and degradations in \textcolor{myred}{red}.}
\label{tab:main}
\resizebox{0.8\textwidth}{!}{%
\begin{tabular}{l l c ccc ccc}
\toprule
\multicolumn{2}{l}{\multirow{2}{*}{\textbf{Method}}} & & \multicolumn{3}{c}{Denoising from $t=250$} & \multicolumn{3}{c}{Denoising from $t=500$} \\
\cmidrule(lr){4-6} \cmidrule(lr){7-9}
& & FID-10K $\downarrow$ & SSCD $\downarrow$ & SSCD$^{\text{norm}}$ $\downarrow$ & Aesthetic $\uparrow$ & SSCD $\downarrow$ & SSCD$^{\text{norm}}$ $\downarrow$ & Aesthetic $\uparrow$ \\
\midrule
\multicolumn{2}{l}{Pre-trained} & 30.3 & 1.257 & - & - & - & - & - \\
\multicolumn{2}{l}{Naive deletion} & 19.61 & 0.874 & 0.607 & 6.118 & 0.726 & 0.641 & 6.077 \\
\multicolumn{2}{l}{EraseDiff} & 117.81 & 0.133 & 0.551 & 3.702 & 0.096 & 0.517 & 4.359 \\
\midrule
\multirow{3}{*}{{GA}} & Base & 359.79 & 0.131 & 0.548 & 3.079 & 0.002 & 0.783 & 3.470 \\
& + Ours & 375.18 & 0.113 & 0.499 & 3.699 & 0.062 & 0.600 & 4.577 \\ \cmidrule{2-9}
& \textit{Gain (\%) } & \textcolor{myred}{-4.28} & \textcolor{myblue}{+13.74} & \textcolor{myblue}{+8.94} & \textcolor{myblue}{+20.14} & \textcolor{myred}{-3$\cdot10^{3}$} & \textcolor{myblue}{+23.37} & \textcolor{myblue}{+31.90} \\
\midrule
\multirow{3}{*}{{SISS}} & Base & 23.42 & 0.336 & 0.430 & 4.094 & 0.299 & 0.501 & 4.845 \\
& + Ours & 23.65 & 0.345 & 0.349 & 5.520 & 0.282 & 0.399 & 6.095 \\ \cmidrule{2-9}
& \textit{Gain (\%) } & \textcolor{myred}{-0.98} & \textcolor{myred}{-2.68} & \textcolor{myblue}{+18.84} & \textcolor{myblue}{+34.83} & \textcolor{myblue}{+5.69} & \textcolor{myblue}{+20.36} & \textcolor{myblue}{+25.80} \\
\midrule
\multirow{3}{*}{{DPO}} & Base & 20.58 & 0.369 & 0.404 & 5.128 & 0.313 & 0.459 & 5.058 \\
& + Ours & 21.15 & 0.344 & 0.332 & 5.614 & 0.292 & 0.393 & 6.138 \\ \cmidrule{2-9}
& \textit{Gain (\%) } & \textcolor{myred}{-2.77} & \textcolor{myblue}{+6.78} & \textcolor{myblue}{+17.82} & \textcolor{myblue}{+9.48} & \textcolor{myblue}{+6.71} & \textcolor{myblue}{+14.38} & \textcolor{myblue}{+21.35} \\
\midrule
\multirow{3}{*}{{KTO}} & Base & 23.48 & 0.363 & 0.366 & 5.355 & 0.289 & 0.442 & 5.879 \\
& + Ours & 23.58 & 0.373 & 0.340 & 5.470 & 0.280 & 0.377 & 6.274 \\ \cmidrule{2-9}
& \textit{Gain (\%)} & \textcolor{myred}{-0.43} & \textcolor{myred}{-2.75} & \textcolor{myblue}{+7.10} & \textcolor{myblue}{+2.15} & \textcolor{myblue}{+3.11} & \textcolor{myblue}{+14.71} & \textcolor{myblue}{+6.72} \\
\bottomrule
\end{tabular}%
}
\vspace{-4mm}
\end{table*}

\subsection{Experimental Results}
\paragraph{Unconditional Data Unlearning.}
For CelebA-HQ, our objective is to delete six randomly sampled celebrity faces selected by \citet{alberti2025data} from pre-trained unconditional diffusion models. Unlearning is applied image-by-image in a continual setting. As GA erases all details in a continual manner, we initialize the model as a pretrained point after each individual deletion.
As explained in Section \ref{sec:scenario}, we first inject noise to certain time steps $t=250,500$ (only $t=250$ in \cite{alberti2025data}) and compare the denoised image using the unlearned model and the original image.

In \Figref{fig:metric_grid_final}, similar to the toy results in \Figref{fig:toy_exp}, we focus on the middle steps, which show stable forgetting and better quality preservation in image-level unconditional unlearning. For a low-pass filter, a cutoff threshold $r=0.15$ effectively maintains the quality. 
Thus, we apply our selective unlearning framework to various optimization methods (GA, SISS, DPO, and KTO) in \Tabref{tab:main}. To ensure that a positive value consistently indicates an improvement over metric $V$, the relative gain (\%) is calculated as  $(V_{\text{ours}} - V_{\text{base}}) / V_{\text{base}}$ for higher is better metric and $(V_{\text{base}} - V_{\text{ours}}) / V_{\text{base}}$ for lower is better metric.
The results indicate that both time and frequency selection achieve forgetting and preserve the quality of unlearned data examples. At $t=250$, it maintains the SSCD while improving SSCD$^{\text{norm}}$, and at $t=500$, it improves both metrics compared to base models. Since our selective method is optimized for the forget data, we observe a minor increase in FID for retained data. 

The comparison between SSCD and SSCD$_{\text{norm}}$ highlights the limitations of using raw similarity to evaluate unlearning. For instance, deletion-focused methods like EraseDiff and GA achieve low SSCD scores. While the results are well-aligned with target forgetting, it is merely an artifact of severe image quality degradation. Their high SSCD$_{\text{norm}}$ scores correctly reveal that their perturbation direction is ineffective for true unlearning. In contrast, our selective framework demonstrates a superior unlearning direction by achieving significant SSCD$_{\text{norm}}$ gains.

\begin{figure*}[!t]
    \centering
    \begin{subfigure}[b]{0.3\textwidth}
        \centering
        \includegraphics[width=\linewidth]{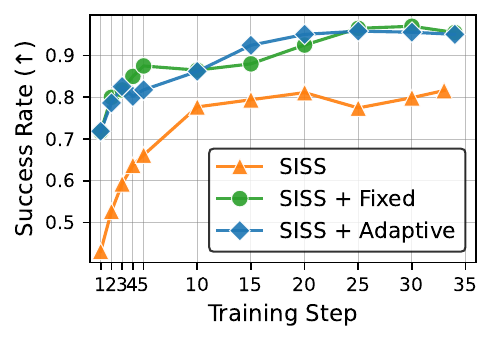}
        \caption{Unlearning Ratio (Full ASR)}
        \label{fig:unlearning_success_rate_1}
    \end{subfigure}%
    \begin{subfigure}[b]{0.3\textwidth}
        \centering
        \includegraphics[width=\linewidth]{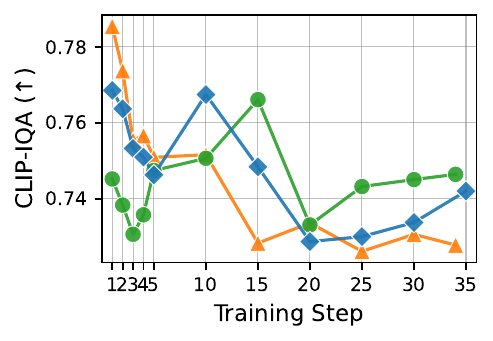}
        \caption{CLIP-IQA (Quality)}
        \label{fig:CLIP-IQA}
    \end{subfigure}%
    \begin{subfigure}[b]{0.38\textwidth}
        \centering
        \includegraphics[width=\linewidth]{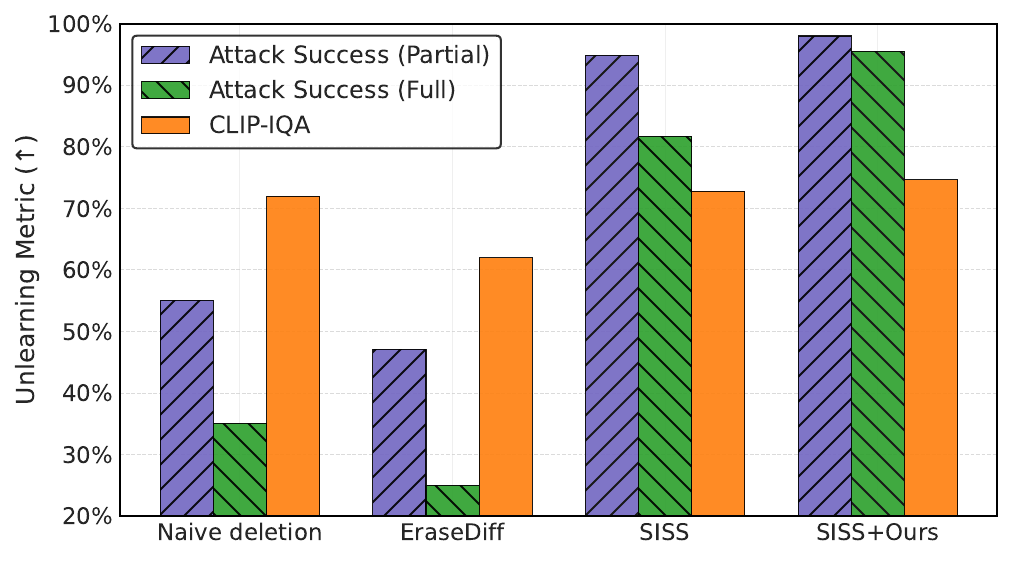}
        \caption{Overall Comparison}
        \label{fig:final_step_bar_sorted}
    \end{subfigure}%
    \caption{{ Unlearning attack success rate (ASR) and CLIP-IQA when unlearning each of the memorized 45 prompts. As shown in (a) and (b), selective unlearning shows a faster and higher unlearning rate.} The adaptive method does not show a big difference from the fixed one. As shown in (c), our method shows strong forgetting while preserving quality compared to existing unlearning methods. }
    \label{fig:unlearning_success_rate}
    \vspace{-5mm}
\end{figure*}

\paragraph{Conditional Text-to-Image Data Unlearning.} 
For Stable Diffusion, we evaluated to unlearn 45 memorized prompts selected by \citet{alberti2025data} from the LAION datasets. {Fully-memorized prompts consistently generate exact replicas of the training images, whereas partially-memorized prompts contain weaker memorization prompts, thus easier to unlearn.}
For conditional cases, we unlearn only time steps $t>t_S$, utilizing the condition on a text embedding of ``memorized prompt". As shown in \Figref{fig:unlearning_success_rate}, our method achieves a higher unlearning success rate and faster convergence than the baseline SISS throughout the training process. \Figref{fig:final_step_bar_sorted} further confirms that our approach yields superior unlearning (lower attack success) and better image quality (higher CLIP-IQA). 

\begin{figure*}[!t]
    \centering
    \begin{subfigure}[b]{0.45\textwidth}
        \centering
        \begin{minipage}[b]{0.3\linewidth}
            \includegraphics[width=\linewidth]{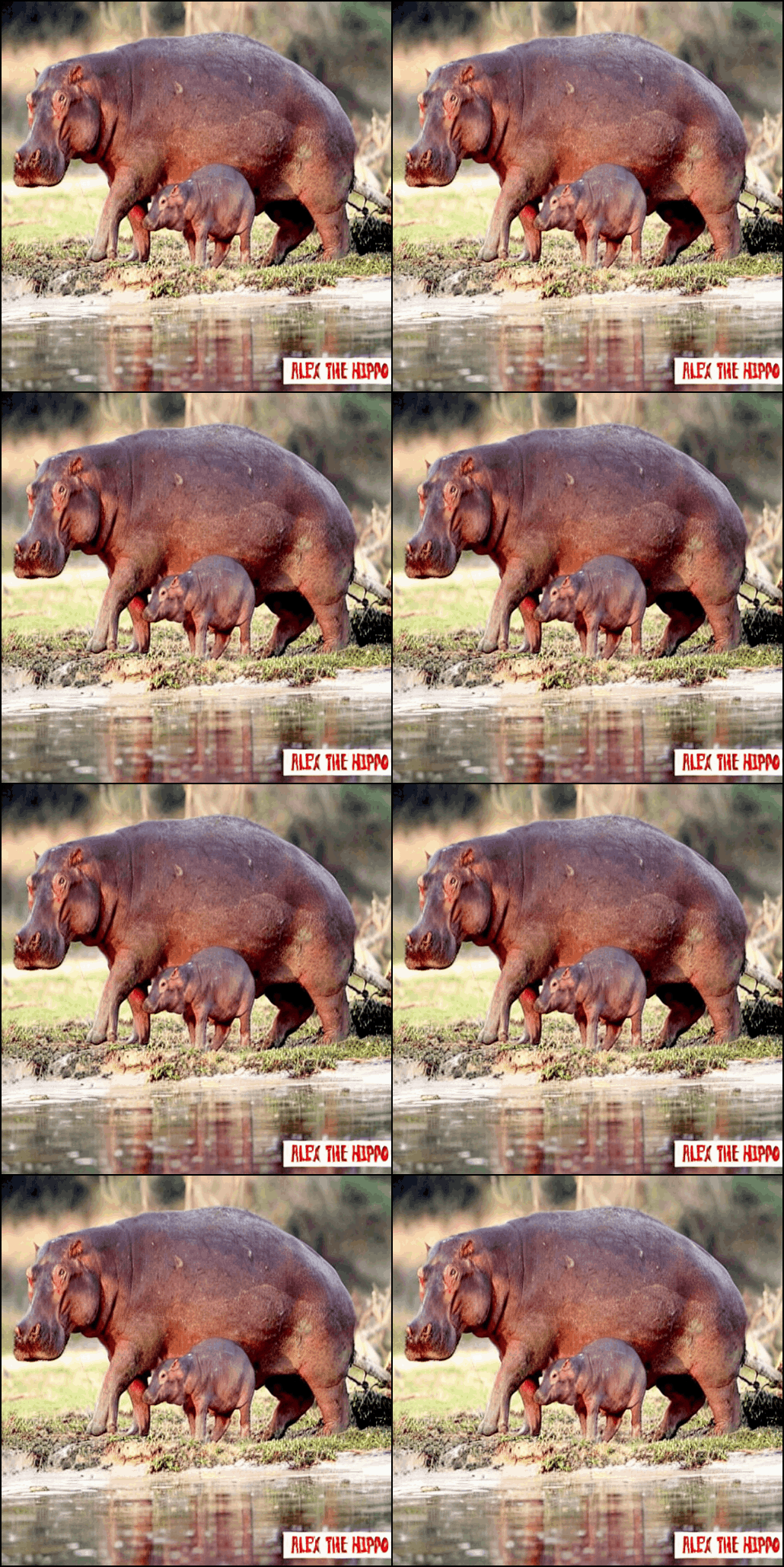}
        \end{minipage}%
        \hfill
        \begin{minipage}[b]{0.3\linewidth}
            \includegraphics[width=\linewidth]{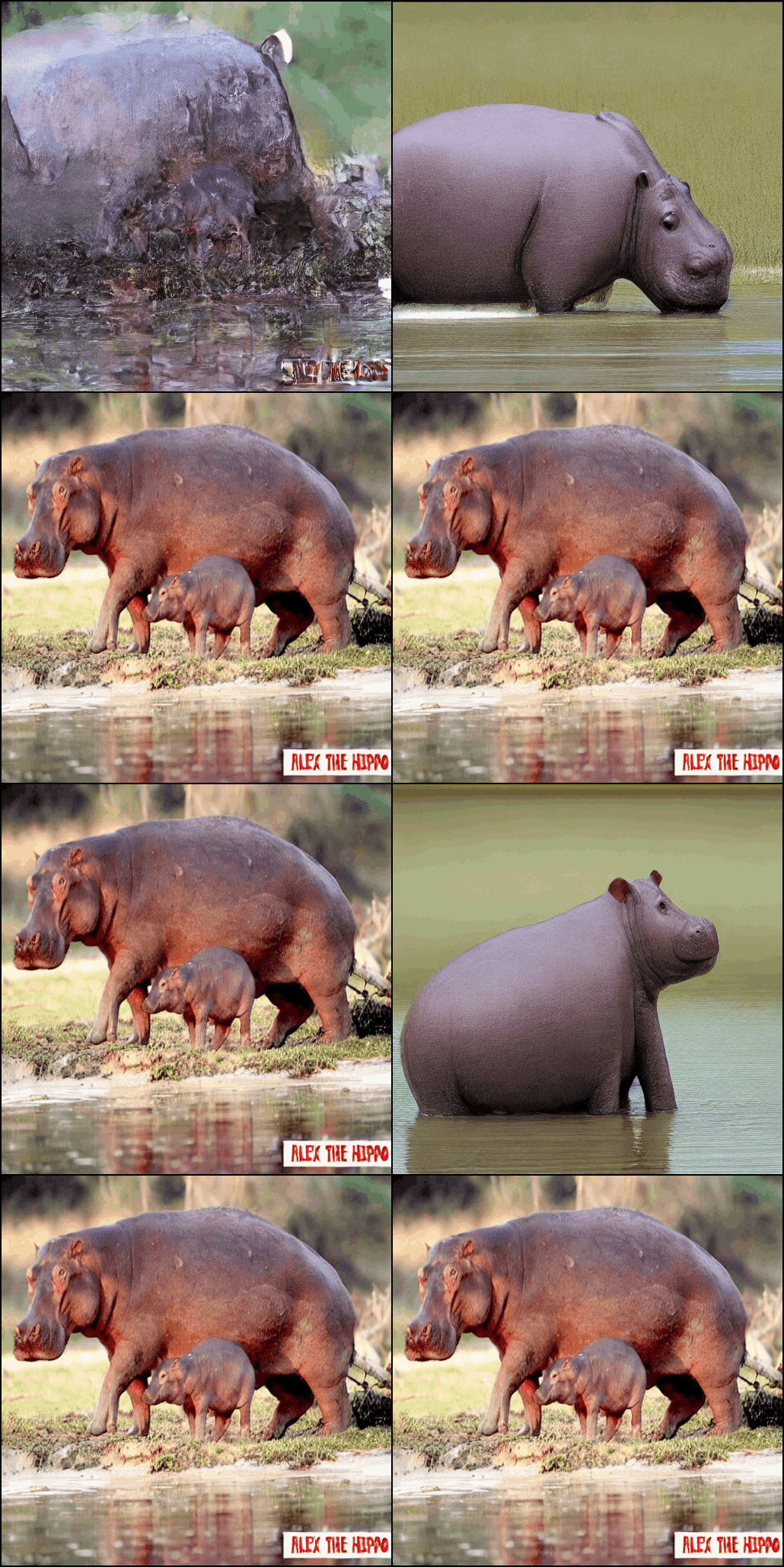}
        \end{minipage}%
        \hfill
        \begin{minipage}[b]{0.3\linewidth}
            \includegraphics[width=\linewidth]{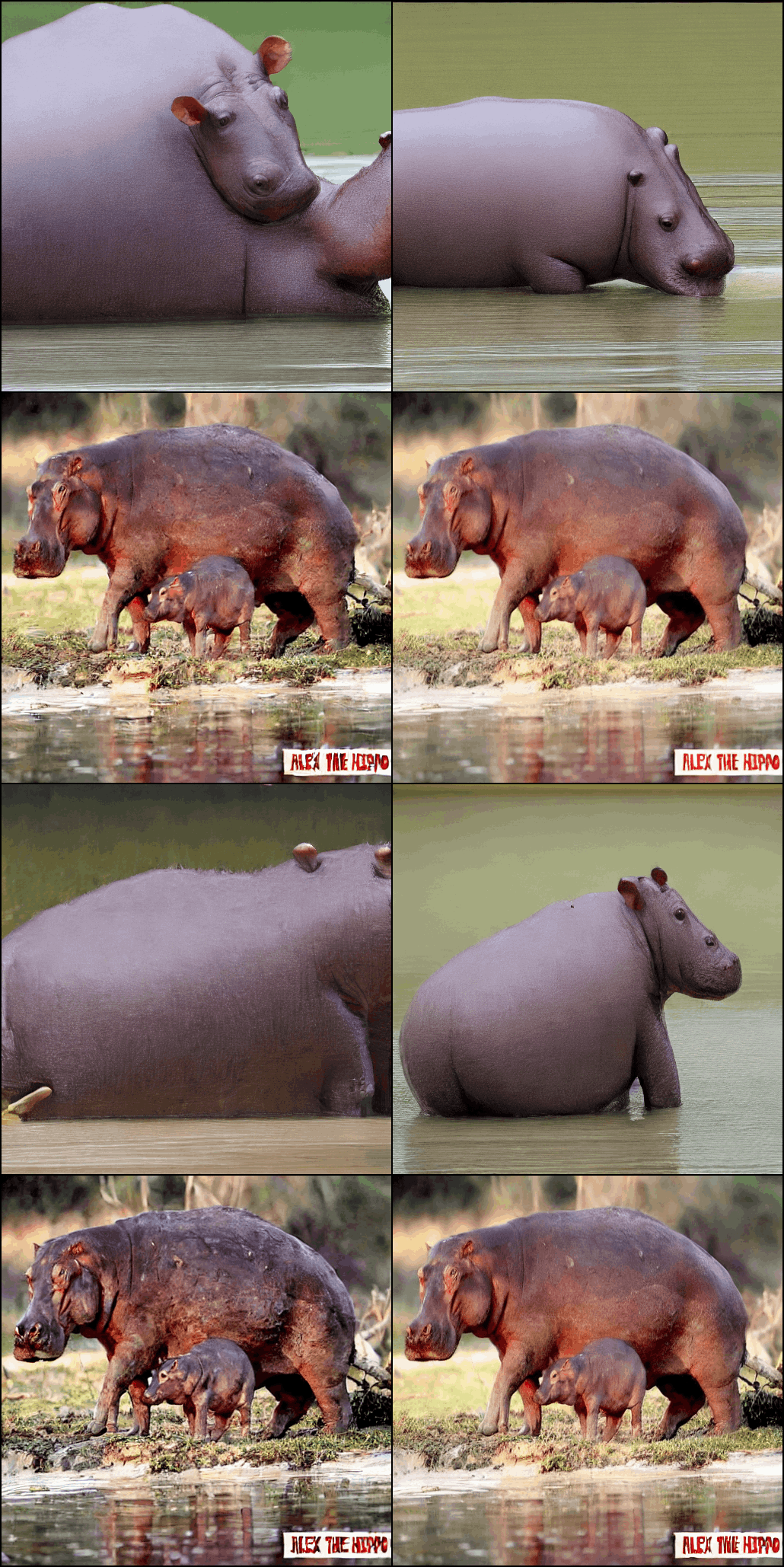}
        \end{minipage}
        \caption{Generated samples of SISS at early steps}
        \label{fig:hippo_siss}
    \end{subfigure}
    \hfill
    \begin{subfigure}[b]{0.45\textwidth}
        \centering
        \begin{minipage}[b]{0.3\linewidth}
            \includegraphics[width=\linewidth]{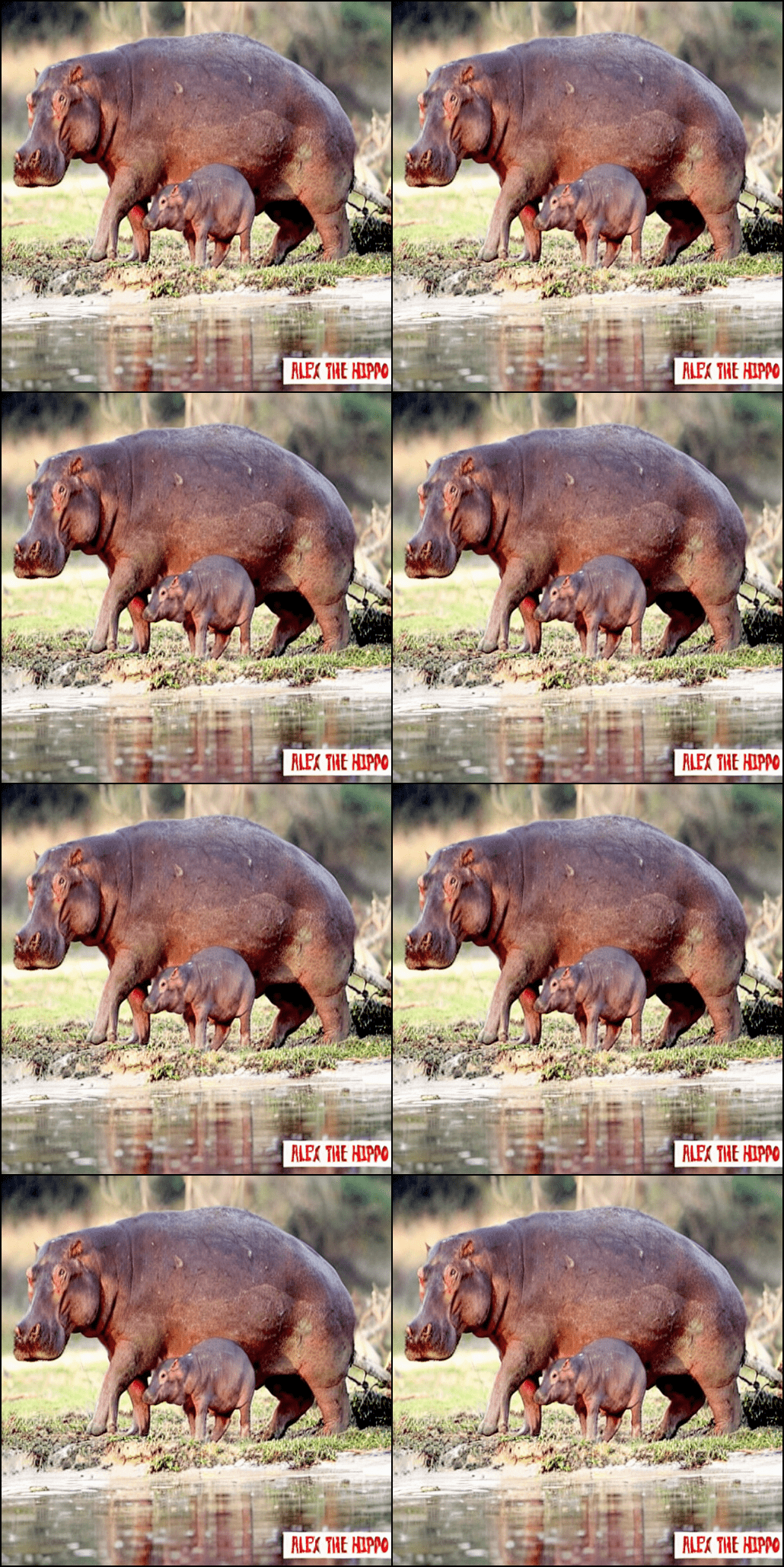}
        \end{minipage}%
        \hfill
        \begin{minipage}[b]{0.3\linewidth}
            \includegraphics[width=\linewidth]{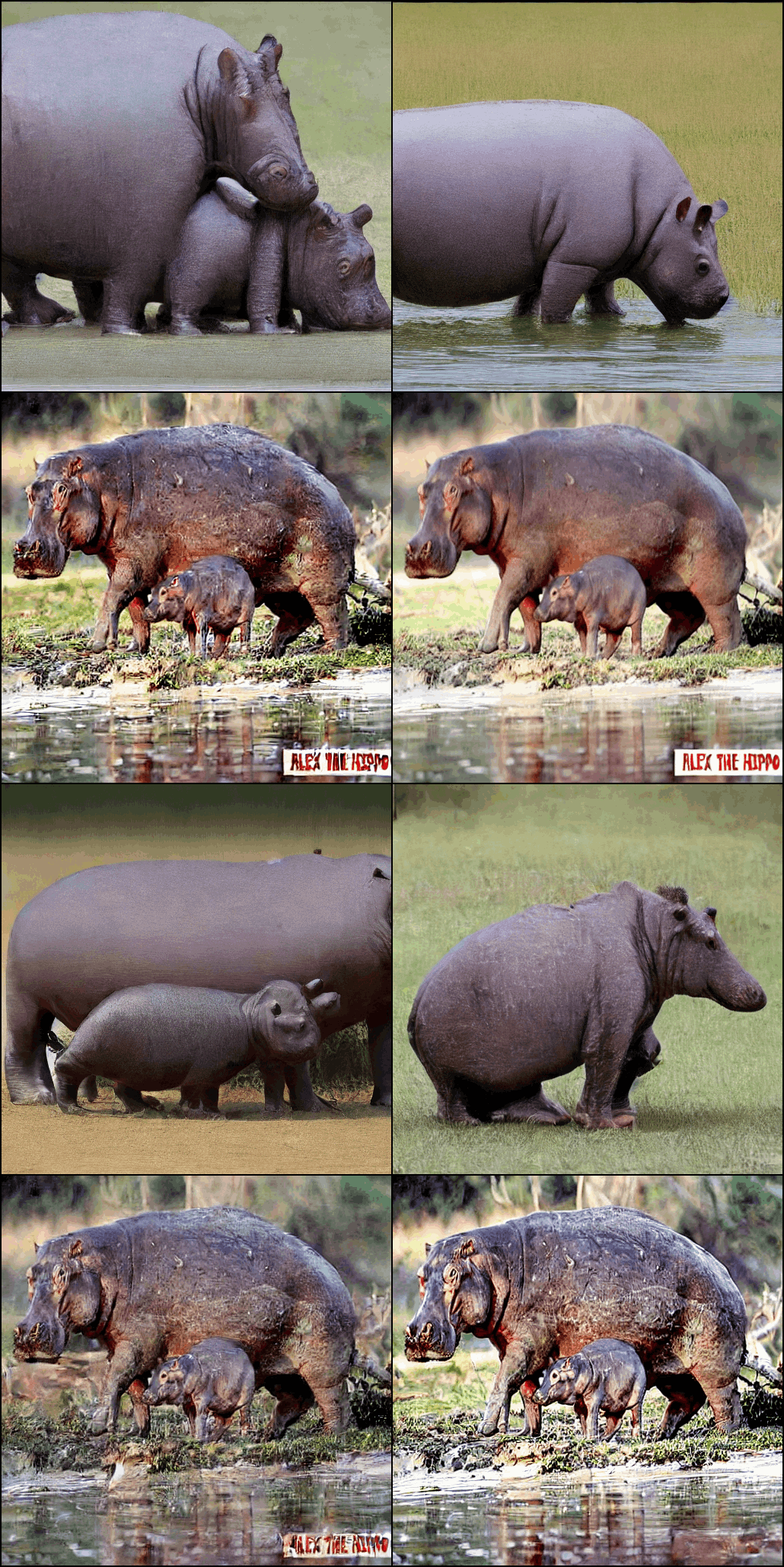}
        \end{minipage}%
        \hfill
        \begin{minipage}[b]{0.3\linewidth}
            \includegraphics[width=\linewidth]{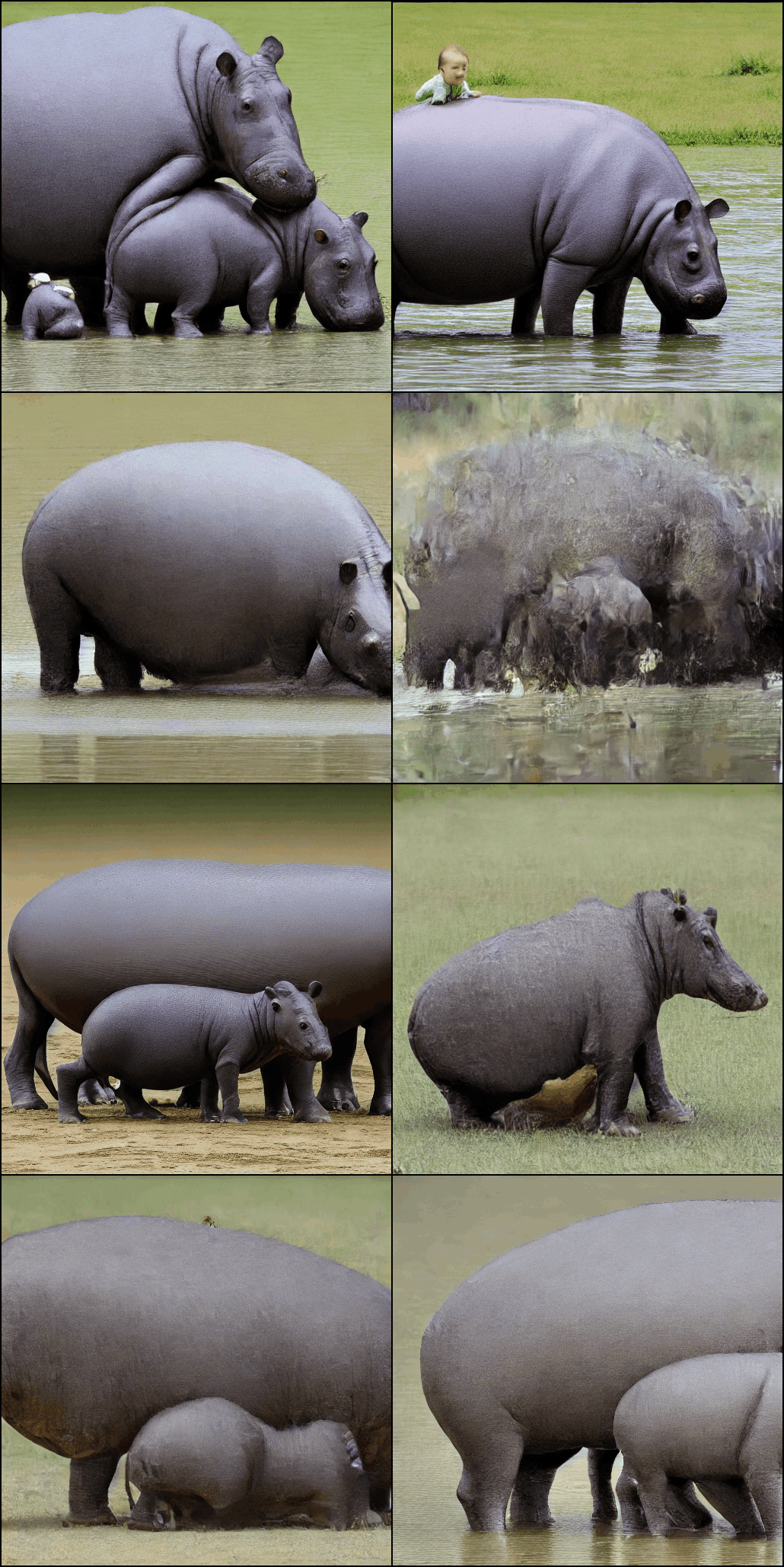}
        \end{minipage}
        \caption{Generated samples of SISS+Ours at early steps}
        \label{fig:hippo_ours}
    \end{subfigure}
    \caption{Visualization of generated images from the memorization with \textbf{fully-memorized prompt} ``Mothers influence on her young hippo'' at early unlearning steps $[0,1,2]$. The proposed method shows faster forgetting performance while maintaining quality.}
    \label{fig:hippo}
    \vspace{-3mm}
\end{figure*}

\subsection{Additional Experiments}
\label{sec:add_exp}
\textbf{Adaptive method.$ $} 
As illustrated in \Figref{fig:unlearning_success_rate} (a-b), the adaptive method in Algorithm \ref{alg:adaptive_unlearning} shows similar performance with the fixed window. However, the adaptive method introduces computational overhead in a pre-sampling step to determine $t_S=\tau$. Therefore, we mainly adopt the fixed interval, leaving the adaptive method for future exploration.  Furthermore, we also observed that frequency threshold scheduling can also improve the quality, as shown in \Tabref{tab:freq_schedule}.

\textbf{Computation.$ $ } 
As shown in \Tabref{tab:computation_cost} using Stable Diffusion, our method requires an additional computational cost for FFT of  5.99\% per epoch. Additionally, the adaptive method requires a sampling step before training begins. However, since our method converges significantly faster, the actual total training time can be reduced. Inference time remains unchanged for our methods.

\textbf{Membership inference attack (MIA).$ $}
We further conduct a membership inference attack (MIA)~\citep{duan2023diffusion,kong2024efficient} evaluation based on diffusion loss, testing on a member set, hold-out dataset, and unlearned set. As shown in~\Tabref{tab:mia}, existing methods suffer from over-forgetting, with relative errors reaching $\times 107.98$ (SecMI) and $\times  278.22$ (PIAN) on the unlearned set, far exceeding the hold-out baseline ($\times 1.32$ and $\times 1.12$). Our method substantially mitigates this collateral distortion ($\times 5.76$ and $\times 53.20$).

\textbf{Supplementary Experiments.$ $ } 
Refer to Appendix \ref{app:result} for supplementary experiments, including additional base analyses (Figures \ref{fig:app_nn}-\ref{fig:psa3}), evaluations on different backbones such as transformer-based architectures and Stable Diffusion v3 (Tables \ref{tab:main_ldm} and \ref{tab:sd3_results}), concept unlearning results (Figure \ref{fig:nudity}), LLM-based evaluations for anomaly detection (Table \ref{tab:llm_anomaly}), the effect of unlearning on nearby samples (Table \ref{tab:nn}), filter band and frequency cutoff scheduling ablations (Tables \ref{tab:filter_band} and \ref{tab:freq_schedule}), and statistical reliability across independent runs (Table \ref{tab:variance}). Refer to Appendix \ref{app:vis} for additional visualizations.

\begin{figure}[!ht]
\begin{minipage}[t]{0.41\textwidth}
    \centering
    \captionof{table}{Computational costs in seconds.}
    \label{tab:computation_cost}
    \small
    \resizebox{\linewidth}{!}{%
    \setlength{\tabcolsep}{5pt}
    \begin{tabular}{l c c c c}
        \toprule
        Metric & SISS & Ours & Adaptive & Sampling \\
        \midrule
        1 Epoch (s) & 8.51 & 9.02 & -- & -- \\
        Total (s) & 297.68 & 315.61 & 370.66 & 55.05 \\
        \bottomrule
    \end{tabular} }
\end{minipage}
\hfill
\begin{minipage}[t]{0.58\textwidth}
    \centering
    \captionof{table}{MIA-based evaluation after unlearning.}
    \label{tab:mia}
    \small
    \resizebox{\linewidth}{!}{%
    \setlength{\tabcolsep}{4pt}
    \begin{tabular}{l c c c c c c}
        \toprule
        & \multicolumn{2}{c}{Orig.} & \multicolumn{2}{c}{SISS} & \multicolumn{2}{c}{Ours} \\
        \cmidrule(lr){2-3}\cmidrule(lr){4-5}\cmidrule(lr){6-7}
        & member & hold-out & member & unlearned & member & unlearned \\
        \midrule
        SecMI~\citep{duan2023diffusion} & 9.873 & 13.04 & 11.03 & 1191 & 9.873 & 56.35 \\
        PIAN~\citep{kong2024efficient}         & 1.423 & 1.596 & 1.490 & 414.6 & 1.369 & 72.84 \\
        \bottomrule
    \end{tabular}}
\end{minipage}
\vspace{-5mm}
\end{figure}

\section{Conclusion}
In this paper, we address the critical issues of quality degradation and slow convergence in data unlearning for diffusion models. Our analysis reveals that forgetting is not a uniform process, but an effective region exists across time steps and spectral domains. 
As a limitation, this work does not provide a method for automatically discovering the optimal unlearning regions. We believe that this work assists the social impact of machine unlearning within generative models.

\bibliography{example_paper}
\bibliographystyle{icml2026_fogen}

\newpage
\clearpage
\appendix
\appendix
\newpage
\section{Justification of Proposition \ref{prop:3.1}}
\label{app:proof}

\newcommand{\Prob}{\mathbb{P}}
{

\noindent\textbf{Setup.} 
Let $\mathbb{P}_{\mathrm{orig}}$ and $\mathbb{P}_m$ ($m \in \{\mathrm{full}, \mathrm{sel}\}$) be the path measures of the original and unlearned diffusion models. We denote the score perturbation as 
\begin{equation}
    \Delta s_m(x_t, t) := s_{\theta_m}(x_t, t) - s_{\theta_{\mathrm{orig}}}(x_t, t).
\end{equation} 
By Girsanov's theorem~\cite{liptser1977statistics}, the KL divergence between the original and unlearned path measures is 
\begin{equation}
     D_{\mathrm{KL}}(\mathbb{P}_{\mathrm{orig}} \,\|\, \mathbb{P}_m) = \frac{1}{2}\int_0^T g(t)^2 \mathbb{E}_{x_t} \left[ \|\Delta s_m(x_t, t)\|_2^2 \right] dt. 
\end{equation}
To evaluate unlearning efficacy, we decompose this total unlearning-induced distortion into a selected target window $\mathcal{S}$ (e.g., selected time-steps or specific frequencies) and its complement $\mathcal{S}^c$. Let 
\begin{equation}
D_{\mathrm{KL}}^{\mathcal{W}}(\mathbb{P}_{\mathrm{orig}} \,\|\, \mathbb{P}_m) := \frac{1}{2}\int_{\mathcal{W}} g(t)^2 \mathbb{E}_{x_t} [\|\Delta s_m\|_2^2] dt    
\end{equation}
denote the regional divergence over a window $\mathcal{W}$. Thus, the total divergence can be formulated as 
 \begin{equation}
     D_{\mathrm{KL}}(\mathbb{P}_{\mathrm{orig}} \,\|\, \mathbb{P}_m) = D_{\mathrm{KL}}^{\mathcal{S}}(\mathbb{P}_{\mathrm{orig}} \,\|\, \mathbb{P}_m) + D_{\mathrm{KL}}^{\mathcal{S}^c}(\mathbb{P}_{\mathrm{orig}} \,\|\, \mathbb{P}_m). 
 \end{equation}
For $\beta > 0$, we define the  forgetting--preservation utility as:
\begin{equation}\label{eq:trade_off}
    \mathcal{J}_{\beta}(m) := D_{\mathrm{KL}}^{\mathcal{S}}(\mathbb{P}_{\mathrm{orig}} \,\|\, \mathbb{P}_m) - \beta D_{\mathrm{KL}}^{\mathcal{S}^c}(\mathbb{P}_{\mathrm{orig}} \,\|\, \mathbb{P}_m).
\end{equation}
Note that $D_{\mathrm{KL}}^{\mathcal{S}}(\mathbb{P}_{\mathrm{orig}} \,\|\, \mathbb{P}_m)$ is a proxy for the \textit{forgetting action}. If $\mathcal{S}$ is chosen to contain the components that encode the forget-set information, a larger divergence on $\mathcal{S}$ means the method is actively modifying the targeted information. Conversely, $D_{\mathrm{KL}}^{\mathcal{S}^c}(\mathbb{P}_{\mathrm{orig}} \,\|\, \mathbb{P}_m)$ is a proxy for \textit{collateral distortion}. Since $\mathcal{S}^c$ is not the region we intend to unlearn, score changes there are undesirable side effects that can degrade retained content and sample quality.

\noindent\textbf{Assumptions.} 
\begin{enumerate}
    \item \textit{Finite divergence}: $D_{\mathrm{KL}}(\mathbb{P}_{\mathrm{orig}} \,\|\, \mathbb{P}_m) < \infty$.
    \item \textit{Focused updates in $\mathcal{S}$}: Due to the frequent update on the selected region under the same training budget, selective unlearning amplifies changes in the target region ($\|\Delta s_{\mathrm{sel}}\|_2 \ge \mu \|\Delta s_{\mathrm{full}}\|_2$ for $\mu \ge 1$ in $\mathcal{S}$) and weaker changes in $\mathcal{S}^c$ ($\|\Delta s_{\mathrm{sel}}\|_2 \le \eta \|\Delta s_{\mathrm{full}}\|_2$ for $\eta \in [0,1]$ in $\mathcal{S}^c$). Empirical validations are presented in Appendix~\ref{app:proof_emp}.
\end{enumerate}

\begin{proposition}
Under the assumptions above, the trade-off utility of $\mathcal{J}_{\beta}(\mathrm{sel})$ is larger than $\mathcal{J}_{\beta}(\mathrm{full)}$. For every  $\beta>0$ and $\lambda := \frac{D_{\mathrm{KL}}^{\mathcal{S}^c}(\mathbb{P}_{\mathrm{orig}} \,\|\, \mathbb{P}_{\mathrm{full}})}{D_{\mathrm{KL}}(\mathbb{P}_{\mathrm{orig}} \,\|\, \mathbb{P}_{\mathrm{full}})} \in [0,1]$, their gap satisfies:
\begin{equation}
    \mathcal{J}_{\beta}(\mathrm{sel}) - \mathcal{J}_{\beta}(\mathrm{full}) \ge \left[ (\mu^2-1)(1-\lambda) + \beta(1-\eta^2)\lambda \right] D_{\mathrm{KL}}(\mathbb{P}_{\mathrm{orig}} \,\|\, \mathbb{P}_{\mathrm{full}}),
\end{equation}  
where $\lambda$ indicates the proportion of collateral distortion compared to the full-unlearning.
\end{proposition}

\begin{proof}
From the definition of the utility function given in Eq.~\ref{eq:trade_off}, the difference is:
\begin{align*}
    \mathcal{J}_{\beta}(\mathrm{sel}) - \mathcal{J}_{\beta}(\mathrm{full}) &= \left( D_{\mathrm{KL}}^{\mathcal{S}}(\mathbb{P}_{\mathrm{orig}} \,\|\, \mathbb{P}_{\mathrm{sel}}) - D_{\mathrm{KL}}^{\mathcal{S}}(\mathbb{P}_{\mathrm{orig}} \,\|\, \mathbb{P}_{\mathrm{full}}) \right) \\
    &\quad - \beta \left( D_{\mathrm{KL}}^{\mathcal{S}^c}(\mathbb{P}_{\mathrm{orig}} \,\|\, \mathbb{P}_{\mathrm{sel}}) - D_{\mathrm{KL}}^{\mathcal{S}^c}(\mathbb{P}_{\mathrm{orig}} \,\|\, \mathbb{P}_{\mathrm{full}}) \right).
\end{align*}
By Assumption 2, the condition $\|\Delta s_{\mathrm{sel}}\|_2 \ge \mu \|\Delta s_{\mathrm{full}}\|_2$ in $\mathcal{S}$ ensures the divergence in the forget-window satisfies:
$$D_{\mathrm{KL}}^{\mathcal{S}}(\mathbb{P}_{\mathrm{orig}} \,\|\, \mathbb{P}_{\mathrm{sel}}) \ge \mu^2 D_{\mathrm{KL}}^{\mathcal{S}}(\mathbb{P}_{\mathrm{orig}} \,\|\, \mathbb{P}_{\mathrm{full}}).$$
Moreover, the condition $\|\Delta s_{\mathrm{sel}}\|_2 \le \eta \|\Delta s_{\mathrm{full}}\|_2$ in $\mathcal{S}^c$ (Assumption 3) ensures the divergence in the non-target window satisfies:
$$D_{\mathrm{KL}}^{\mathcal{S}^c}(\mathbb{P}_{\mathrm{orig}} \,\|\, \mathbb{P}_{\mathrm{sel}}) \le \eta^2 D_{\mathrm{KL}}^{\mathcal{S}^c}(\mathbb{P}_{\mathrm{orig}} \,\|\, \mathbb{P}_{\mathrm{full}}).$$
Substituting these bounds into the utility difference yields:
$$\mathcal{J}_{\beta}(\mathrm{sel}) - \mathcal{J}_{\beta}(\mathrm{full}) \ge (\mu^2-1) D_{\mathrm{KL}}^{\mathcal{S}}(\mathbb{P}_{\mathrm{orig}} \,\|\, \mathbb{P}_{\mathrm{full}}) + \beta(1-\eta^2) D_{\mathrm{KL}}^{\mathcal{S}^c}(\mathbb{P}_{\mathrm{orig}} \,\|\, \mathbb{P}_{\mathrm{full}}).$$
By the definitions of $\lambda$ and total $D_{\mathrm{KL}}$, we can express the regional divergences as:
$$D_{\mathrm{KL}}^{\mathcal{S}^c}(\mathbb{P}_{\mathrm{orig}} \,\|\, \mathbb{P}_{\mathrm{full}}) = \lambda D_{\mathrm{KL}}(\mathbb{P}_{\mathrm{orig}} \,\|\, \mathbb{P}_{\mathrm{full}}), \quad D_{\mathrm{KL}}^{\mathcal{S}}(\mathbb{P}_{\mathrm{orig}} \,\|\, \mathbb{P}_{\mathrm{full}}) = (1-\lambda) D_{\mathrm{KL}}(\mathbb{P}_{\mathrm{orig}} \,\|\, \mathbb{P}_{\mathrm{full}}).$$
Substituting these exact expressions into the lower bound directly yields the stated inequality:
\begin{align*}
    \mathcal{J}_{\beta}(\mathrm{sel}) - \mathcal{J}_{\beta}(\mathrm{full}) &\ge (\mu^2-1)(1-\lambda)D_{\mathrm{KL}}(\mathbb{P}_{\mathrm{orig}} \,\|\, \mathbb{P}_{\mathrm{full}}) + \beta(1-\eta^2)\lambda D_{\mathrm{KL}}(\mathbb{P}_{\mathrm{orig}} \,\|\, \mathbb{P}_{\mathrm{full}}) \\
    &= \left[ (\mu^2-1)(1-\lambda) + \beta(1-\eta^2)\lambda \right] D_{\mathrm{KL}}(\mathbb{P}_{\mathrm{orig}} \,\|\, \mathbb{P}_{\mathrm{full}}).
\end{align*}
\end{proof}

\noindent\textbf{Remark 1.} \textit{The bracketed term $(\mu^2-1)(1-\lambda) + \beta(1-\eta^2)\lambda$ is always non-negative under our assumptions ($\mu \ge 1, \eta \le 1, \beta>0, \lambda \in [0,1]$). Then, the improvement in utility is strictly positive.}

\noindent\textbf{Remark 2.} \textit{The utility gap $\mathcal{J}_{\beta}(\mathrm{sel}) - \mathcal{J}_{\beta}(\mathrm{full})\geq0$ indicates our selective mechanism focuses score updates on the selected forget-region while mitigating unwanted distortion in non-selected parts.}

\subsection{Empirical Validation of Assumptions} 
\label{app:proof_emp}

Our theoretical analysis relies on two key assumptions: enhanced forgetting in the target region (Assumption 2, $\mu \ge 1$) and suppressed distortion in the non-target region (Assumption 2, $\eta \le 1$). We empirically validate both assumptions across the time and frequency domains.

\paragraph{Enhanced Forgetting ($\mu \ge 1$).} 
To verify that targeted training induces stronger score updates within the specified region compared to full unlearning, we evaluate the $\mu$-ratio ($\|\Delta s_{\mathrm{sel}}\|_2^2 / \|\Delta s_{\mathrm{full}}\|_2^2$) strictly within the targeted window $\mathcal{S}$. As shown in \Tabref{tab:mu_ratio}, the relative ratios are strictly greater than 1 across all targeted time windows (\Tabref{tab:window_ratio_time_mu}) and frequency bands (\Tabref{tab:window_ratio_freq_mu}). This confirms that selective unlearning strengthens its update energy on the selected region.

\begin{table}[h]
\centering
\caption{Empirical validation of $\mu^2 \ge 1$. We report the ratio $\|\Delta s_{\mathrm{sel}}\|_2^2 / \|\Delta s_{\mathrm{full}}\|_2^2$ strictly within the target region $\mathcal{S}$. Values $>1$ indicate that targeted training induces stronger score updates compared to full unlearning. $W_1$, $W_2$, and $W_3$ denote the early $[0, 250)$, middle $[250, 750)$, and late $[750, 1000]$ diffusion time windows. Low, Mid, and High correspond to the spatial frequency bands.}
\label{tab:mu_ratio}
\begin{subtable}[t]{0.45\linewidth}
\centering
\caption{Time-selective Unlearning}
\label{tab:window_ratio_time_mu}
\resizebox{\linewidth}{!}{
\begin{tabular}{lc}
\toprule
\textbf{Target Region ($\mathcal{S}$)} & \textbf{Ratio} ($\|\Delta s_{\mathrm{sel}}\|_2^2 / \|\Delta s_{\mathrm{full}}\|_2^2$) \\
\midrule
$W_1$ (Early) & 1.2515 \\
$W_2$ (Middle) & 1.4950 \\
$W_3$ (Late) & 1.3095 \\
\bottomrule
\end{tabular}
}
\end{subtable}
\hfill
\begin{subtable}[t]{0.45\linewidth}
\centering
\caption{Frequency-selective Unlearning}
\label{tab:window_ratio_freq_mu}
\resizebox{\linewidth}{!}{
\begin{tabular}{lc}
\toprule
\textbf{Target Region ($\mathcal{S}$)} & \textbf{Ratio} ($\|\Delta s_{\mathrm{sel}}\|_2^2 / \|\Delta s_{\mathrm{full}}\|_2^2$) \\
\midrule
Low-pass & 1.2403 \\
Mid-pass & 13.3800 \\
High-pass & 1.3850 \\
\bottomrule
\end{tabular}
}
\end{subtable}
\end{table}

\paragraph{Suppressed Distortion ($\eta \le 1$).} 
To quantify the bounded out-of-region effect, we measure the normalized perturbation across different evaluation domains when the model is trained on a specific target region. The results in \Tabref{tab:eta_ratio} indicate that selective unlearning predominantly affects the intended phase or band compared to non-target counterparts. The off-diagonal elements in both the time domain (\Tabref{tab:window_ratio_time_eta}) and the frequency domain (\Tabref{tab:window_ratio_freq_eta}) experimentally support $\eta < 1$.

\begin{table}[h]
\centering
\caption{Empirical validation of $\eta^2 \le 1$. We report the normalized perturbation $\|\Delta s_{\mathrm{sel}}\|_2^2$ across evaluation regions (rows) when trained on a specific target $\mathcal{S}$ (columns). Diagonal elements are scaled to 1, while off-diagonal elements show the bounded effects of collateral distortion.}
\label{tab:eta_ratio}
\begin{subtable}[t]{0.45\linewidth}
\centering
\caption{Time-selective (Cross-Region)}
\label{tab:window_ratio_time_eta}
\resizebox{\linewidth}{!}{
\begin{tabular}{lccc}
\toprule
& \multicolumn{3}{c}{\textbf{Target Region ($\mathcal{S}$)}} \\
\cmidrule(lr){2-4}
\textbf{Eval Region} & $W_1$ & $W_2$ & $W_3$ \\
\midrule
$W_1$ & 1.0000 & 0.2511 & 0.0009 \\
$W_2$ & 0.7532 & 1.0000 & 0.0010 \\
$W_3$ & 0.4939 & 0.7037 & 1.0000 \\
\bottomrule
\end{tabular}
}
\end{subtable}
\hfill
\begin{subtable}[t]{0.45\linewidth}
\centering
\caption{Frequency-selective (Cross-Region)}
\label{tab:window_ratio_freq_eta}
\resizebox{\linewidth}{!}{
\begin{tabular}{lccc}
\toprule
& \multicolumn{3}{c}{\textbf{Target Region ($\mathcal{S}$)}} \\
\cmidrule(lr){2-4}
\textbf{Eval Region} & Low & Mid & High \\
\midrule
Low  & 1.0000 & 0.2535 & 0.1681 \\
Mid  & 0.4116 & 1.0000 & 0.0511 \\
High & 0.0502 & 0.0814 & 1.0000 \\
\bottomrule
\end{tabular}
}
\end{subtable}
\end{table}

\subsection{Time and Frequency Understanding for $\lambda$}
\label{sec:theory_time_vC}

\paragraph{Time-Selective Unlearning.}
Defining the target region as a specific time window $\mathcal{S} = \mathcal{T}_s \subset [0, T]$ formulates time-selective unlearning. In this case, the ratio $\lambda_t$ represents the fraction of the full-unlearning distortion that occurs outside the targeted time window $\mathcal{T}_s$:
\begin{equation}\label{eq:lambda_time_vC}
    \lambda_t \;:=\; \frac{\int_{[0,T]\setminus\mathcal{T}_s} g(t)^2\,\mathbb{E}_{x_t}\!\left[\|\Delta s_{\mathrm{full}}(x_t, t)\|_2^2\right] dt}{\int_0^T g(t)^2\,\mathbb{E}_{x_t}\!\left[\|\Delta s_{\mathrm{full}}(x_t, t)\|_2^2\right] dt} \;\in\;[0,1].
\end{equation}

\paragraph{Frequency-Selective Unlearning.}
\label{sec:theory_freq_vC}
Plancherel's theorem~\citep{yosida2012functional} establishes an equivalence between spatial and spectral energy:
\begin{equation}\label{eq:plancherel_vC}
    \|\Delta s_m(x_t, t)\|_2^2 \;=\; \sum_{\boldsymbol{\omega}} \|\widehat{\Delta s}_m(\boldsymbol{\omega},t)\|_2^2,
\end{equation}
where $\widehat{\Delta s}_m(\boldsymbol{\omega},t)$ is the discrete Fourier transform of the score perturbation across spatial dimensions at frequency $\boldsymbol{\omega}$. By applying a low-pass filter with cutoff $r$, we define our targeted forgetting region in the frequency domain as $\mathcal{S} = \Omega_r := \{|\boldsymbol{\omega}|\le r\}$. Consequently, the collateral distortion ratio $\lambda_\omega$, corresponding to the unselected high-frequency band, is defined as:
\begin{equation}\label{eq:lambda_freq_vC}
    \lambda_\omega \;:=\; \frac{\int_0^T g(t)^2\!\sum_{|\boldsymbol{\omega}|>r}\mathbb{E}_{x_t}\!\left[\|\widehat{\Delta s}_{\mathrm{full}}(\boldsymbol{\omega},t)\|_2^2\right] dt}{\int_0^T g(t)^2\!\sum_{\boldsymbol{\omega}}\mathbb{E}_{x_t}\!\left[\|\widehat{\Delta s}_{\mathrm{full}}(\boldsymbol{\omega},t)\|_2^2\right] dt} \;\in\;[0,1].
\end{equation}

\section{Experimental Details}
\label{app:exp}

\paragraph{Experimental Setups.}
We basically follow the experimental setups in SISS \citep{alberti2025data}. We first make note of an important detail in their paper. 

All the diffusion models were provided in the Huggingface \texttt{diffusers} library with a U-Net backbone. For the CelebA-HQ dataset, we used a pretrained checkpoint released by~\citep{ho2020denoising} at \url{https://huggingface.co/google/ddpm-celebahq-256}. For the Stable Diffusion experiments, we used version 1.4 at \url{https://huggingface.co/CompVis/stable-diffusion-v1-4} as the pretrained checkpoint and used a 50-step DDIM sampler~\citep{song2020denoising}.

Regarding hyperparameters, we follow \citep{alberti2025data} without additional tuning.

For the experimental setups, we mainly use NVIDIA A40 GPUs with the PyTorch library and utilize NVIDIA A100 GPUs for parallel runs.

\paragraph{Memorized Prompts of Stable Diffusion.}
\citet{alberti2025data} constructed 45 prompts in Stable Diffusion v1.4. Since only one LAION image corresponds to each prompt, synthetic datasets are generated by sampling 128 images and applying k-means clustering for classification. A ``fully-memorized” prompt refers to a case where Stable Diffusion repeatedly reproduces the same outcome, whereas a ``partially-memorized” prompt is obtained by manually adding or deleting tokens, producing more diverse outputs that are easier to unlearn.

\paragraph{Preference Optimization.}

Direct Preference Optimization (DPO) \citep{rafailov2023direct} is widely used to evaluate preference alignment in language models. Originally developed for Reinforcement Learning from Human Feedback (RLHF), it has recently also been applied to diffusion fine-tuning \citep{wallace2024diffusion}. For unlearning, Negative Preference Optimization (NPO) \citep{zhang2024negative} has been proposed as an alternative to gradient descent. Unlike gradient ascent \citep{zhang2024negative}, NPO leverages the initial model as a reference point, which helps mitigate overfitting by keeping the optimization close to the initialization. In the diffusion setting, \citet{wang2025diffusionnpo} applied NPO for alignment. Another line of work, Kahneman–Tversky Optimization (KTO) \citep{ethayarajhmodel}, has its own strength since the method does not require positive–negative pairs. \citet{li2024aligning} extended KTO to diffusion for pair-free feedback alignment. 
In this paper, we follow the formulation of forget loss with DPO \citep{wallace2024diffusion} and KTO \citep{li2024aligning}.

{
To formulate this, let the per-sample diffusion loss be $\ell(\theta, \rvx) = \mathbb{E}_{t}[w_t \| s_\theta(\rvx_t, t) - \nabla_{\rvx_t} \log q(\rvx_t|\rvx) \|_2^2]$. We define the implicit reward as the loss reduction relative to the reference model: $r(\theta, \rvx) = \ell(\theta_{\text{ref}}, \rvx) - \ell(\theta, \rvx)$. By maximizing the implicit reward gap defined by the difference in diffusion loss relative to the reference model $\theta_{\text{ref}}$, the framework effectively minimizes the likelihood of generating forget concepts while preserving general capabilities. The objectives for DPO and KTO are then given by for forget data $\rvx_f$ and retain data $\rvx_r$:
\begin{align}
\mathcal{L}_{\text{DPO}}(\theta) &= - \mathbb{E}_{\rvx_r \sim \mathcal{D}_R, \rvx_f \sim \mathcal{D}_F} \Bigl[ \log \sigma \Bigl( \beta \bigl( r(\theta, \rvx_r) - r(\theta, \rvx_f) \bigr) \Bigr) \Bigr]. \label{eq:dpo} \\
\mathcal{L}_{\text{KTO}}(\theta) &= \mathbb{E}_{\rvx_r \sim \mathcal{D}_R} \Bigl[ 1 - \sigma \bigl( \beta r(\theta, \rvx_r) - z_{\text{ref}} \bigr) \Bigr] \nonumber \\
&\quad + \lambda \mathbb{E}_{\rvx_f \sim \mathcal{D}_F} \Bigl[ 1 - \sigma \bigl( z_{\text{ref}} - \beta r(\theta, \rvx_f) \bigr) \Bigr]. \label{eq:kto}
\end{align}
$\sigma$ is the sigmoid function, $\beta$ is a temperature hyperparameter, and $z_{\text{ref}}$ is the reference point for the reward.

\paragraph{Algorithm of Adaptive Search.}
Based on \citet{jain2025classifier}, we present pseudocode of the adaptive transition point $\tau=t_S$ search method in Algorithm \ref{alg:adaptive_unlearning}.

\begin{algorithm}[!ht]
\caption{{Adaptive Transition Time Search and Selective Unlearning}}
\label{alg:adaptive_unlearning}
\begin{algorithmic}[1]

\Require Pretrained model $\theta^{*}$; prompt embedding $e_p$; unconditional embedding $e_\emptyset$;
\Require Forget set $\mathcal{D}_F$; retain set $\mathcal{D}_R$; total time steps $T$

\Statex \textbf{Phase 1: Adaptive search for transition time $\tau$}

\State Sample $x_T \sim \mathcal{N}(0,\mathbf{I})$
\State $\tau \gets \text{None}$
\State $d_{\text{prev}} \gets +\infty$

\For{$t$ following the denoising schedule from $T$ to $1$}
    \State $d_t \gets \|\epsilon_\theta(x_t, e_p) - \epsilon_\theta(x_t, e_\emptyset)\|_2$
    \If{$\tau = \text{None}$ \textbf{ and } $d_t > d_{\text{prev}}$}
        \State $\tau \gets t_{\text{prev}}$ \Comment{first local minimum along the trajectory}
    \EndIf
    \State $d_{\text{prev}} \gets d_t$
    \State $t_{\text{prev}} \gets t$
    \State $x_{t-1} \gets \mathrm{DenoisingStep}(x_t, \epsilon_\theta)$
\EndFor

\Statex \textbf{Phase 2: Seletive Unlearning}

\State $t_S \gets \tau$
\State Perform unlearning for timesteps $t \in [t_S, T]$

\end{algorithmic}
\end{algorithm}

}
\section{Supplementary Experiments}
\label{app:result}

{
\subsection{Different Backbones}
To evaluate the adaptability of latent models for image-level generation, we conduct experiments using the Latent Diffusion Model (LDM) \citep{rombach2022high} pretrained on the CelebHQ dataset\footnote{\url{https://huggingface.co/CompVis/ldm-celebahq-256}}. The results are presented in \Tabref{tab:main_ldm}.
Furthermore, we extend our evaluation to Stable Diffusion v3.0. built upon the Diffusion Transformer (DiT) architecture \citep{peebles2023scalable}. Since Stable Diffusion v3.0 has practical guardrail methods to prevent the memorization observed in previous Stable Diffusion v1.4, it is difficult to generate the memorized example in Stable 1.4. Thus, we rather calculate the SSCD and SSCD$^{\text{norm}}$ while using the same prompts. The corresponding results are reported in \Tabref{tab:sd3_results}. 
The gains of our method are smaller compared to Stable Diffusion v1.4 due to the flow-matching-based property, but we still observe that controlling time and frequency helps unlearning.

Tables \ref{tab:main_ldm} and \ref{tab:sd3_results} demonstrate the adaptability and effectiveness of our proposed method across diverse architectures. We did not perform additional hyperparameter tuning for these experiments.

\begin{table}[!ht]
\centering
\caption{{Comparison of unlearning methods on latent diffusion models with a relative Gain. For each method, we show the baseline scores, the scores with our method applied (+ Ours), and the resulting relative gain (\%) in a separate row. For the 'Gain (\%)' row, a higher value always indicates better performance.} Positive gains are colored in \textcolor{myblue}{blue}, and degradations are in \textcolor{myred}{red}.}
\label{tab:main_ldm}
\resizebox{0.8\textwidth}{!}{%
\begin{tabular}{l l c ccc ccc}
\toprule
\multicolumn{2}{l}{\multirow{2}{*}{\textbf{Method}}} & & \multicolumn{3}{c}{Denoising from $t=250$} & \multicolumn{3}{c}{Denoising from $t=500$} \\
\cmidrule(lr){4-6} \cmidrule(lr){7-9}
& & FID-10K $\downarrow$ & SSCD $\downarrow$ & SSCD$^{\text{norm}}$ $\downarrow$ & Aesthetic $\uparrow$ & SSCD $\downarrow$ & SSCD$^{\text{norm}}$ $\downarrow$ & Aesthetic $\uparrow$ \\
\midrule

\multirow{3}{*}{{SISS}} & Base & 17.800 & 0.431 & 0.547 & 4.601 & 0.281 & 0.588 & 5.092 \\
& + Ours & 16.960 & 0.437 & 0.520 & 5.076 & 0.250 & 0.536 & 4.955 \\ \cmidrule{2-9}
& \textit{Gain (\%) } & \textcolor{myblue}{+4.72} & \textcolor{myred}{-1.54} & \textcolor{myblue}{+4.95} & \textcolor{myblue}{+10.32} & \textcolor{myblue}{+10.90} & \textcolor{myblue}{+8.84} & \textcolor{myred}{-2.69} \\
\midrule
\multirow{3}{*}{{DPO}} & Base & 28.630 & 0.428 & 0.573 & 4.505 & 0.372 & 0.630 & 4.880 \\
& + Ours & 28.630 & 0.372 & 0.630 & 4.880 & 0.295 & 0.546 & 4.996 \\ \cmidrule{2-9}
& \textit{Gain (\%) } & \textcolor{black}{0.00} & \textcolor{myblue}{+13.02} & \textcolor{myred}{-9.89} & \textcolor{myblue}{+8.34} & \textcolor{myblue}{+20.63} & \textcolor{myblue}{+13.34} & \textcolor{myblue}{+2.38} \\
\midrule
\multirow{3}{*}{{KTO}} & Base & 28.630 & 0.372 & 0.630 & 4.880 & 0.458 & 0.472 & 5.812 \\
& + Ours & 31.360 & 0.498 & 0.576 & 5.072 & 0.457 & 0.493 & 6.446 \\ \cmidrule{2-9}
& \textit{Gain (\%)} & \textcolor{myred}{-9.54} & \textcolor{myred}{-33.65} & \textcolor{myblue}{+8.58} & \textcolor{myblue}{+3.93} & \textcolor{myblue}{+0.19} & \textcolor{myred}{-4.48} & \textcolor{myblue}{+10.90} \\
\bottomrule
\end{tabular}%
}
\end{table}

\begin{table}[!ht]
    \centering
    \caption{{Quantitative comparison on Stable Diffusion v3.0. with Diffusion Transformer (DiT) backbone. Arrows indicate the direction of better performance ($\uparrow$ for higher, $\downarrow$ for lower).}}
    \label{tab:sd3_results}
    \begin{tabular}{l c c}
        \toprule
        {Measure} & {SISS} & {SISS + Ours} \\
        \midrule
        CLIP-IQA (Partial) ($\uparrow$) & 0.863 & 0.856 \small{(-0.80\%)} \\
        CLIP-IQA (Full) ($\uparrow$)    & 0.886 & 0.890 \small{(+0.50\%)} \\
        \midrule
        SSCD (Partial) ($\downarrow$)      & 0.349 & 0.302 \small{(-13.67\%)} \\
        SSCD (Full) ($\downarrow$)         & 0.421 & 0.362 \small{(-14.06\%)} \\
        SSCD Norm (Partial) ($\downarrow$) & 0.704 & 0.698 \small{(-0.87\%)} \\
        SSCD Norm (Full) ($\downarrow$)    & 0.724 & 0.707 \small{(-2.30\%)} \\
        \bottomrule
    \end{tabular}
\end{table}

\subsection{Concept Unlearning}
As stated in the introduction, unlearning specific concepts, such as nudity or violence, is another major direction in unlearning research. We examine whether the choice of timesteps also affects concept generation in terms of “memorization of certain concepts.” We utilize the nudity prompts used in Ring-a-bell \cite{tsai2024ringabell}
using the GitHub\footnote {\url{https://github.com/jaehong31/SAFREE/blob/main/datasets/nudity-ring-a-bell.csv}}  of SAFREE \citep{yoon2025safree}.

We compare SISS and SISS+Ours using the same time window used SD 1.4, and compute the CLIP-IQA quality score. For the unlearning measure, we measure the nudity detection rate using NudeNet\footnote {\url{https://github.com/notAI-tech/NudeNet}}, where an attack is counted if the NudeNet classifier probability exceeds 0.45. As shown in \Figref{fig:nudity}, the proposed one shows faster convergence and superior forgetting behavior.

\begin{figure*}[!ht]
    \centering
    \begin{subfigure}[b]{0.33\textwidth}
        \centering
        \includegraphics[width=\linewidth]{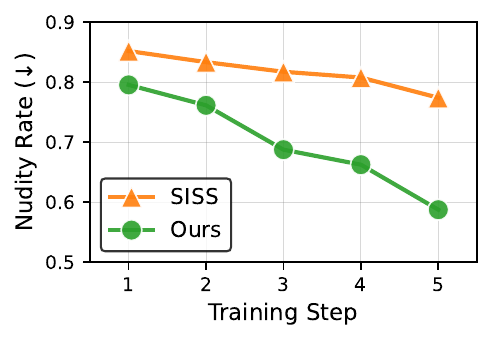}
        \caption{Nudity Detection rate}
    \end{subfigure}%
        \begin{subfigure}[b]{0.33\textwidth}
        \centering
        \includegraphics[width=\linewidth]{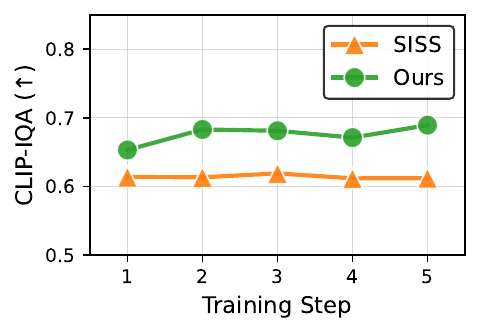}
        \caption{CLIP-IQA (Quality)}
    \end{subfigure}%
    \caption{{Unlearning with concept ``nudity'' using Ring-a-bell dataset. Our methods show faster unlearning convergence and a high nudity deletion rate.}}
    \label{fig:nudity}
    \vspace{-3mm}
\end{figure*}

\subsection{Importance of Unlearned Quality}
We argue that significant degradation or distinct artifacts in these outputs can create a privacy vulnerability. If the unlearned model generates evident anomalies for specific queries, it provides a clue to adversaries, allowing them to identify which data points were deleted (i.e., membership inference via outlier detection).

To verify this hypothesis, we conducted an experiment using advanced Large Language Models (LLMs) as anomaly detectors. We provided the LLMs with generated images and instructed them to evaluate whether the images are realistic (in-distribution) or abnormal (out-of-distribution) based on the following prompt:

\begin{quote}
\textit{Prompt: ``You are an image quality and anomaly detection expert. You are given two image sets: The first set contains reference (in-distribution) images. Assign a score to each query image: 1.0 = Realistic / In-distribution (no visible generation problem), 0.0 = Abnormal / Out-of-distribution (clear generation failure or artifacts).''}
\end{quote}

As shown in \Tabref{tab:llm_anomaly}, baseline methods like SISS exhibited low scores, indicating that their outputs are easily distinguishable as anomalies. In contrast, our method maintains higher generation quality, effectively masking the trace of unlearning.
Furthermore, recent studies \citep{bertran2024reconstruction, gao2025meta} warn that unlearning mechanisms are vulnerable against relearning attacks, which can recover deleted data using residual information such as embedding similarity. Therefore, ensuring the quality of unlearned samples is crucial not only for aesthetics but also for preventing privacy attacks.

}

\begin{table}[!ht]
    \centering
    \caption{{Anomaly detection scores evaluated by various LLMs on unlearned outputs. A higher score indicates the generated image is perceived as realistic (in-distribution), making it harder to detect that unlearning has occurred.}}
    \label{tab:llm_anomaly}
    \begin{tabular}{l c c}
        \toprule
        {LLM Evaluator} & {SISS} & {SISS + Ours} \\
        \midrule
        ChatGPT-5         & 0.22 & 0.67 \\
        Claude Sonnet 4.5 & 0.13 & 0.83 \\
        Gemini 2.5 Flash  & 0.00 & 0.50 \\
        \bottomrule
    \end{tabular}
\end{table}

{

To investigate the impact of unlearning on similar samples, we choose the five nearest neighbors for each forgotten image, and report the mean SSCD between their original and reconstructed images as well as the aesthetic score of the reconstructed samples in~\Tabref{tab:nn}. As data unlearning focuses on individual deletion, the SSCD on these nearest images does not decrease. However, we do observe a slight drop in aesthetic quality after unlearning. By implementing our selective unlearning, this aesthetic degradation is mitigated. Consequently, selective unlearning helps preserve both the semantics and the overall quality of nearby examples.

\subsection{Ablation on Frequency Filtering}

\textbf{Filter band ablation.$ $}
We evaluate low-pass (our method), mid-pass, and high-pass filters for unlearning in~\Tabref{tab:filter_band}. The results indicate that deleting low-frequency information while preserving high-frequency details performs the best, similar to not deleting Phase~III in time steps. Otherwise, both unlearning and quality preservation fail.

\textbf{Frequency cutoff scheduling.$ $}
Our frequency selection can be further improved with different filtering schedules. Therefore, we test four additional variants during training, including Linear Increase ($0.05 \to 0.2$), Linear Decay ($0.2 \to 0.05$), and Stepwise strategies, to assess the impact of filtering high-frequency content over the unlearning fine-tuning. As shown in~\Tabref{tab:freq_schedule}, the Linear Increase strategy, which retains a greater amount of high-frequency information during the later training stages, results in the highest image quality. This indicates that frequency filtering should be dynamically adjusted to preserve fine-grained details during the later stages of unlearning.

\subsection{Statistical reliability}
To verify that the gap is statistically significant, we further evaluate the SISS and KTO experiments in~\Tabref{tab:variance} across five independent runs, which yield small standard deviations.

\begin{table}[t]
    \centering
    \caption{Reconstruction on the five nearest neighbors of each forgotten image. Selective unlearning preserves both semantics (SSCD) and visual quality (aesthetic score) of nearby samples.}
    \label{tab:nn}
    \small
    \setlength{\tabcolsep}{6pt}
    \begin{tabular}{l c c c c}
        \toprule
        & \multicolumn{2}{c}{$t = 250$} & \multicolumn{2}{c}{$t = 500$} \\
        \cmidrule(lr){2-3}\cmidrule(lr){4-5}
        Model & SSCD $\downarrow$ & Aesthetic $\uparrow$ & SSCD $\downarrow$ & Aesthetic $\uparrow$ \\
        \midrule
        Orig.   & 0.8817 & 6.3894 & 0.7405 & 6.7483 \\
        SISS    & 0.8812 & 6.2441 & 0.7394 & 6.5261 \\
        + Ours  & 0.8797 & 6.4467 & 0.7439 & 6.5873 \\
        \bottomrule
    \end{tabular}
    \vspace{-3mm}
\end{table}

\begin{table}[t]
    \centering
    \caption{Effect of filter band on selective unlearning. The low-pass choice, which preserves high-frequency detail, yields the best trade-off across both timesteps.}
    \label{tab:filter_band}
    \small
    \setlength{\tabcolsep}{6pt}
    \begin{tabular}{l c c c c}
        \toprule
        & \multicolumn{2}{c}{$t = 250$} & \multicolumn{2}{c}{$t = 500$} \\
        \cmidrule(lr){2-3}\cmidrule(lr){4-5}
        Filter band & SSCD $\downarrow$ & Aesthetic $\uparrow$ & SSCD $\downarrow$ & Aesthetic $\uparrow$ \\
        \midrule
        Low-pass (ours) & \textbf{0.331} & \textbf{5.553} & \textbf{0.264} & \textbf{5.904} \\
        Mid-pass        & 0.364 & 4.374 & 0.329 & 5.043 \\
        High-pass       & 0.379 & 4.143 & 0.227 & 4.670 \\
        \bottomrule
    \end{tabular}
    \vspace{-3mm}
\end{table}

\begin{table}[!ht]
    \centering
    \caption{Different schedules of the low-pass cutoff $r_t$ during unlearning fine-tuning. Linear Increase ($0.05 \to 0.2$) gives the best quality at both evaluation timesteps.}
    \label{tab:freq_schedule}
    \small
    \setlength{\tabcolsep}{3pt}
    \begin{tabular}{l c c c c c c}
        \toprule
        & \multicolumn{3}{c}{$t = 250$} & \multicolumn{3}{c}{$t = 500$} \\
        \cmidrule(lr){2-4}\cmidrule(lr){5-7}
        Schedule & SSCD $\downarrow$ & SSCD$^{\text{norm}}$ $\downarrow$ & Aesth.\ $\uparrow$ & SSCD $\downarrow$ & SSCD$^{\text{norm}}$ $\downarrow$ & Aesth.\ $\uparrow$ \\
        \midrule
        SISS              & 0.336 & 0.430 & 4.094 & 0.299 & 0.501 & 4.845 \\
        Fixed             & 0.345 & 0.349 & 5.520 & 0.282 & 0.399 & \textbf{6.095} \\
        Linear Increase   & \textbf{0.323} & \textbf{0.322} & \textbf{5.709} & 0.260 & \textbf{0.382} & 5.832 \\
        Linear Decay      & 0.429 & 0.422 & 5.296 & 0.317 & 0.448 & 5.722 \\
        Stepwise Increase & 0.328 & 0.331 & 5.631 & \textbf{0.255} & 0.392 & 5.918 \\
        Stepwise Decay    & 0.431 & 0.419 & 5.186 & 0.317 & 0.438 & 5.535 \\
        \bottomrule
    \end{tabular}
    \vspace{-3mm}
\end{table}

\begin{table}[!ht]
    \centering
    \caption{Mean $\pm$ standard deviation over five independent runs on CelebA-HQ. Adding our selective scheme yields consistent gains on the perturbation-normalized metric SSCD$^{\text{norm}}$ and on the aesthetic score, with negligible variance.}
    \label{tab:variance}
    \scriptsize
    \setlength{\tabcolsep}{3pt}
    \begin{tabular}{l c c c c c c c}
        \toprule
        & & \multicolumn{3}{c}{$t = 250$} & \multicolumn{3}{c}{$t = 500$} \\
        \cmidrule(lr){3-5}\cmidrule(lr){6-8}
        Method & FID $\downarrow$ & SSCD $\downarrow$ & SSCD$^{\text{norm}}$ $\downarrow$ & Aesth.\ $\uparrow$ & SSCD $\downarrow$ & SSCD$^{\text{norm}}$ $\downarrow$ & Aesth.\ $\uparrow$ \\
        \midrule
        SISS    & $23.42_{\pm 0.31}$ & $0.336_{\pm 0.002}$ & $0.430_{\pm 0.003}$ & $4.094_{\pm 0.054}$ & $0.299_{\pm 0.006}$ & $0.501_{\pm 0.004}$ & $4.845_{\pm 0.051}$ \\
        + Ours  & $23.65_{\pm 0.35}$ & $0.345_{\pm 0.003}$ & $0.349_{\pm 0.002}$ & $5.520_{\pm 0.063}$ & $0.282_{\pm 0.004}$ & $0.399_{\pm 0.003}$ & $6.095_{\pm 0.049}$ \\
        KTO     & $23.48_{\pm 0.15}$ & $0.363_{\pm 0.002}$ & $0.366_{\pm 0.001}$ & $5.355_{\pm 0.062}$ & $0.289_{\pm 0.007}$ & $0.442_{\pm 0.005}$ & $5.879_{\pm 0.089}$ \\
        + Ours  & $23.58_{\pm 0.12}$ & $0.373_{\pm 0.003}$ & $0.340_{\pm 0.003}$ & $5.470_{\pm 0.049}$ & $0.280_{\pm 0.010}$ & $0.377_{\pm 0.006}$ & $6.274_{\pm 0.024}$ \\
        \bottomrule
    \end{tabular}
    \vspace{-3mm}
\end{table}
}
\newpage
\subsection{Base Experiments}

We illustrate some details through a toy example. For \Figref{fig:toy_exp}, we construct a multi-layer perceptron diffusion model trained with the DDPM objective \citep{ho2020denoising}. The model is trained until convergence. After training, we delete the forget samples ($\times$) using gradient ascent, while applying gradient descent to certain retain data points (◆).

Except for \Figref{fig:toy_exp}, all experiments in Section 3 are conducted on the CelebHQ dataset used in the main table \ref{tab:main}. We now provide additional results.
In \Figref{fig:distribution_time}, we utilize the DINOv3 embedding \citep{simeoni2025dinov3} to calculate the similarity between data samples. We also measured in pixel-wise similarity in \Figref{fig:pixel}, where we failed to observe similar patterns. The nearest images of individual data samples are in \Figref{fig:app_nn}. Individual gradient norms are in \Figref{fig:gradient_norms_supp}.
Finally, we report the remaining results of \Figref{fig:psa} in Figures \ref{fig:psa2} and \ref{fig:psa3}.

\clearpage
\newpage
\begin{figure}[!ht]
    \centering
    \begin{subfigure}[!ht]{0.49\linewidth}
        \centering
        \includegraphics[width=\linewidth]{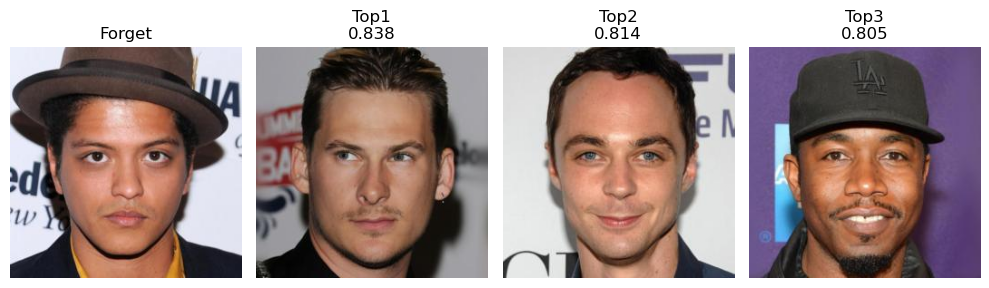}
    \end{subfigure}
    \hfill
    \begin{subfigure}[!ht]{0.49\linewidth}
        \centering
        \includegraphics[width=\linewidth]{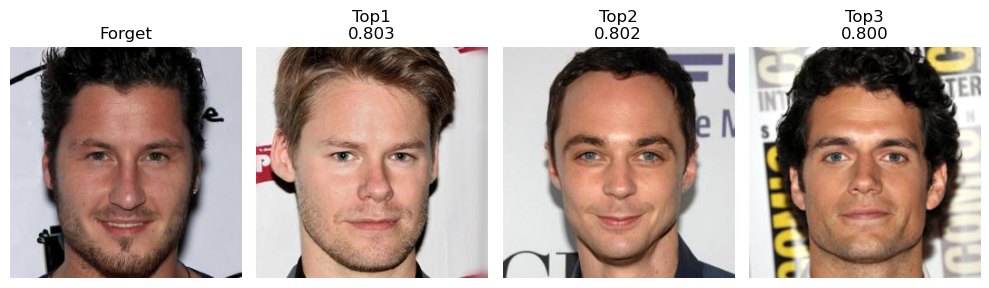}
    \end{subfigure}

    \begin{subfigure}[!ht]{0.49\linewidth}
        \centering
        \includegraphics[width=\linewidth]{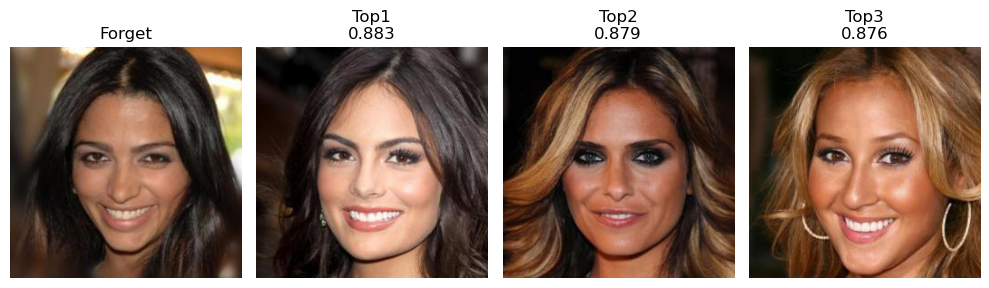}
    \end{subfigure}
    \hfill
    \begin{subfigure}[!ht]{0.49\linewidth}
        \centering
        \includegraphics[width=\linewidth]{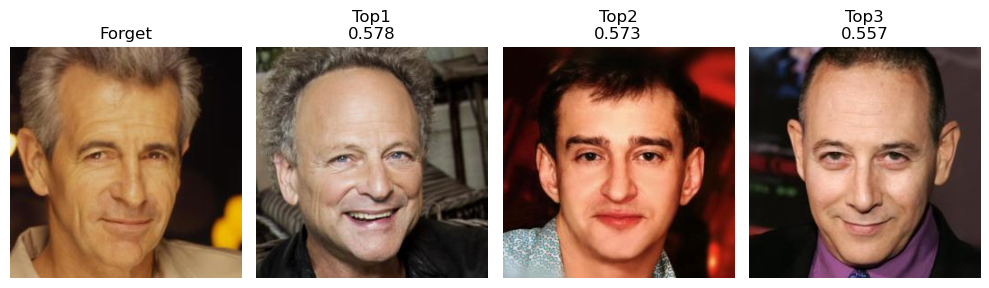}
    \end{subfigure}

    \begin{subfigure}[!ht]{0.49\linewidth}
        \centering
        \includegraphics[width=\linewidth]{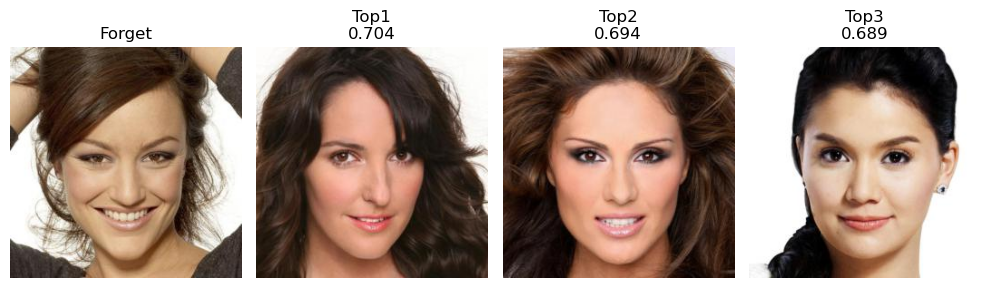}
    \end{subfigure}
    \hfill
    \begin{subfigure}[!ht]{0.49\linewidth}
        \centering
        \includegraphics[width=\linewidth]{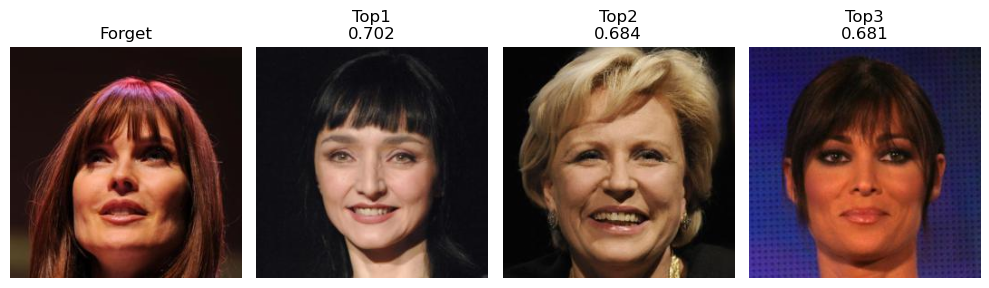}
    \end{subfigure}
    \caption{Forget samples and their nearest neighborhood on DINOv3 embedding \citep{simeoni2025dinov3}.}
    \label{fig:app_nn}
\end{figure}

\begin{figure}[!ht]
    \centering
    \begin{minipage}[!ht]{0.38\linewidth}
        \centering
        \includegraphics[width=\linewidth]{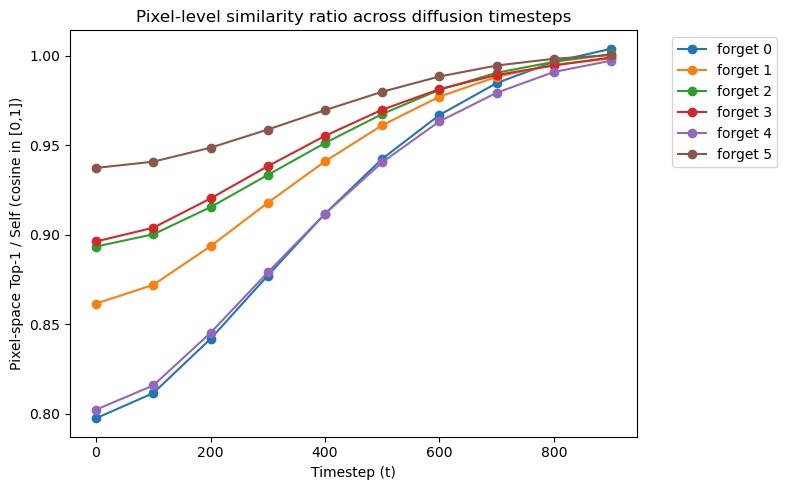}
        \caption{Similarity in pixel-level.}
        \label{fig:pixel}
    \end{minipage}
    \hfill
    \begin{minipage}[!ht]{0.6\linewidth}
        \centering
        \includegraphics[width=\linewidth]{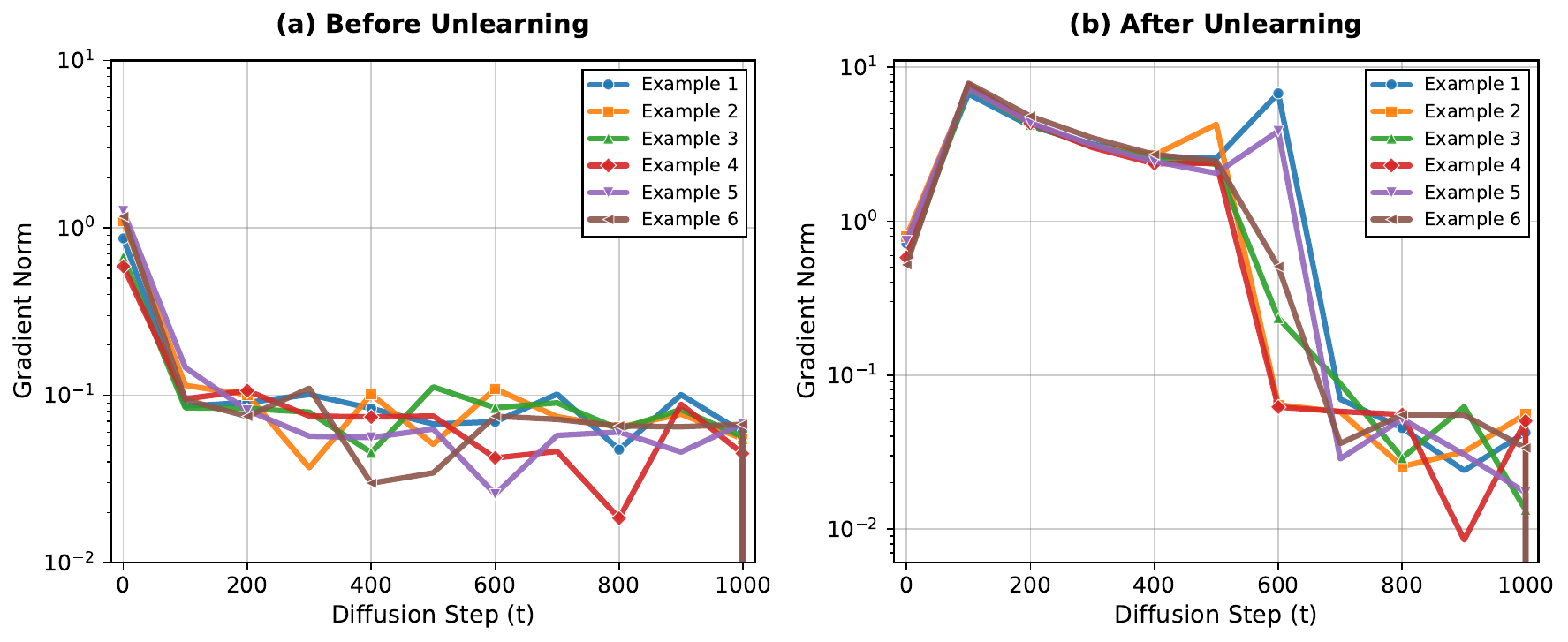}
        \caption{Gradient norms for each data sample before and after unlearning.}
        \label{fig:gradient_norms_supp}
    \end{minipage}
\end{figure}

\begin{figure*}[!ht]
    \centering
    \begin{subfigure}[!ht]{0.32\linewidth}
        \centering
        \includegraphics[width=\linewidth]{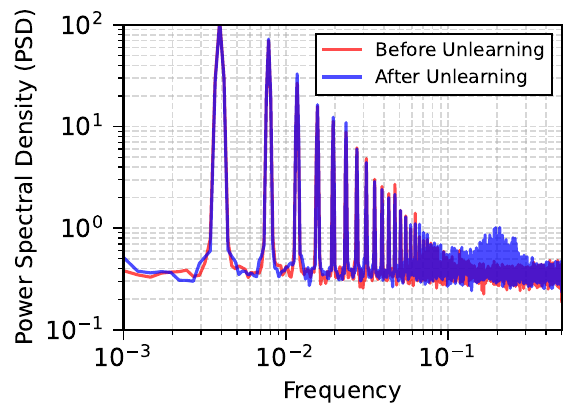}\vspace{-2mm}
    \end{subfigure}
    \begin{subfigure}[!ht]{0.32\linewidth}
        \centering
        \includegraphics[width=\linewidth]{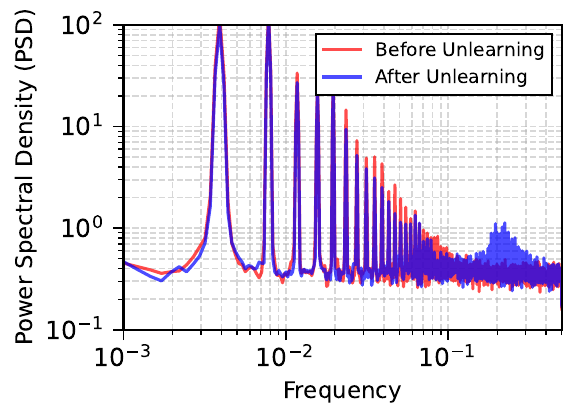}\vspace{-2mm}
    \end{subfigure}
    \hfill
    \begin{subfigure}[!ht]{0.32\linewidth}
        \centering
        \includegraphics[width=\linewidth]{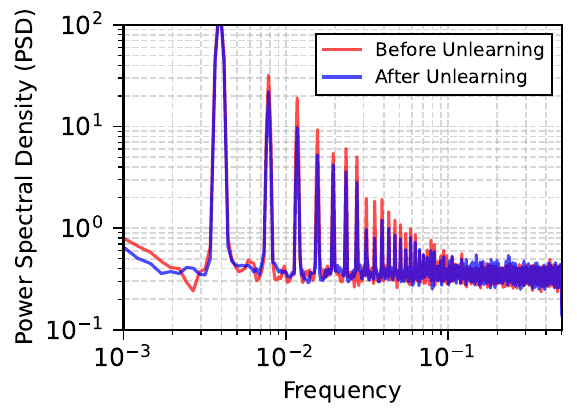}\vspace{-2mm}
    \end{subfigure}
    \hfill
    \caption{Additional results of Power Spectral Density (PSD) before and after unlearning on the forget dataset. As shown in the right figure, when the generated data from the unlearned model preserves aesthetic quality, the difference in high-frequency components is diminished.}
        \label{fig:psa2}
\end{figure*}

\begin{figure*}[!ht]
    \centering
    \begin{subfigure}[!ht]{0.32\linewidth}
        \centering
        \includegraphics[width=\linewidth]{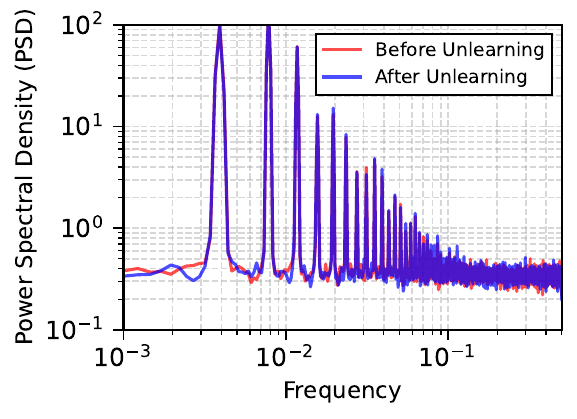}\vspace{-2mm}
    \end{subfigure}
    \begin{subfigure}[!ht]{0.32\linewidth}
        \centering
        \includegraphics[width=\linewidth]{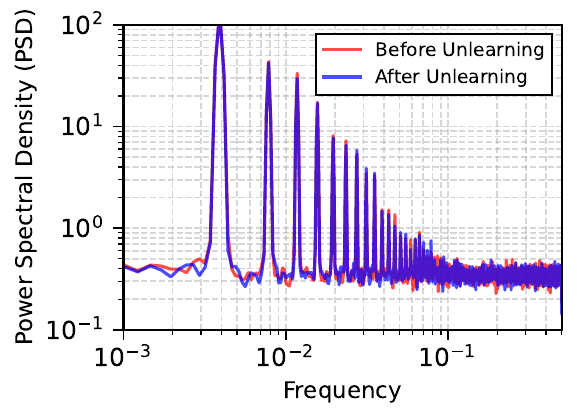}\vspace{-2mm}

    \end{subfigure}
    \hfill
    \begin{subfigure}[!ht]{0.32\linewidth}
        \centering
        \includegraphics[width=\linewidth]{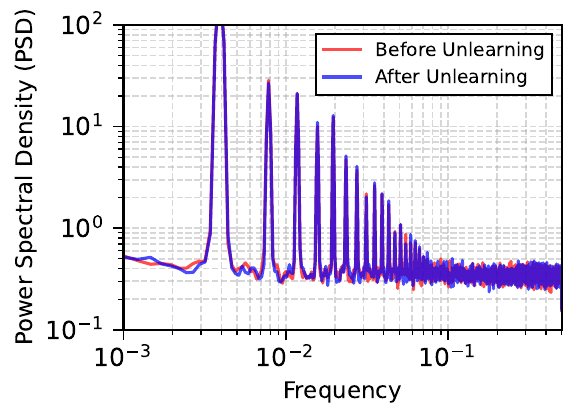}\vspace{-2mm}
    \end{subfigure}
    \hfill
    \caption{Additional results of Power Spectral Density (PSD) before and after unlearning on the retain dataset.}
        \label{fig:psa3}
\end{figure*}

\newpage

\section{Visualization}
    \label{app:vis}
    
    \paragraph{CelebA-HQ Results.} We first visualize the results for each method: SISS, SISS+Ours, and variational results on different time steps in Figures \ref{img1} to \ref{img5}.
    
    \begin{figure}[!ht]
        \centering \vspace{-2mm}
            \includegraphics[width=\linewidth]{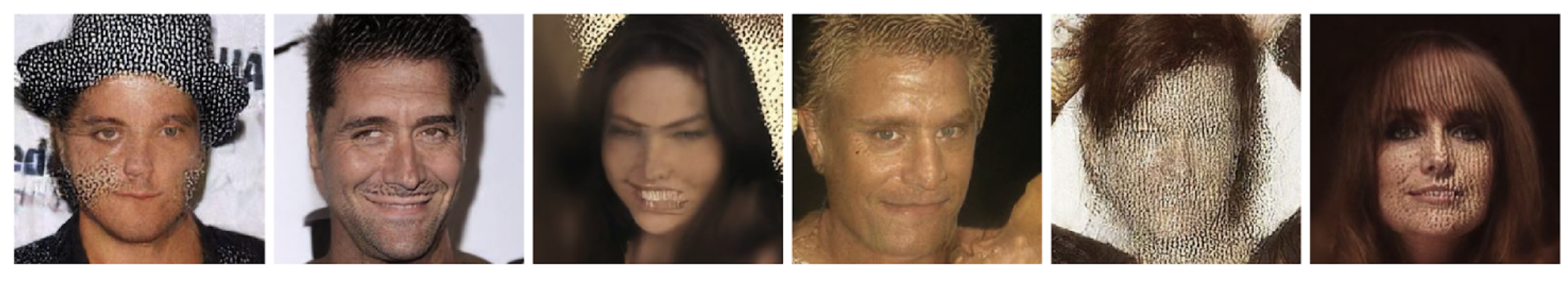}\vspace{-2mm}
            \caption{Unlearning with SISS.}
            \label{img1}
    \end{figure}
    
    \begin{figure}[!ht]
        \centering
        \begin{minipage}[b]{0.163\linewidth}
            \includegraphics[width=\linewidth]{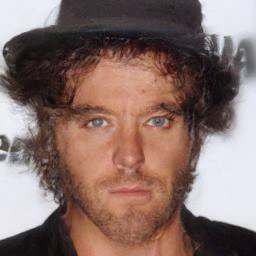}
        \end{minipage}%
        \begin{minipage}[b]{0.163\linewidth}
            \includegraphics[width=\linewidth]{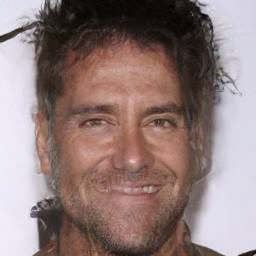}
        \end{minipage}%
        \begin{minipage}[b]{0.163\linewidth}
            \includegraphics[width=\linewidth]{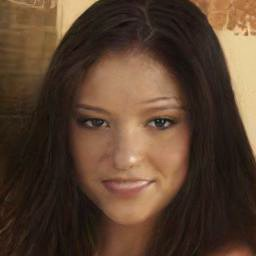}
        \end{minipage}%
        \begin{minipage}[b]{0.163\linewidth}
            \includegraphics[width=\linewidth]{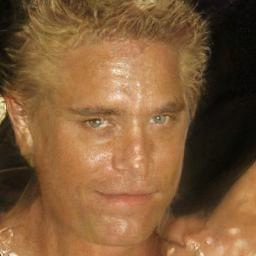}
        \end{minipage}%
        \begin{minipage}[b]{0.163\linewidth}
            \includegraphics[width=\linewidth]{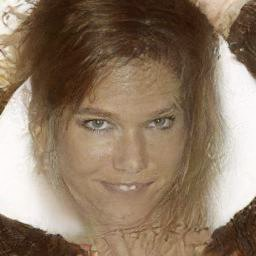}
        \end{minipage}%
        \begin{minipage}[b]{0.163\linewidth}
            \includegraphics[width=\linewidth]{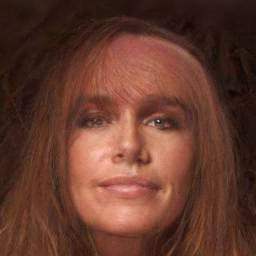}
        \end{minipage}%
        \caption{Unlearing with SISS+Ours.}
        \label{img2}
    
    \end{figure}
    
    \begin{figure}[!ht]
        \centering \vspace{-2mm}
            \includegraphics[width=\linewidth]{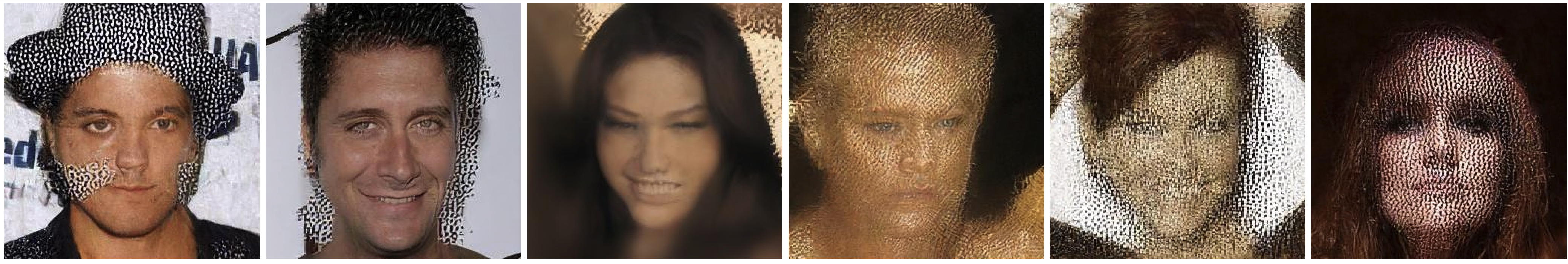}\vspace{-2mm}
            \caption{Unlearning on time steps [0, 500].}
        \label{img3}
    
    \end{figure}
    
    \begin{figure}[!ht]
        \centering \vspace{-2mm}
            \includegraphics[width=\linewidth]{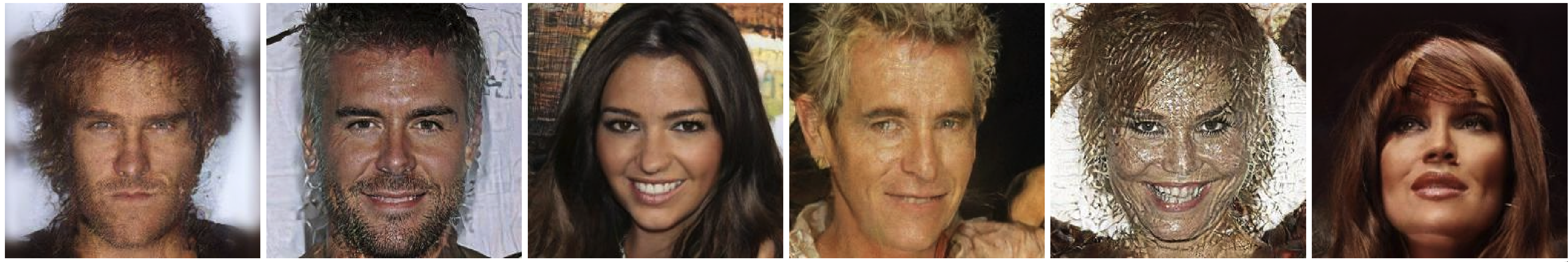}\vspace{-2mm}
            \caption{Unlearning on time steps [500, 1000].}
            \label{img4}
    
    \end{figure}

    \begin{figure}[!ht]
        \centering \vspace{-2mm}
            \includegraphics[width=\linewidth]{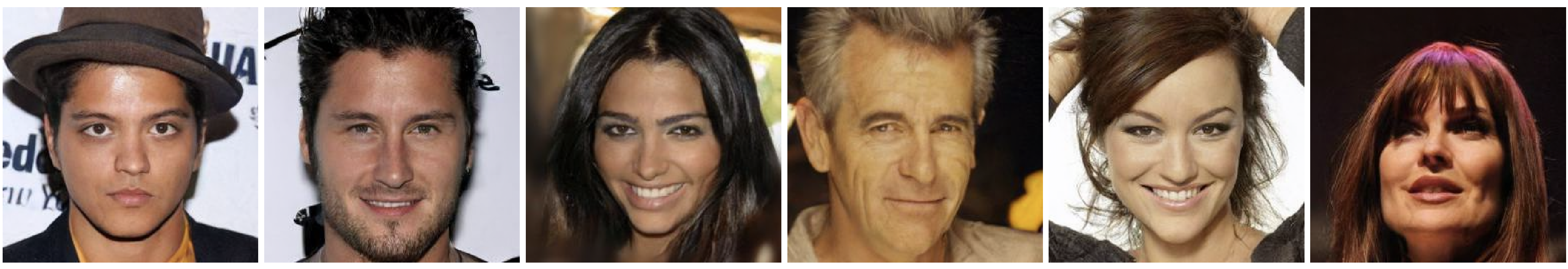}\vspace{-2mm}
            \caption{Unlearning on time steps [750, 1000].}
            \label{img5}
    
    \end{figure}
    
    \newpage
    \paragraph{Stable Diffusion Results.}
    We visualize the results of text-to-image data unlearning in Figures \ref{fig:sd1} to \ref{fig:sd3}.

    \begin{figure*}[!ht]
        \centering
        \begin{subfigure}[b]{0.189\textwidth}
            \centering
            \begin{minipage}[b]{\linewidth}
                \includegraphics[width=\linewidth]{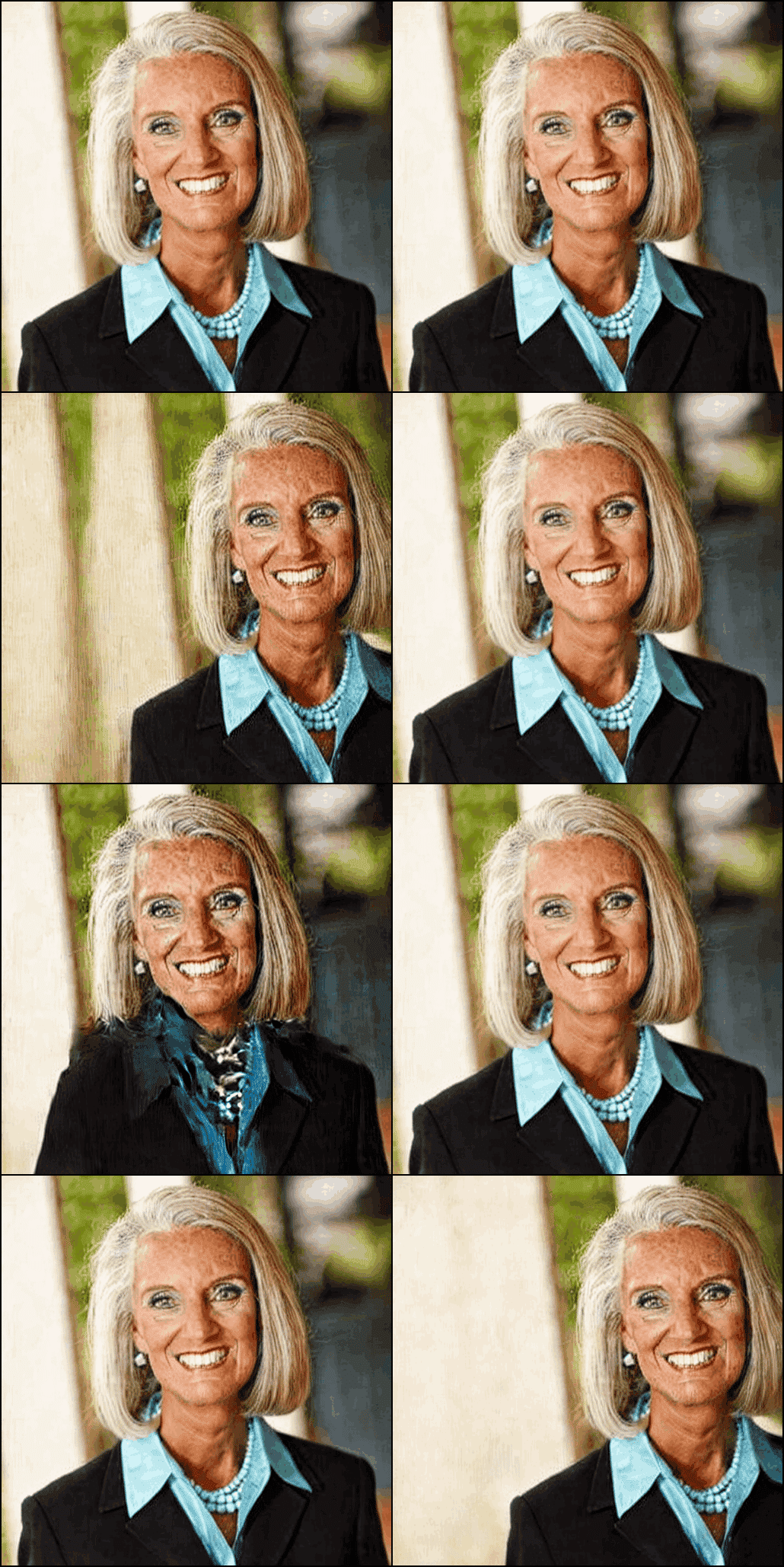}
            \end{minipage}%
            \caption{Memorized}
        \end{subfigure}
        \hfill
        \begin{subfigure}[b]{0.39\textwidth}
            \centering
            \begin{minipage}[b]{0.49\linewidth}
                \includegraphics[width=\linewidth]{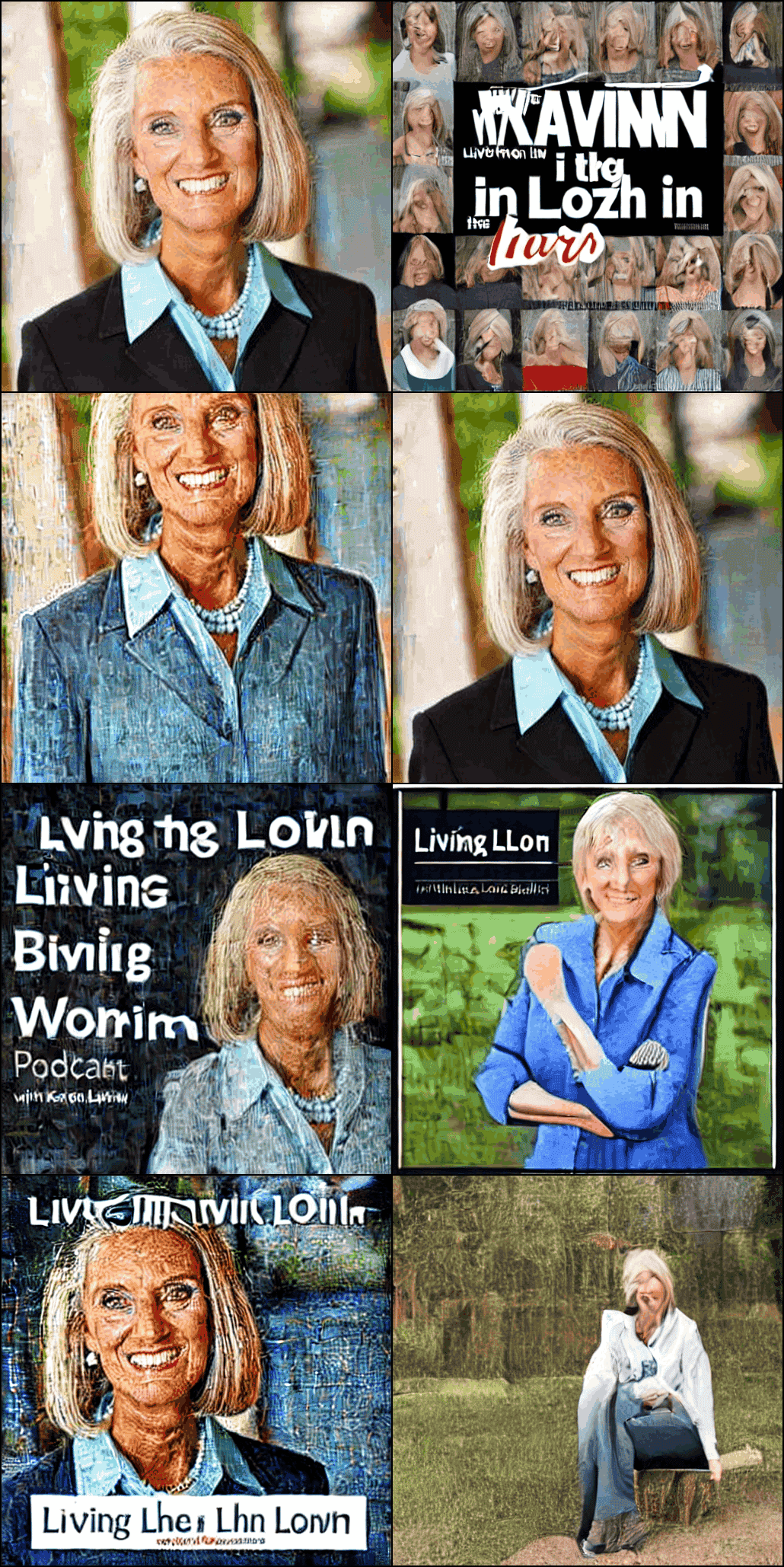}
            \end{minipage}%
            \hfill
            \begin{minipage}[b]{0.49\linewidth}
                \includegraphics[width=\linewidth]{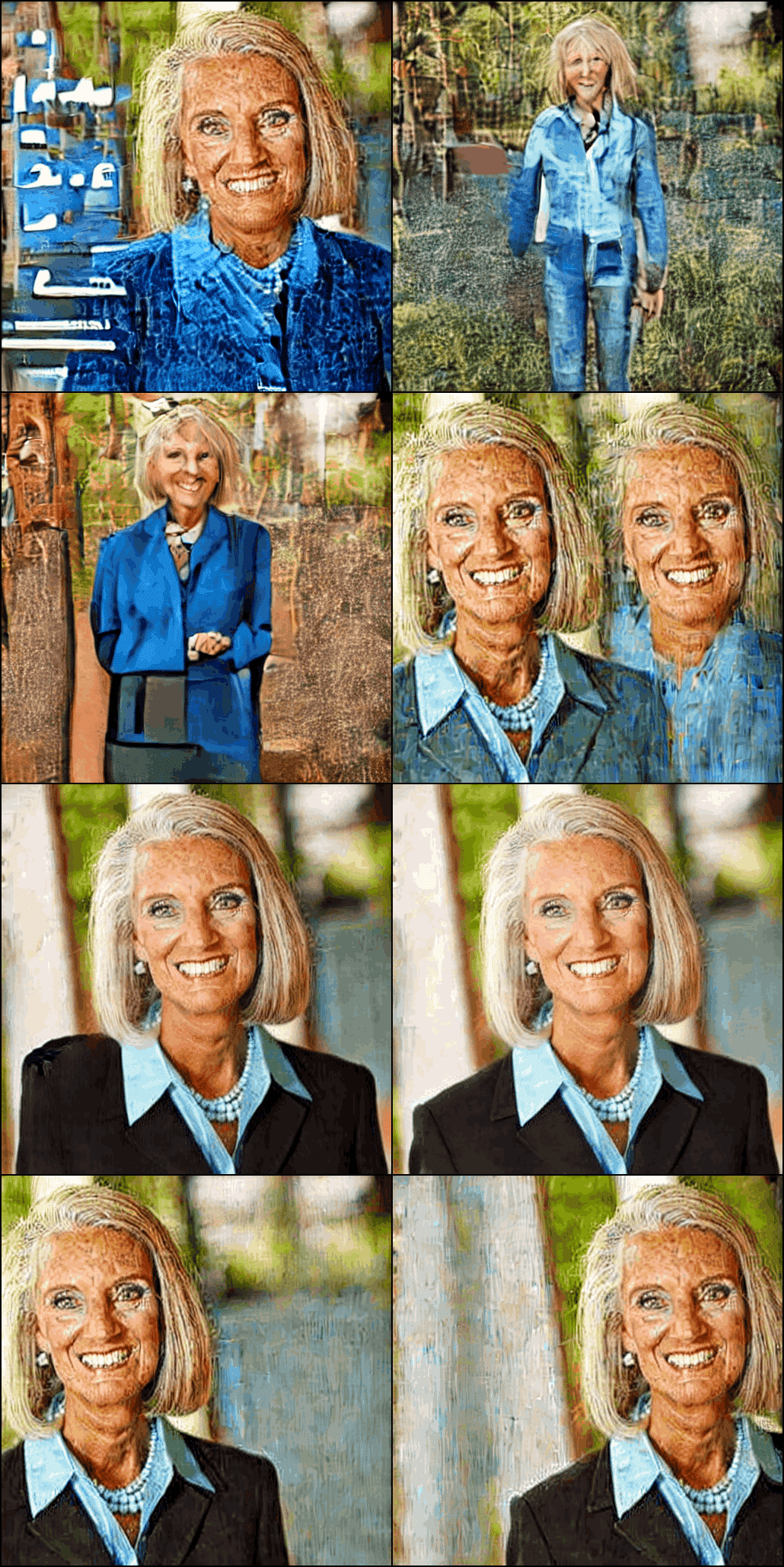}
            \end{minipage}%
            \hfill
            \caption{SISS}
        \end{subfigure}
        \begin{subfigure}[b]{0.39\textwidth}
        \centering
        \begin{minipage}[b]{0.49\linewidth}
            \includegraphics[width=\linewidth]{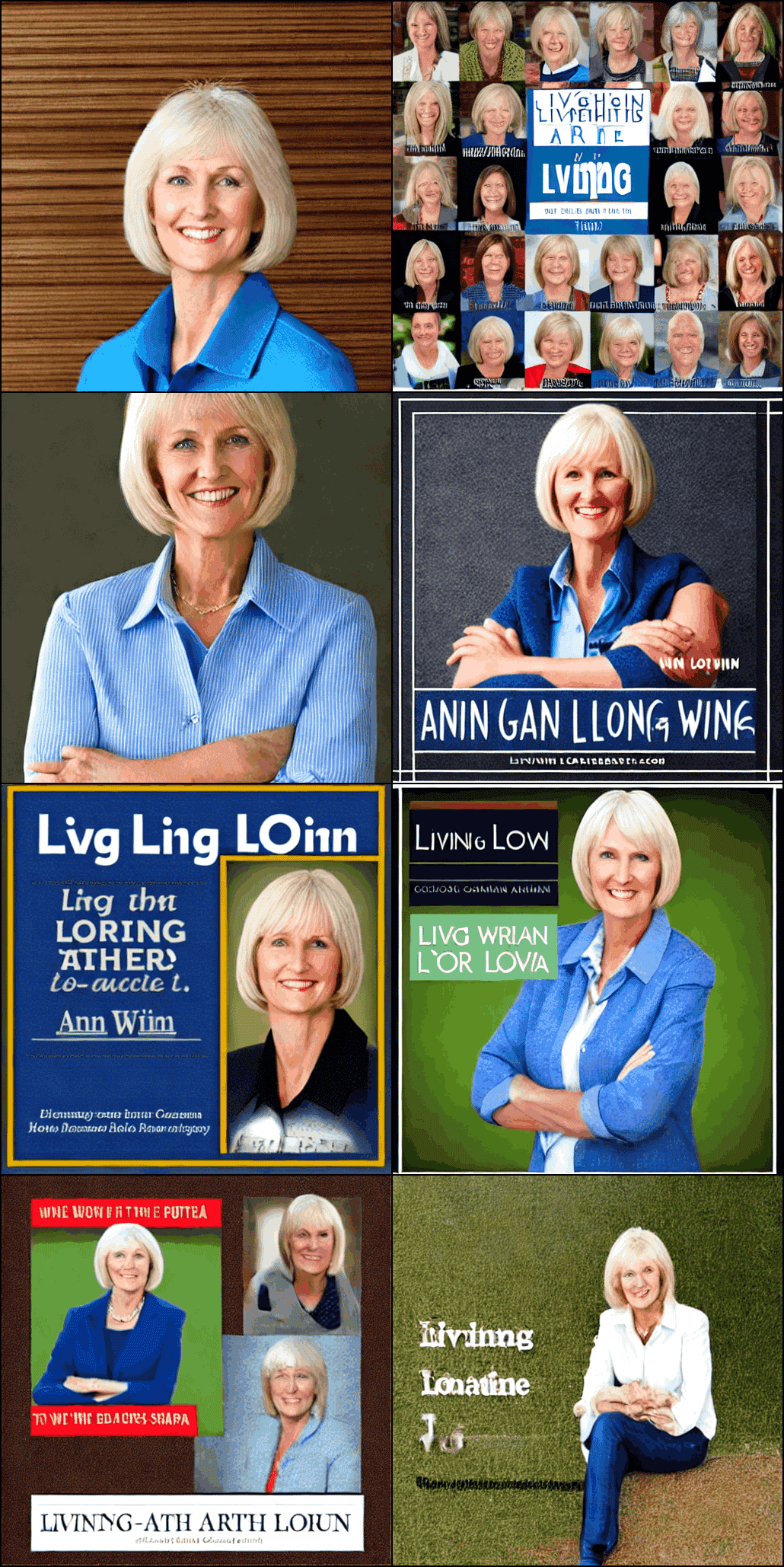}
        \end{minipage}%
        \hfill
        \begin{minipage}[b]{0.49\linewidth}
            \includegraphics[width=\linewidth]{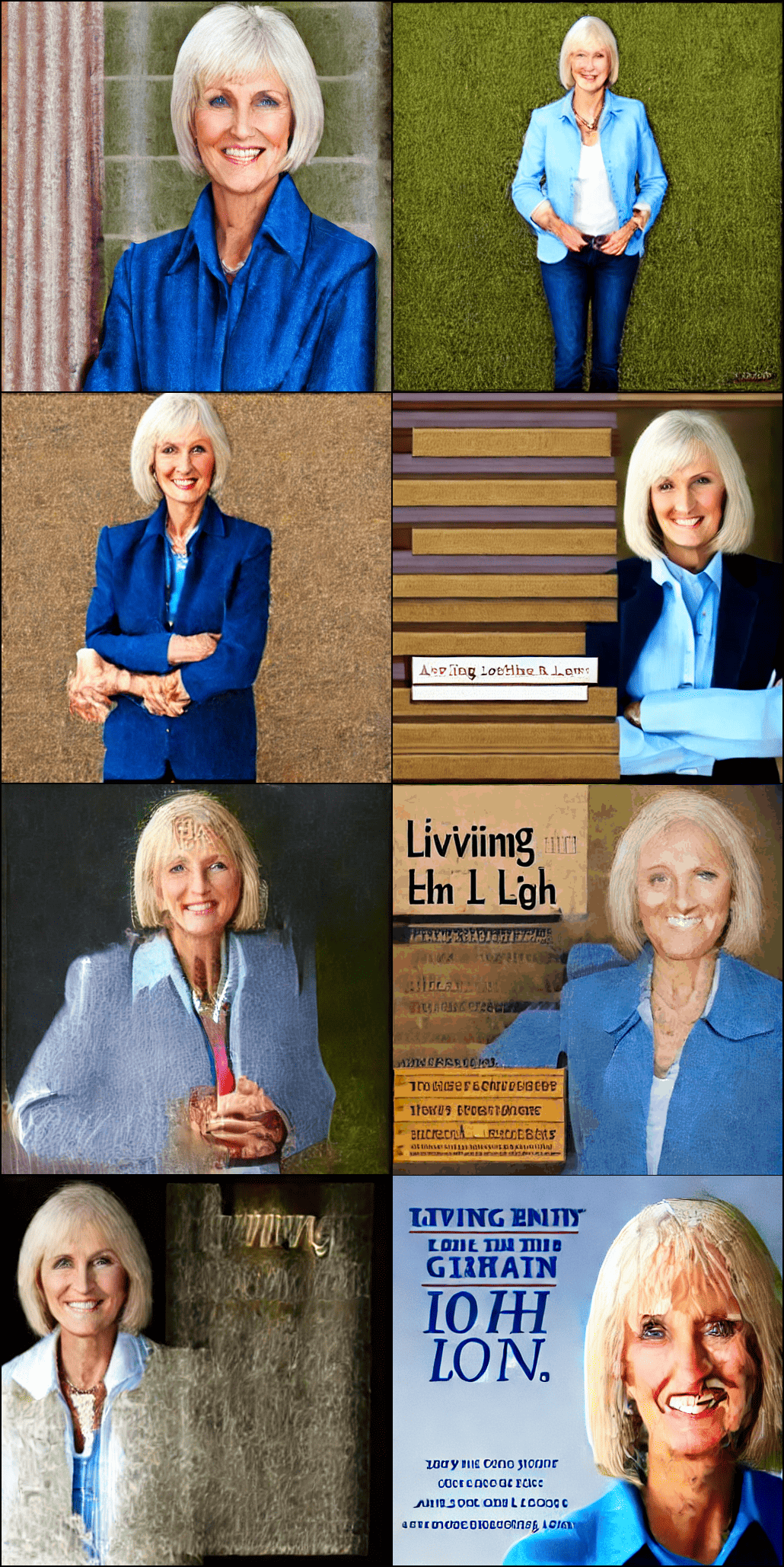}
        \end{minipage}%
        \hfill
        \caption{SISS + Ours}
    \end{subfigure}
        \caption{Visualization of (left) Partially-memorized and (right) Fully-memorized results after unlearning of the prompt ''Living in the Light with Ann Graham Lotz".}
        \label{fig:sd1}
    \end{figure*}
    
    \begin{figure*}[!ht]
        \centering
        \begin{subfigure}[b]{0.325\textwidth}
            \centering
                \begin{minipage}[b]{0.49\linewidth}
                \includegraphics[width=\linewidth]{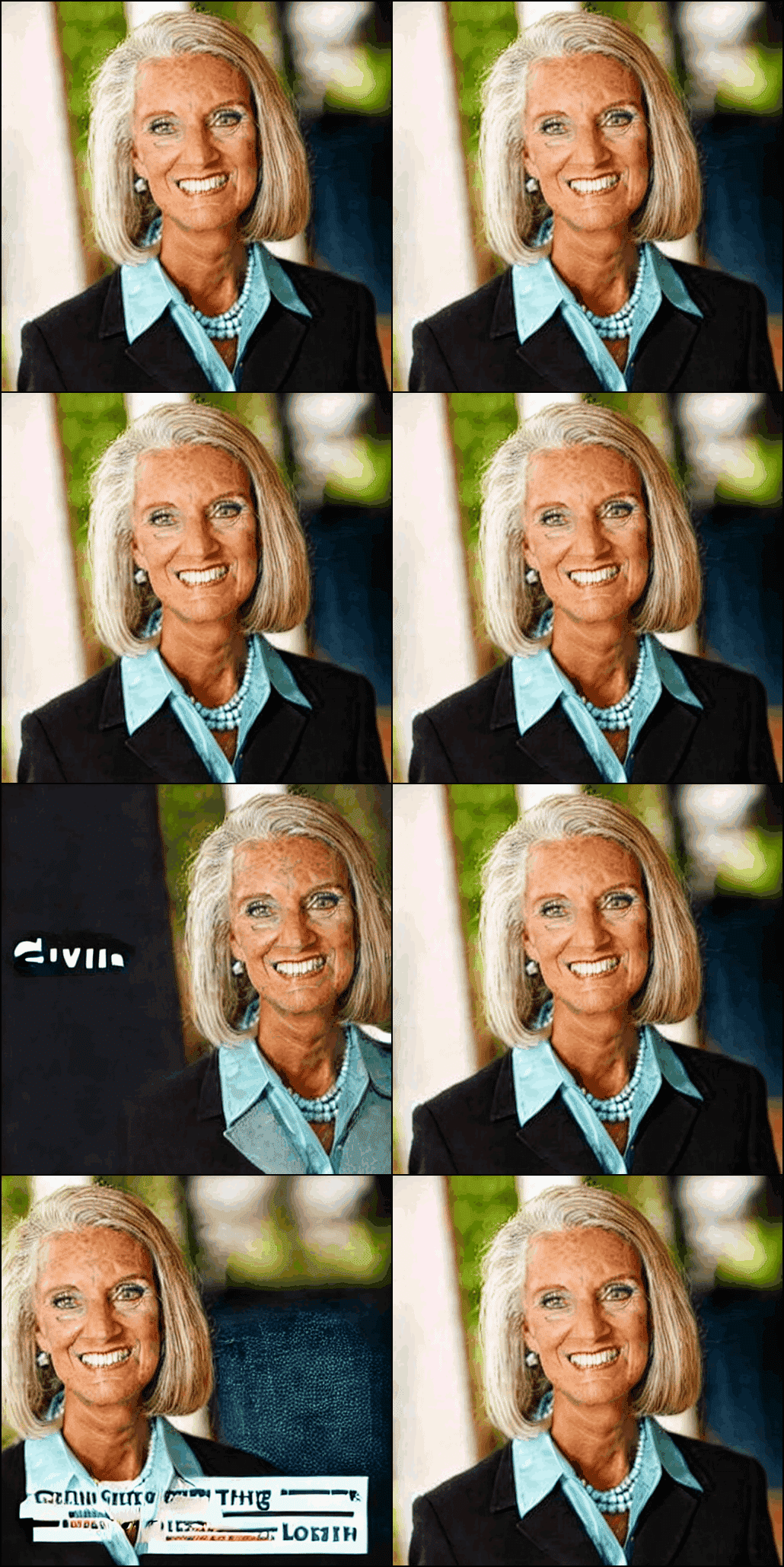}
            \end{minipage}%
            \begin{minipage}[b]{0.49\linewidth}
                \includegraphics[width=\linewidth]{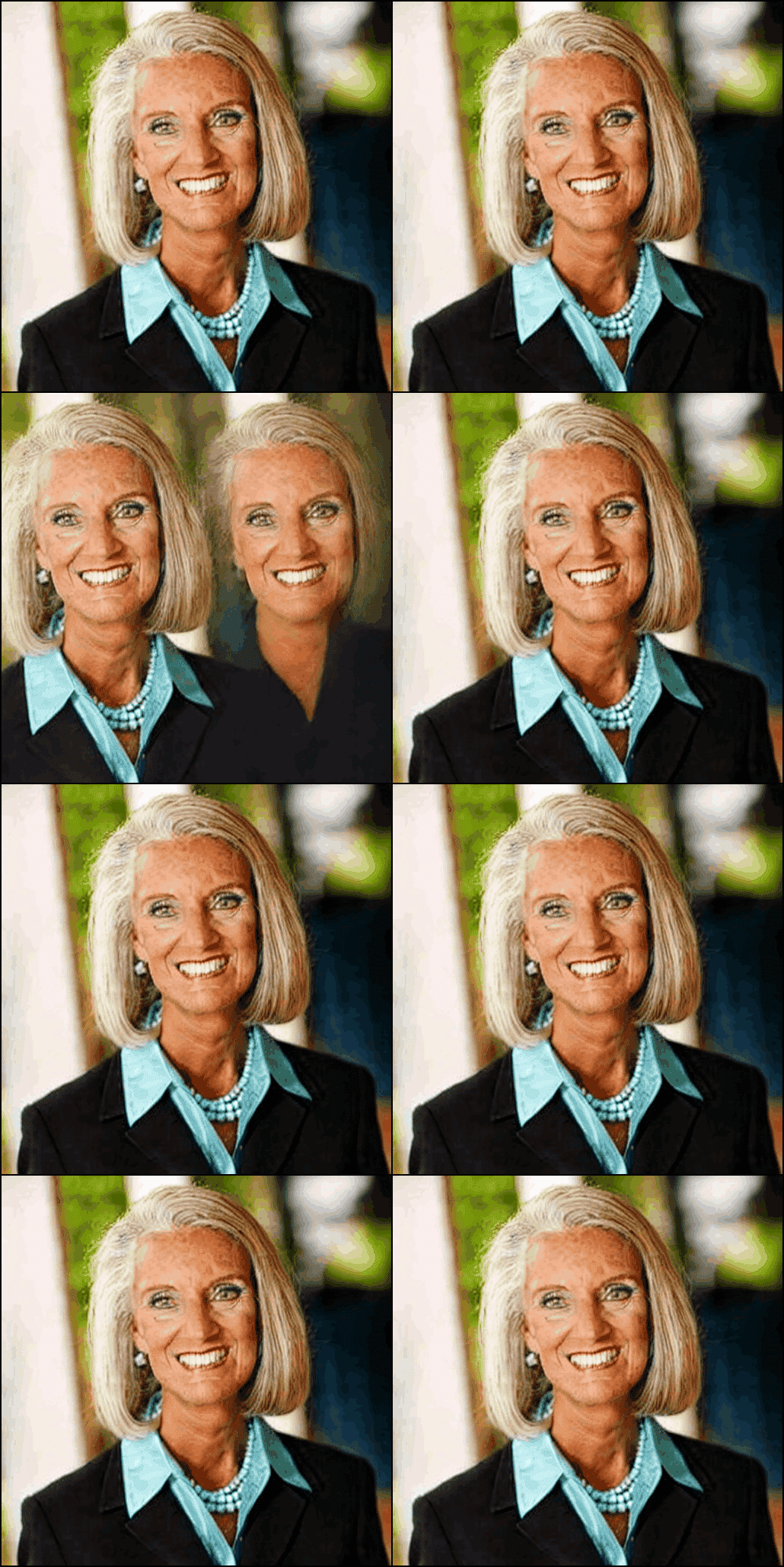}
            \end{minipage}%
            \caption{Time steps on [0,500]}
        \end{subfigure}
        \begin{subfigure}[b]{0.325\textwidth}
            \centering
            \begin{minipage}[b]{0.49\linewidth}
                \includegraphics[width=\linewidth]{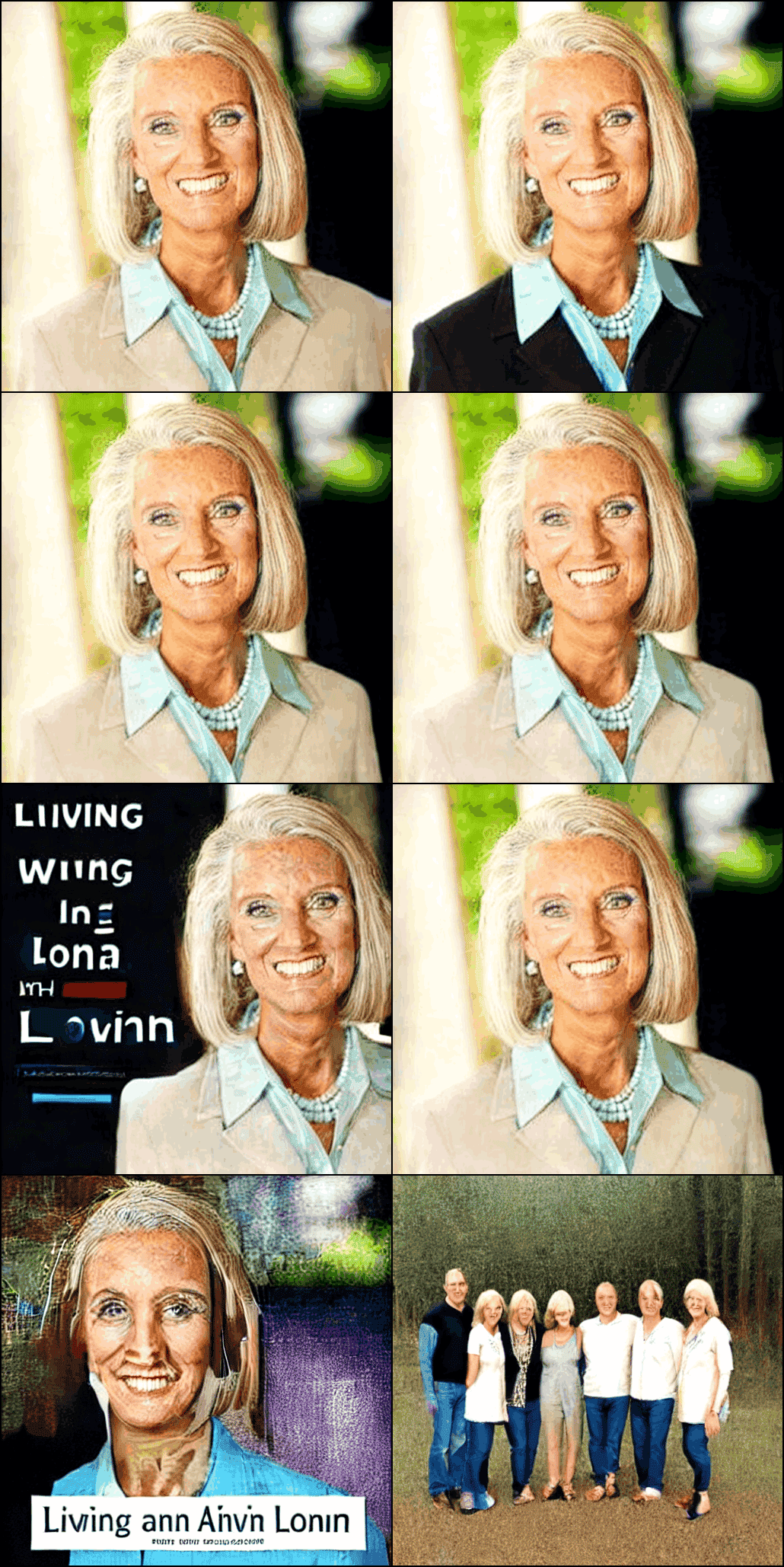}
            \end{minipage}%
            \begin{minipage}[b]{0.49\linewidth}
                \includegraphics[width=\linewidth]{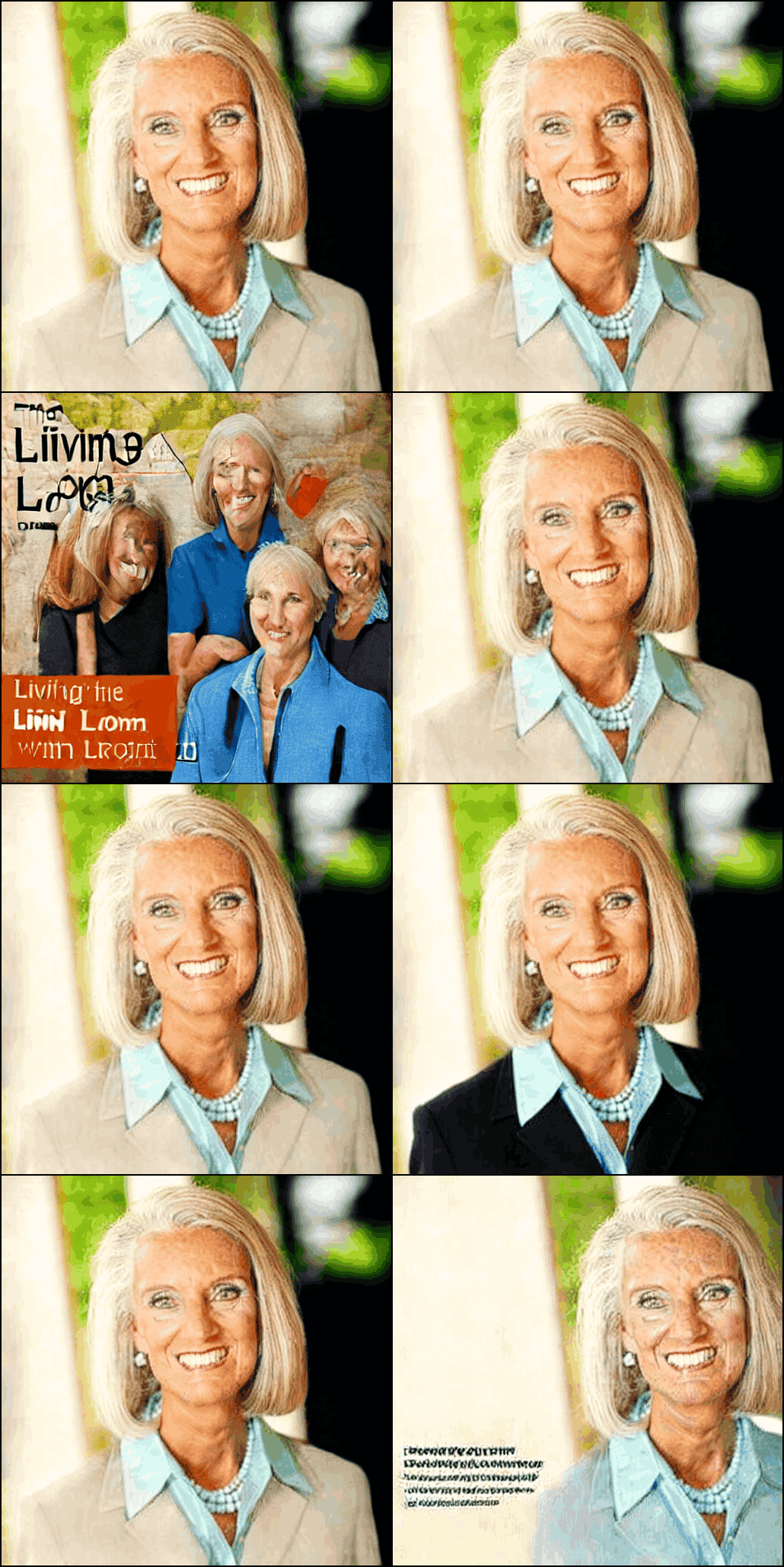}
            \end{minipage}%
            \caption{Time steps on [250,750]}
        \end{subfigure}
        \begin{subfigure}[b]{0.325\textwidth}
            \centering
            \begin{minipage}[b]{0.49\linewidth}
                \includegraphics[width=\linewidth]{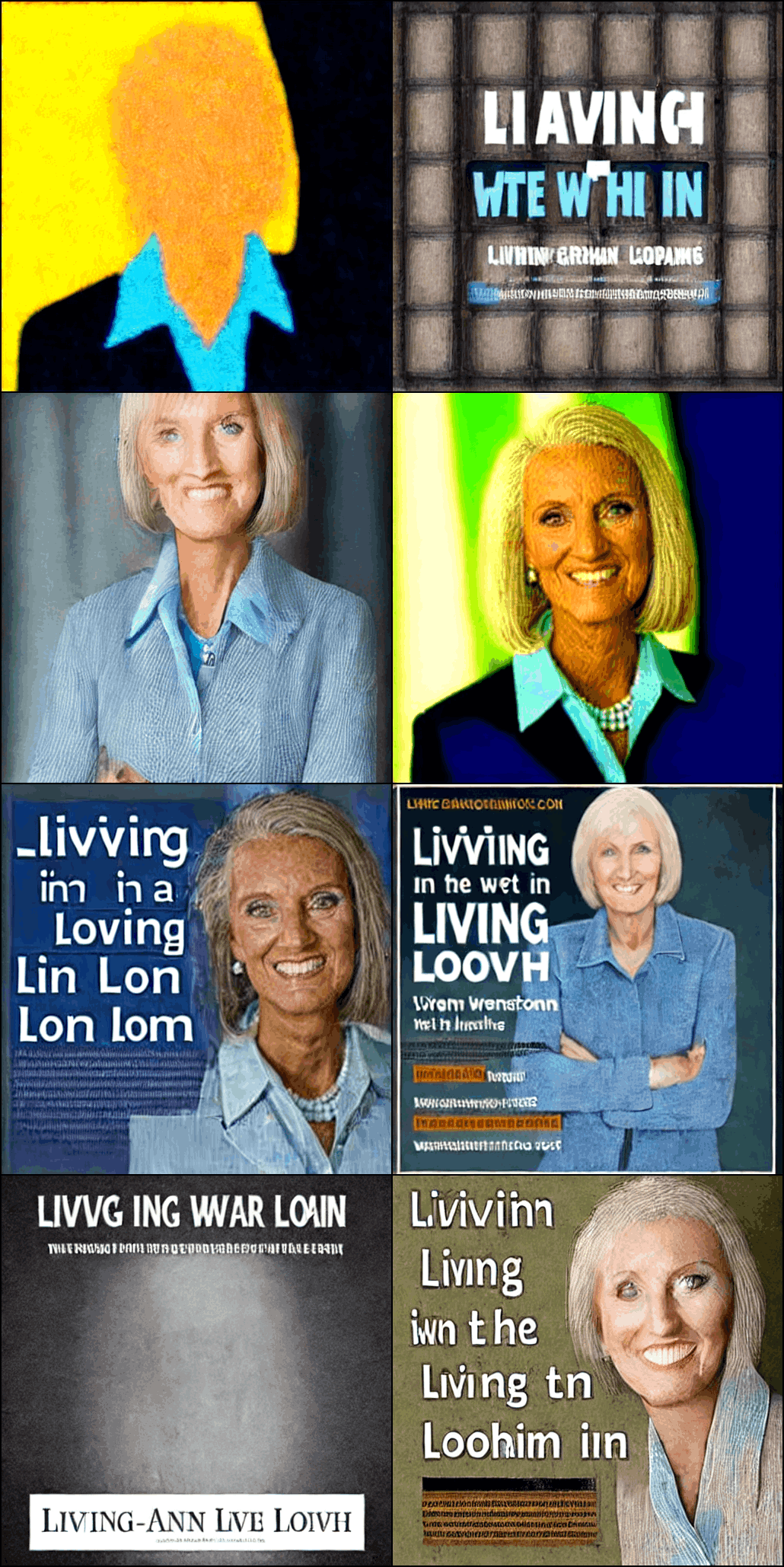}
            \end{minipage}%
            \begin{minipage}[b]{0.49\linewidth}
                \includegraphics[width=\linewidth]{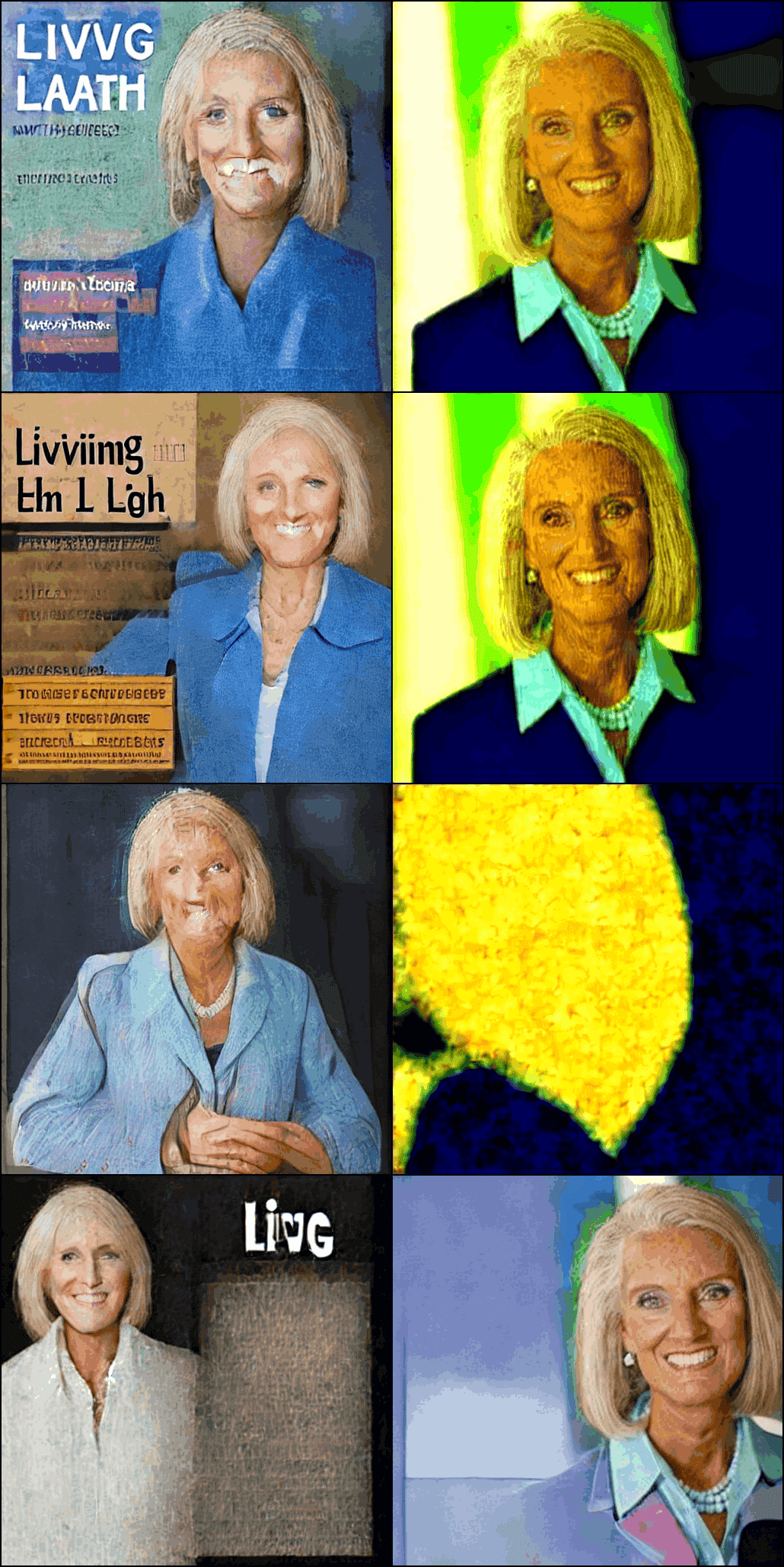}
            \end{minipage}%
            \caption{Time steps [500,1000]}
        \end{subfigure}
        \caption{Visualization of (left) Partially-memorized and (right) Fully-memorized results after unlearning of the prompt ''Living in the Light with Ann Graham Lotz" using different time steps for unlearning.}
        \label{fig:sd1-1}
    \end{figure*}

    \begin{figure*}[!ht]
        \centering
        \begin{subfigure}[b]{0.189\textwidth}
            \centering
            \begin{minipage}[b]{\linewidth}
                \includegraphics[width=\linewidth]{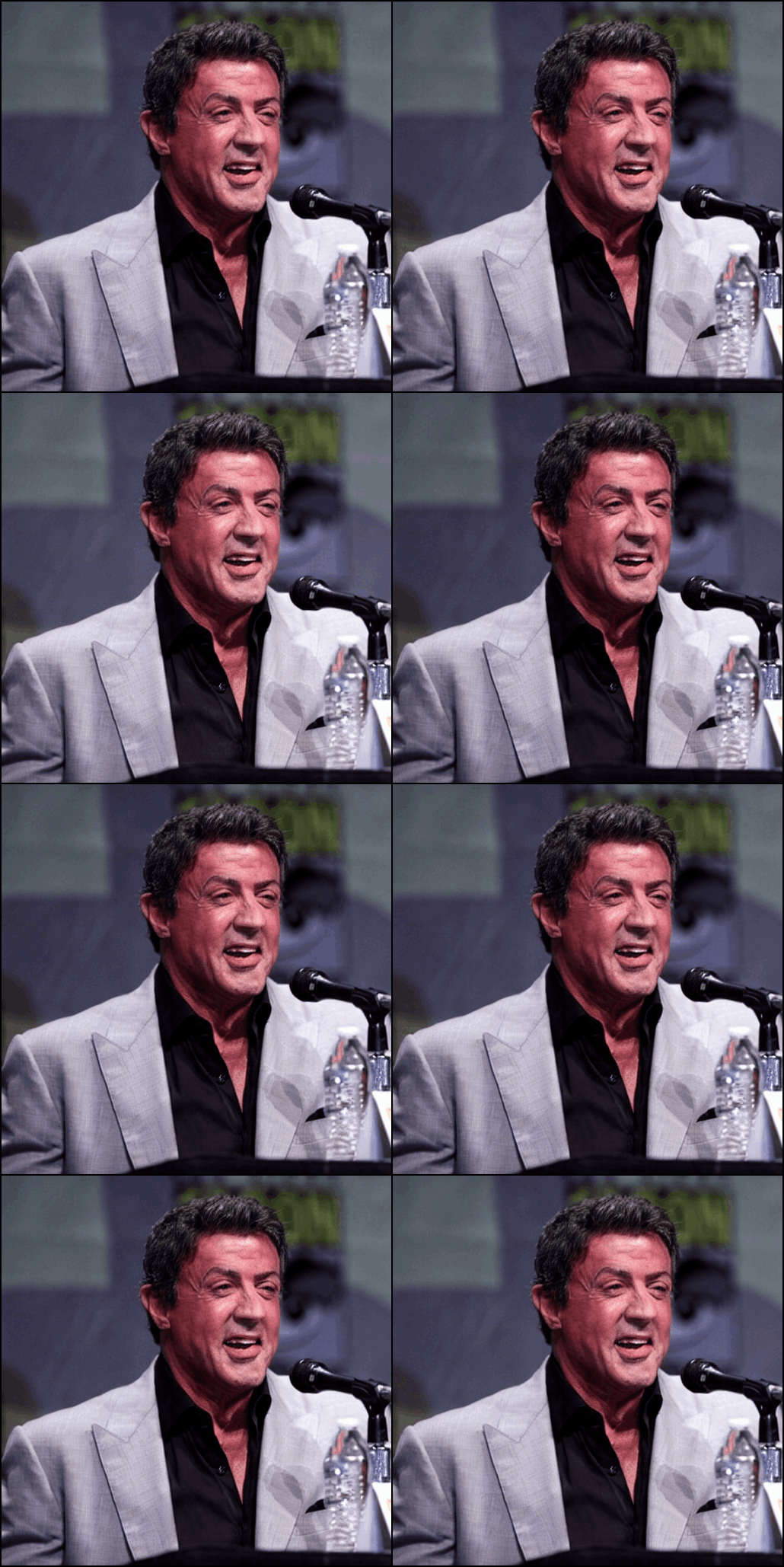}
            \end{minipage}%
            \caption{Memorized}
        \end{subfigure}
        \hfill
        \begin{subfigure}[b]{0.39\textwidth}
            \centering
            \begin{minipage}[b]{0.49\linewidth}
                \includegraphics[width=\linewidth]{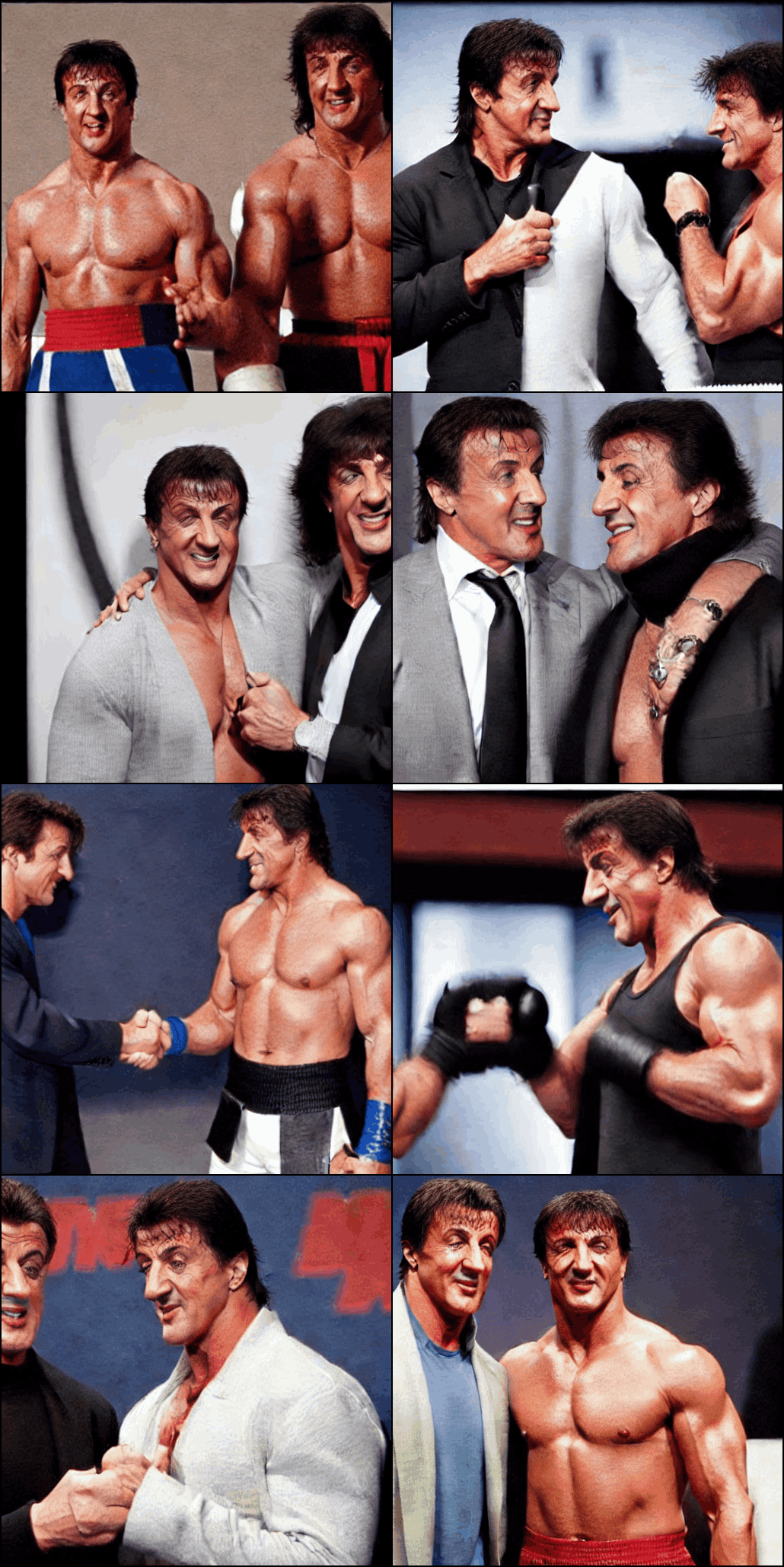}
            \end{minipage}%
            \hfill
            \begin{minipage}[b]{0.49\linewidth}
                \includegraphics[width=\linewidth]{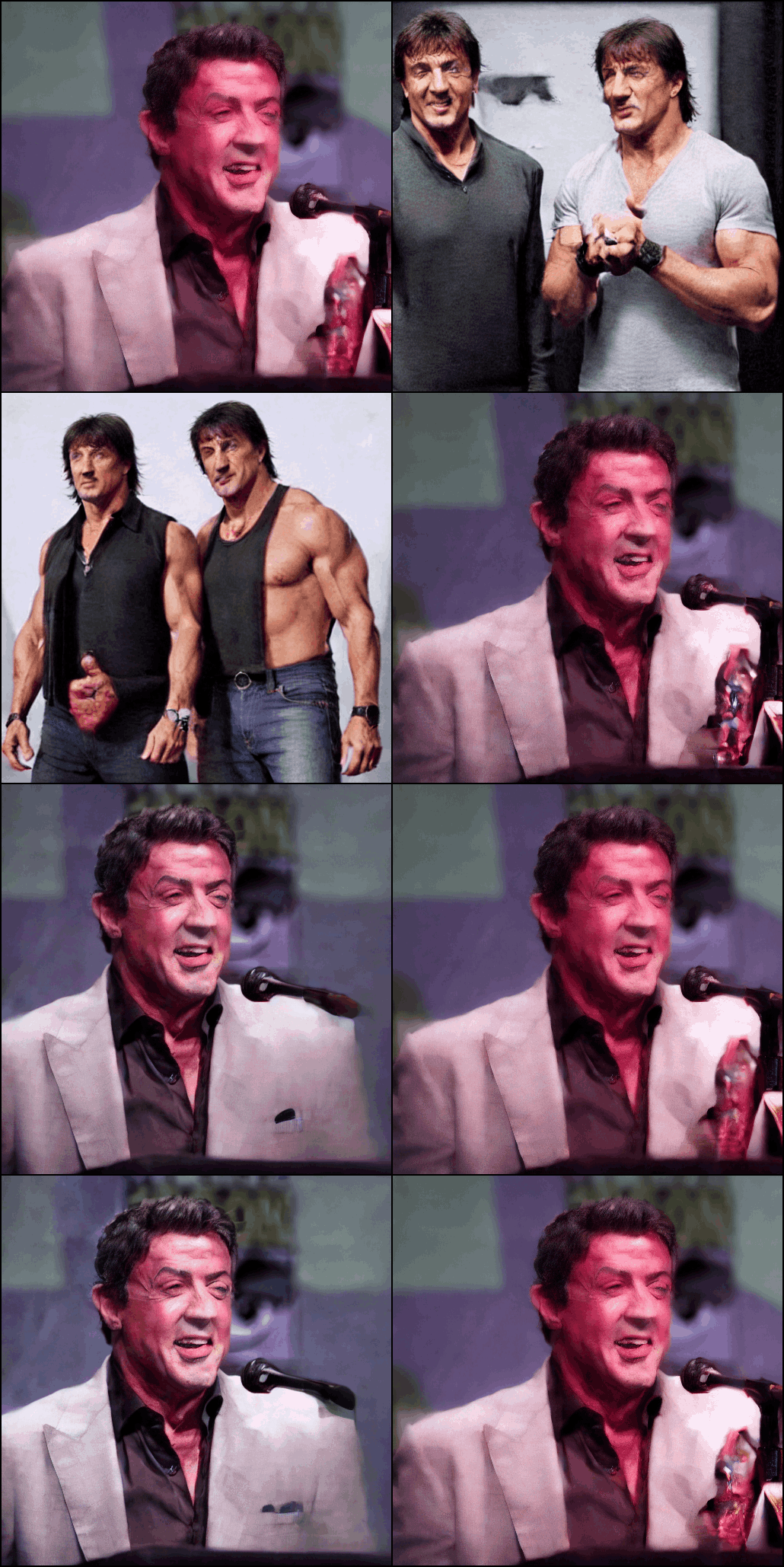}
            \end{minipage}%
            \hfill
            \caption{SISS}
        \end{subfigure}
        \begin{subfigure}[b]{0.39\textwidth}
        \centering
        \begin{minipage}[b]{0.49\linewidth}
            \includegraphics[width=\linewidth]{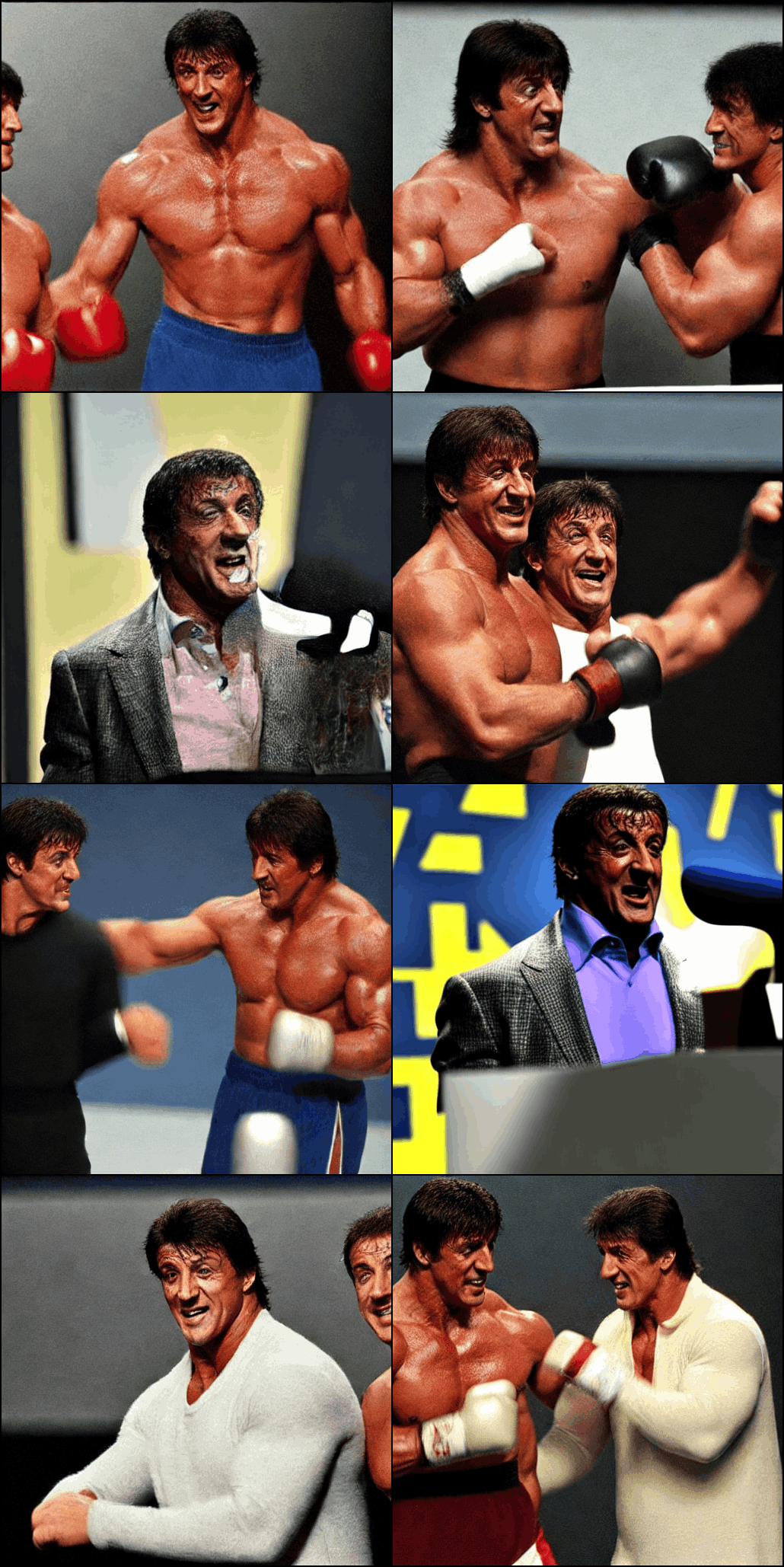}
        \end{minipage}%
        \hfill
        \begin{minipage}[b]{0.49\linewidth}
            \includegraphics[width=\linewidth]{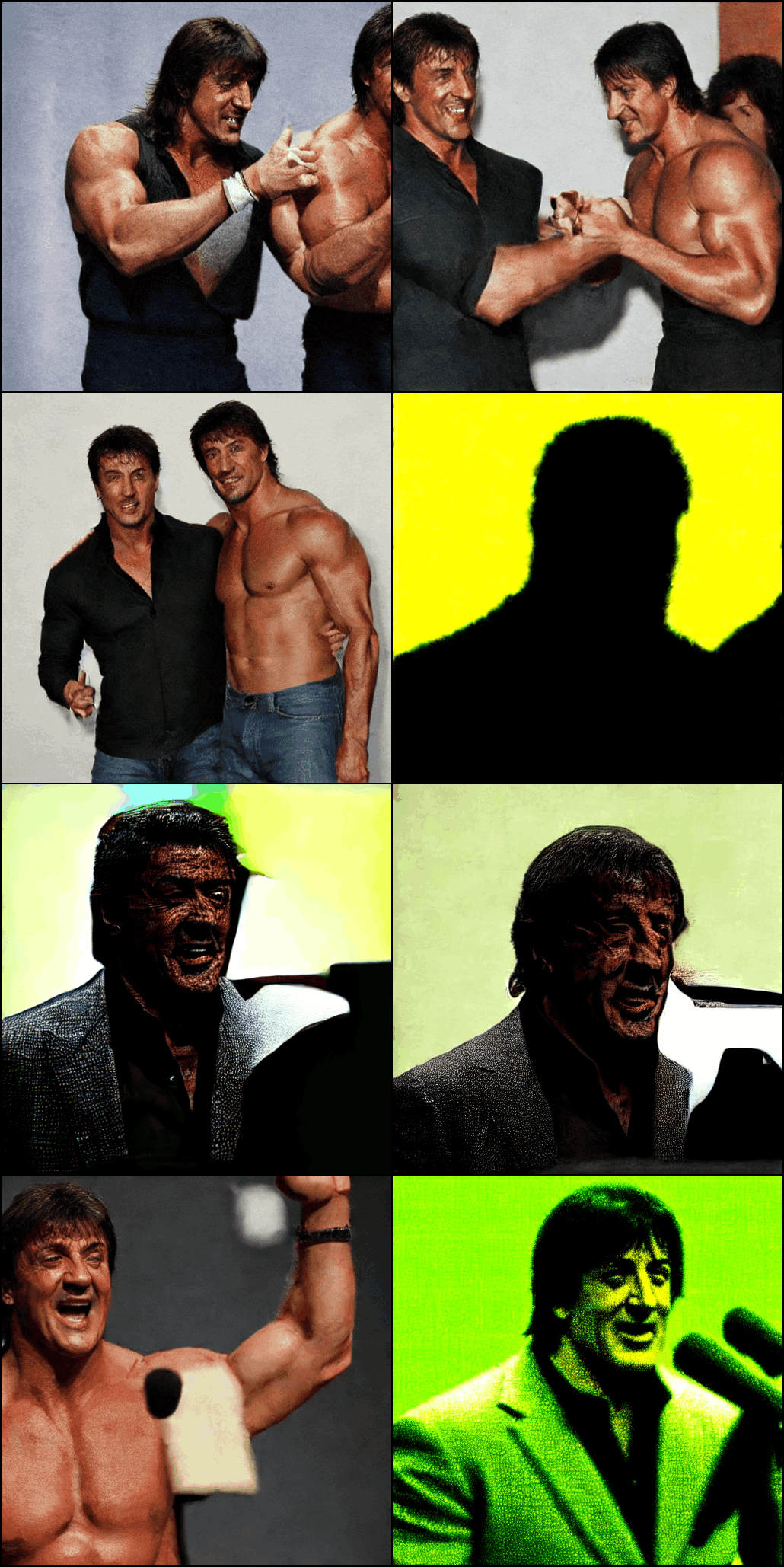}
        \end{minipage}%
        \hfill
        \caption{SISS + Ours}
    \end{subfigure}
        \caption{Visualization of (left) Partially-memorized and (right) Fully-memorized results after unlearning of the prompt ''Rambo 5 und Rocky Spin-Off - Sylvester Stallone gibt Updates".}
        \label{fig:sd2}
    \end{figure*}

    \begin{figure*}[!ht]
        \centering
        \begin{subfigure}[b]{0.189\textwidth}
            \centering
            \begin{minipage}[b]{\linewidth}
                \includegraphics[width=\linewidth]{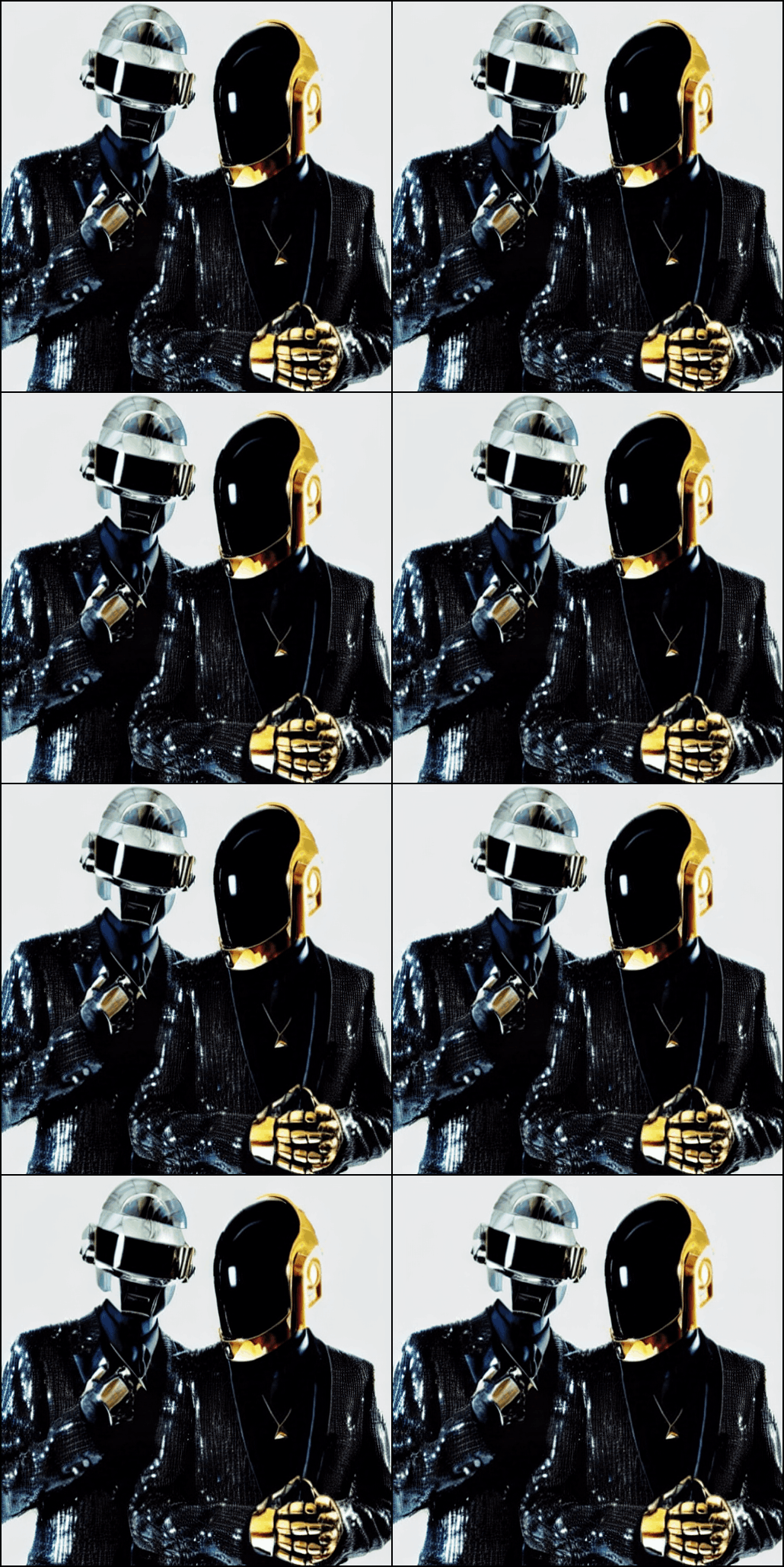}
            \end{minipage}%
            \caption{Memorized}
        \end{subfigure}
        \hfill
        \begin{subfigure}[b]{0.39\textwidth}
            \centering
            \begin{minipage}[b]{0.49\linewidth}
                \includegraphics[width=\linewidth]{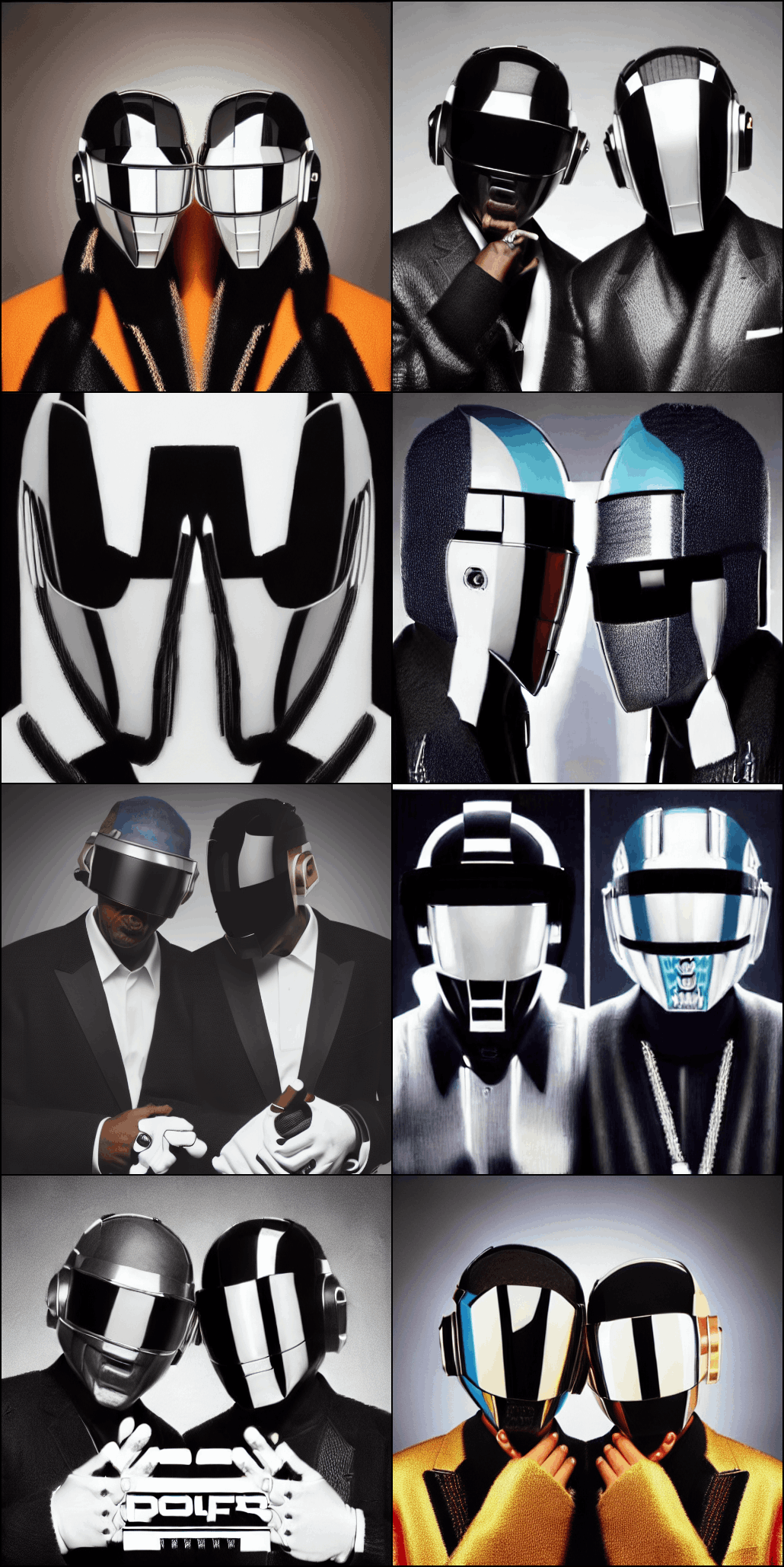}
            \end{minipage}%
            \hfill
            \begin{minipage}[b]{0.49\linewidth}
                \includegraphics[width=\linewidth]{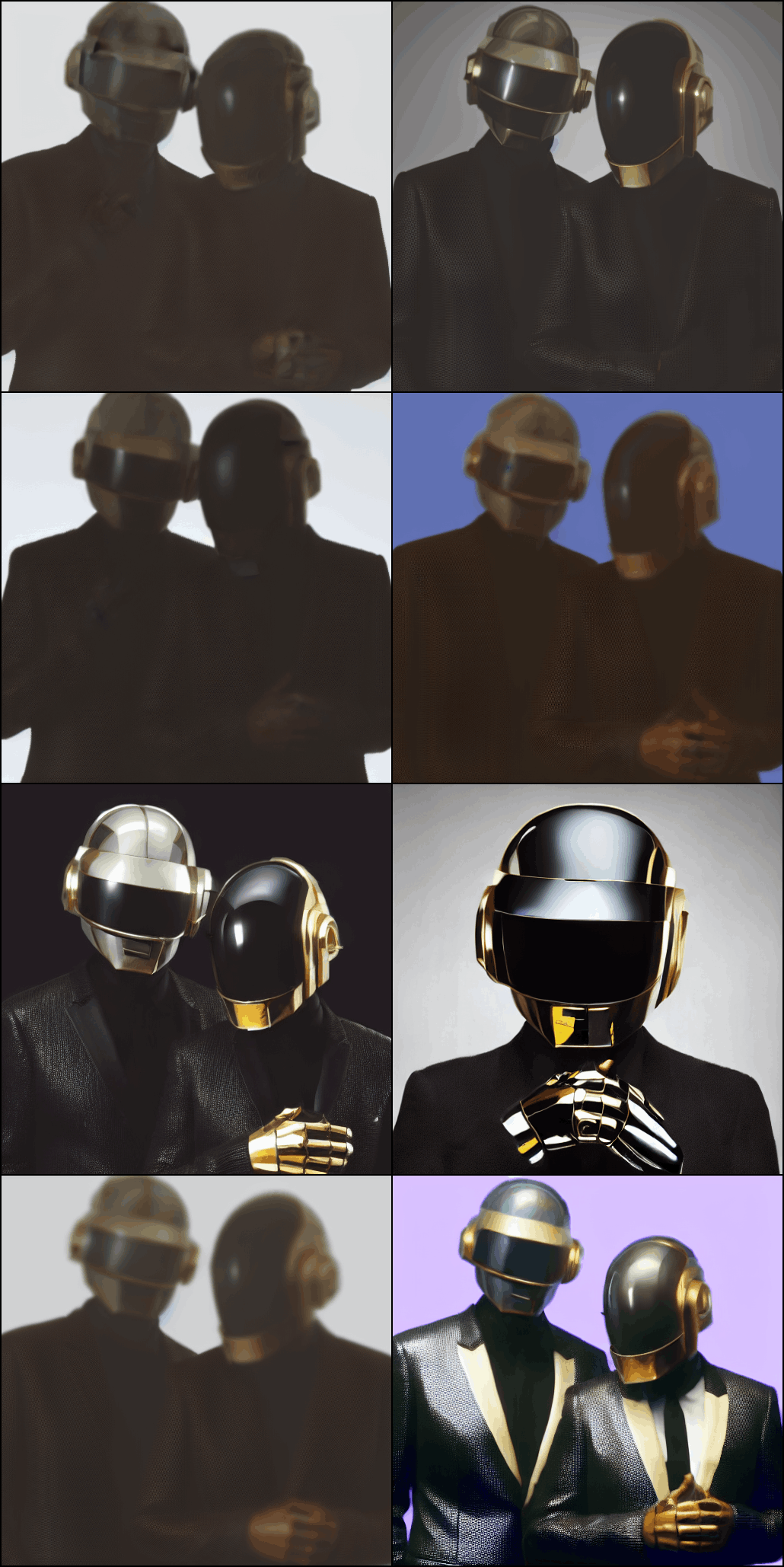}
            \end{minipage}%
            \hfill
            \caption{SISS}
        \end{subfigure}
        \begin{subfigure}[b]{0.39\textwidth}
        \centering
        \begin{minipage}[b]{0.49\linewidth}
            \includegraphics[width=\linewidth]{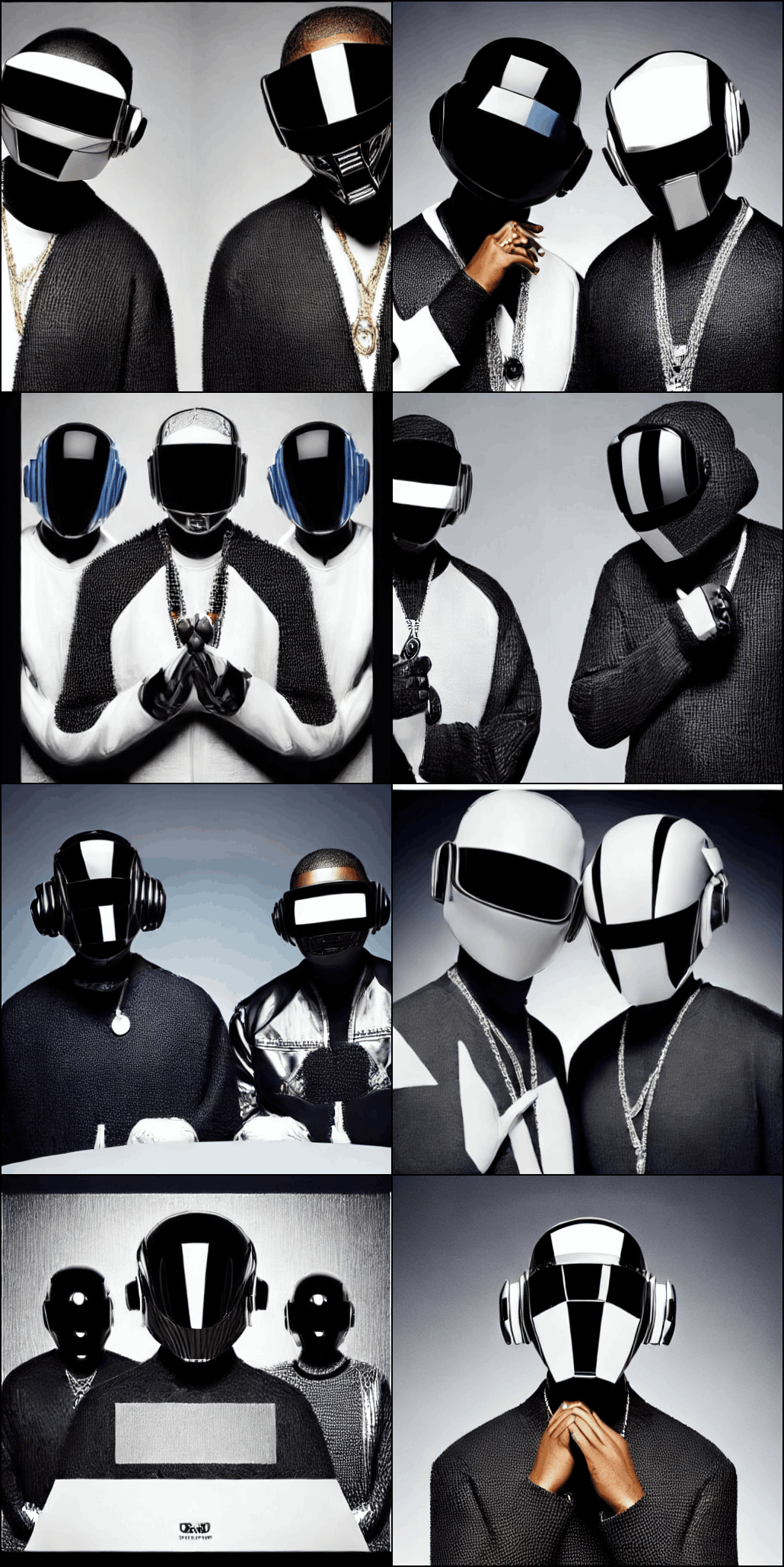}
        \end{minipage}%
        \hfill
        \begin{minipage}[b]{0.49\linewidth}
            \includegraphics[width=\linewidth]{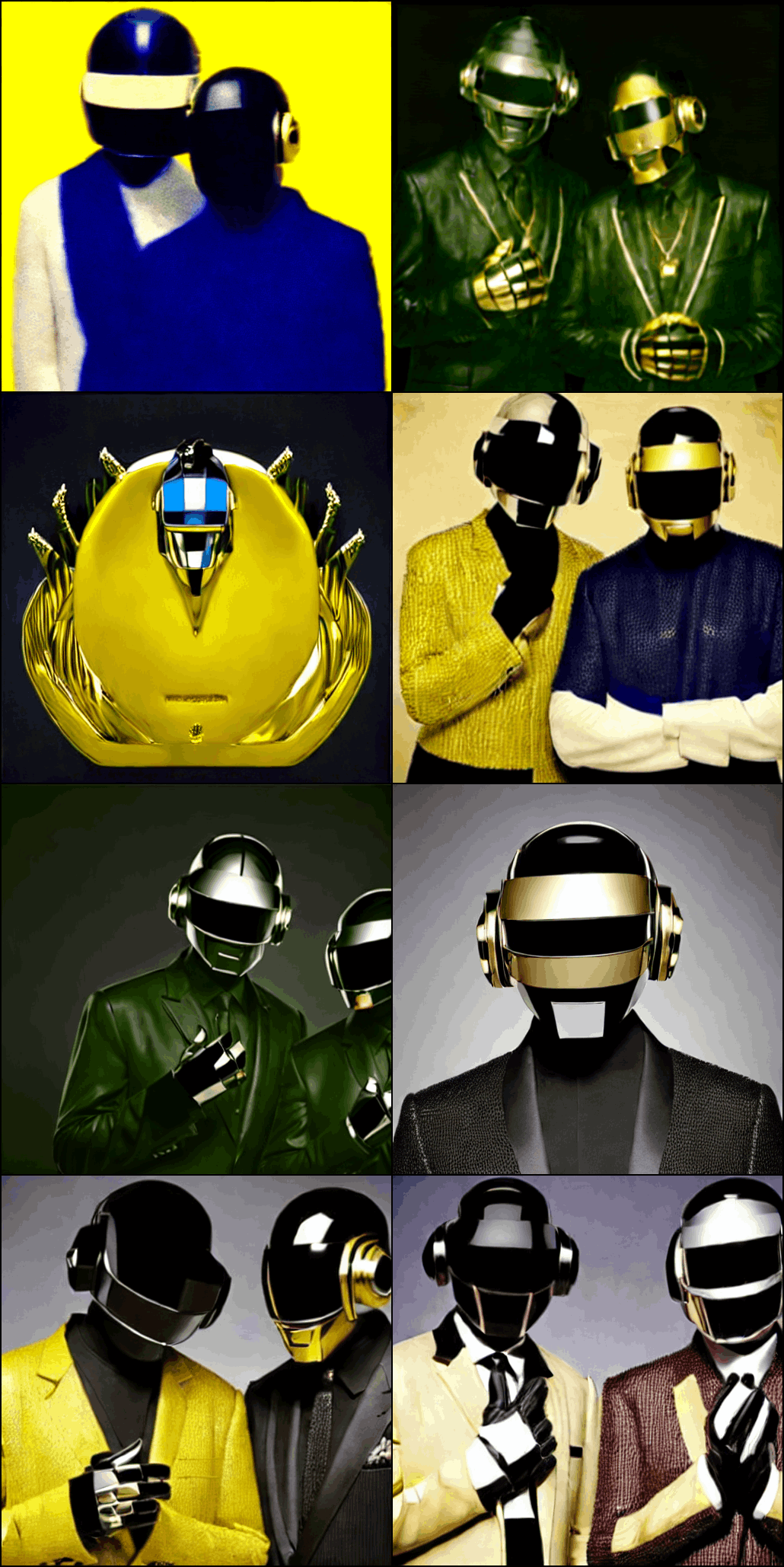}
        \end{minipage}%
        \hfill
        \caption{SISS + Ours}
    \end{subfigure}
        \caption{Visualization of (left) Partially-memorized and (right) Fully-memorized results after unlearning of the prompt ''Daft Punk, Jay Z Collaborate on \{\texttt{"}Computerized\texttt{"}\}".}
        \label{fig:sd3}
    \end{figure*}
    
    \clearpage
    
    {

    


\end{document}